\documentclass[11pt]{article}

\usepackage{lara-report}
\usepackage{afterpage} %

\newif\ifshowcomments
\showcommentsfalse     %

\newcommand{\model}{\texorpdfstring{\modelfont{Lara Mini base}}{Lara Mini base}\xspace}
\newcommand{\modelbase}{\texorpdfstring{\modelfont{Lara Mini base}}{Lara Mini base}\xspace}

\reporttitle{Lara Mini Base}
\reportsubtitle{Language diversity as a resource, not a trade-off}
\reportauthors{Translated, Lara Team\\[2pt]}
\reportshorttitle{Lara Mini base Technical Report}
\reportdate{November 2026}
\reportlinks{\translatedicon~\href{https://laratranslate.com/research}{Research}\qquad\hficon~\href{https://huggingface.co/translated/Lara-4-Mini-Base}{Model card}}
\hypersetup{pdftitle={Lara Mini base Technical Report}, pdfauthor={Translated}}

\begin{document}

\makereporttitle
\begin{abstract}
General-purpose language models are trained mostly on English: over open models we review, at most, non-English pretraining data makes up only a 12\% share.
Models built for European languages give them far more, but score well below general-purpose models of the same size.
A common concern is that, in small models, languages compete for capacity.
We present \model, a model trained from scratch on 6T tokens in English and 44 other languages, which together cover all 24 official languages of the European Union.
As training progresses, the 44 other languages together come to outweigh English.
To give these languages enough capacity at low cost, \model uses a mixture of experts architecture with 7.2B total and 0.6B active parameters.
Among open base models with up to 2B active parameters, it achieves the highest average score across standard English, multilingual, math, and code benchmarks, ahead of models trained on 4 to 6 times as many tokens. 
Its margin is largest on multilingual benchmarks in the languages it supports and on math benchmarks. It supports contexts of up to 128k tokens.
We find that our MoE architectures reduces capacity competition between languages.
For a single user on one GPU, without quantization, it generates text at least 1.8 times faster than any model with a higher average score.
We release the weights of \model under the CC BY-NC-SA 4.0 license.
In addition, we will release \textbf{T-HellaSwag}, a higher-quality machine translation of the commonsense reasoning benchmark HellaSwag into 110 language variants.

\end{abstract}

\begin{center}
  \begin{tikzpicture}
\newcommand{\lrb}[4]{%
  \fill[#3] (axis cs:#1-0.34,40) [rounded corners=2.4pt] -- (axis cs:#1-0.34,#2) -- (axis cs:#1+0.34,#2) [sharp corners] -- (axis cs:#1+0.34,40) -- cycle;
  \node[anchor=south, inner sep=1.6pt, font=\sffamily\bfseries\fontsize{6}{7}\selectfont, text=#4] at (axis cs:#1,#2) {#2};}
\newcommand{\lrlogo}[2]{%
  \begin{scope}
    \clip ($(axis cs:#1,#2)+(-3.6pt,-11pt)$) rectangle ($(axis cs:#1,#2)+(3.6pt,-3.8pt)$);
    \node[anchor=west, inner sep=0pt] at ($(axis cs:#1,#2)+(-3.6pt,-7.4pt)$) {\TranslatedLogo[white]{7.2pt}};
  \end{scope}}
\newcommand{\lrname}[2]{\node[rotate=45, anchor=north east, inner sep=1pt, font=\sffamily\fontsize{5.6}{6.5}\selectfont, text=chartsecondary] at (axis cs:#1,40) {#2};}
\begin{axis}[lrchart, scale only axis, width=0.95\linewidth, height=3.4cm,
  xmin=-0.5, xmax=19.40, ymin=40, ymax=72, ytick={40,50,60,70},
  ymajorgrids, grid style={draw=chartgrid, line width=0.4pt}, axis x line*=bottom, axis y line*=left,
  y axis line style={draw=none}, xtick=\empty, major tick length=0pt,
  yticklabel style={font=\sffamily\fontsize{5.5}{6}\selectfont}, clip=false]
\pgfplotsextra{
  \lrb{0.00}{57.0}{bottom color=brandgradstart, top color=brandgradend}{brandaccent}\lrlogo{0.00}{57.0}\lrname{0.00}{\textcolor{brandaccent}{\model}}
  \lrb{0.90}{53.9}{chartaxis!45}{chartmuted}\lrname{0.90}{granite-4.0-h-tiny}
  \lrb{1.80}{53.8}{chartaxis!45}{chartmuted}\lrname{1.80}{Qwen3-1.7B}
  \lrb{2.70}{53.2}{chartaxis!45}{chartmuted}\lrname{2.70}{Falcon-H1-1.5B-Deep}
  \lrb{3.60}{52.2}{chartaxis!45}{chartmuted}\lrname{3.60}{Hunyuan-1.8B}
  \lrb{5.10}{68.4}{chartaxis!45}{chartmuted}\lrname{5.10}{Trinity-Nano}
  \lrb{6.00}{66.4}{bottom color=brandgradstart, top color=brandgradend}{brandaccent}\lrlogo{6.00}{66.4}\lrname{6.00}{\textcolor{brandaccent}{\model}}
  \lrb{6.90}{66.2}{chartaxis!45}{chartmuted}\lrname{6.90}{granite-4.0-h-tiny}
  \lrb{7.80}{65.7}{chartaxis!45}{chartmuted}\lrname{7.80}{LFM2.5-8B-A1B}
  \lrb{8.70}{65.3}{chartaxis!45}{chartmuted}\lrname{8.70}{Falcon-H1-1.5B-Deep}
  \lrb{10.20}{56.7}{bottom color=brandgradstart, top color=brandgradend}{brandaccent}\lrlogo{10.20}{56.7}\lrname{10.20}{\textcolor{brandaccent}{\model}}
  \lrb{11.10}{53.3}{chartaxis!45}{chartmuted}\lrname{11.10}{Gemma-4-E2B}
  \lrb{12.00}{51.8}{chartaxis!45}{chartmuted}\lrname{12.00}{granite-4.0-h-tiny}
  \lrb{12.90}{50.3}{chartaxis!45}{chartmuted}\lrname{12.90}{Qwen3-1.7B}
  \lrb{13.80}{48.6}{chartaxis!45}{chartmuted}\lrname{13.80}{LFM2.5-8B-A1B}
  \lrb{15.30}{52.5}{bottom color=brandgradstart, top color=brandgradend}{brandaccent}\lrlogo{15.30}{52.5}\lrname{15.30}{\textcolor{brandaccent}{\model}}
  \lrb{16.20}{51.2}{chartaxis!45}{chartmuted}\lrname{16.20}{Falcon-H1-1.5B-Deep}
  \lrb{17.10}{50.9}{chartaxis!45}{chartmuted}\lrname{17.10}{Hunyuan-1.8B}
  \lrb{18.00}{50.6}{chartaxis!45}{chartmuted}\lrname{18.00}{Qwen3-1.7B}
  \lrb{18.90}{48.9}{chartaxis!45}{chartmuted}\lrname{18.90}{granite-4.0-h-tiny}
  \coordinate (g1) at (axis cs:1.80,40);
  \coordinate (g2) at (axis cs:6.90,40);
  \coordinate (g3) at (axis cs:12.00,40);
  \coordinate (g4) at (axis cs:17.10,40);
  \coordinate (pc) at (axis cs:16.65,55.5);
}
\end{axis}
\node[font=\sffamily\bfseries\scriptsize, text=white, fill=brandaccent, rounded corners=3pt, inner sep=3pt, text height=1.6ex, text depth=0.3ex, minimum width=3.2cm] at ($(g1)+(0,-1.95cm)$) {Overall};
\node[font=\sffamily\bfseries\scriptsize, text=brandink, fill=brandtint, rounded corners=3pt, inner sep=3pt, text height=1.6ex, text depth=0.3ex, minimum width=3.2cm] at ($(g2)+(0,-1.95cm)$) {English};
\node[font=\sffamily\bfseries\scriptsize, text=brandink, fill=brandtint, rounded corners=3pt, inner sep=3pt, text height=1.6ex, text depth=0.3ex, minimum width=3.2cm] at ($(g3)+(0,-1.95cm)$) {Multilingual};
\node[font=\sffamily\bfseries\scriptsize, text=brandink, fill=brandtint, rounded corners=3pt, inner sep=3pt, text height=1.6ex, text depth=0.3ex, minimum width=3.2cm] at ($(g4)+(0,-1.95cm)$) {Math and Code};
\node[anchor=south, font=\sffamily\scriptsize, text=chartsecondary, inner sep=0pt] at ($(current axis.north)+(0,0.25cm)$)
  {\textcolor{brandaccent}{\rule{6pt}{6pt}}~\model{} (A0.6B)\quad
   \textcolor{chartaxis!45}{\rule{6pt}{6pt}}~Best other models with up to 2B active parameters, one per organization};
\end{tikzpicture}
  \captionsetup{hypcap=false}
  
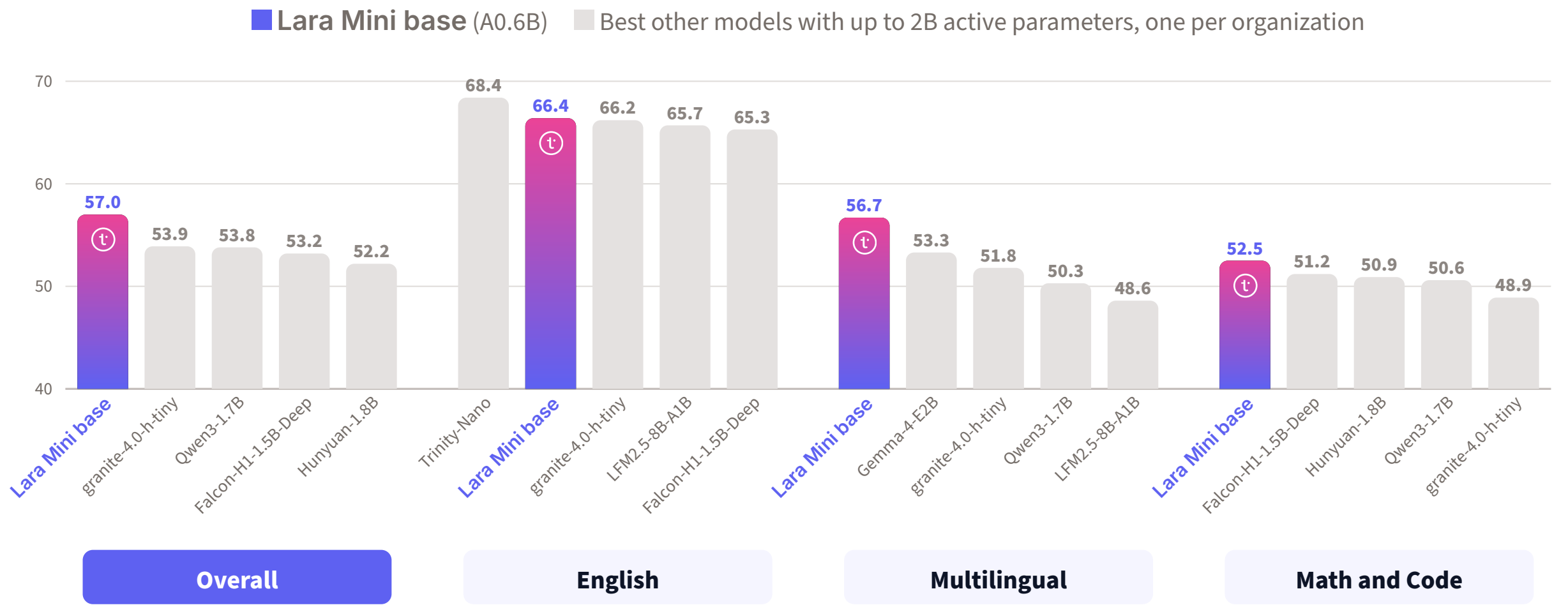
\captionof{figure}{Overall and domain averages of \model and of the four best other models with up to 2B active parameters on each average.}
  \label{fig:teaser}
\end{center}

\clearpage
{\hypersetup{linkcolor=brandink}\tableofcontents}
\clearpage

\section{Introduction}
\label{sec:introduction}

Small base language models have improved mostly by reading ever more tokens, and recent models with only a few billion parameters are trained on more than ten trillion of them~\citep{yang2025qwen3,hf2025smollm3}.
Because high-quality English text is limited, a growing budget also ends up repeating the same data, with diminishing returns~\citep{muennighoff2023scaling}.
Rather than reading the same English text again and again, we suggest that a small model should spend part of its budget on the text it has seen least: text in other languages.

Text in other languages may offer more than new tokens.
Research on multi-task learning has shown that related tasks can improve generalization by sharing learned representations~\citep{caruana1997multitask}.
Similarly, multilingual neural machine translation has demonstrated positive transfer between languages~\citep{firat2016multiway}.
More broadly, representation learning suggests that different observations can help a model discover shared explanatory factors underlying the data~\citep{bengio2013representation}.
These findings motivate a hypothesis: different languages may provide complementary perspectives on the same underlying reality, helping a shared model learn more generalizable representations.
Such benefits could potentially extend beyond multilingual tasks to English reasoning and other capabilities.

Even counting second-language speakers, English has only about 1.5 billion speakers: most of the world's population does not speak it~\citep{eberhard2025ethnologue}.
Text in these languages is also widely available in open corpora~\citep{nguyen2024culturax,penedo2025fineweb2,oepen2025hplt3}.
Yet the general-purpose open base models that disclose their mixture contribute at most 12\% of their tokens to this text (\cref{tab:ml-share}).
Models built for European languages give it 21\% to 51\%, but in our evaluation they score far below general-purpose models of the same size.
For example, EuroLLM-1.7B and Apertus-v1.1-1.5B reach an average benchmark score of 27.6 and 31.2, against 53.8 for Qwen3-1.7B (\cref{tab:app-full-results}).
At first glance, these scores suggest that multilingual data cost a small model its general ability.

This leaves an open question: can a model with a small active-parameter budget be strong across many languages without sacrificing its English?
Earlier work found that languages compete for the capacity of a small model, so adding languages can hurt each language~\citep {conneau2020xlmr,chang2024multilinguality}.
Recent scaling laws show that this trade-off depends on the size of the model and on how data are allocated across languages~\citep{longpre2026atlas}.
The answer, therefore, depends not only on how much multilingual data a model is trained on but also on how it is chosen and used.

We address this question with \model, a multilingual mixture of experts base model with 7.2B total and 0.6B active parameters, pretrained from scratch on 6T tokens.
We built \model around its data.
From openly released datasets, we assembled a deduplicated and filtered pool in English and 44 other languages, which together cover all 24 official languages of the European Union.
The first three training stages (main, transition and anneal) read 6T tokens from this pool and move toward high-quality, question-answer, code and math data.
English makes up only 41\% of the 6T tokens.
English falls from 44\% of the main stage to 27\% of the anneal, so that the other languages grow from 46\% to 55\% of the text and outweigh English in the anneal (\cref{sec:data}).
Of the 6T training tokens, 97\% are the first read of a document (\cref{sec:data-mixture}).
A fourth and last stage of 100B tokens then extends the context of \model to 128k tokens (\cref{sec:context-extension}).
To support many languages while keeping inference cost low, every layer of \model has a fine-grained mixture of experts that activates 4 of 128 experts per token.
To keep the inference state small, the layers interleave Gated DeltaNet~\citep{yang2025gdn}, sliding-window and global attention, the latter in only 7 of the 34 layers (\cref{sec:architecture}).

We compare \model with 56 open base models on standard English, multilingual, math and code benchmarks, all evaluated by us with the same harness and settings.
As in \citet{muennighoff2025olmoe}, we group the models by active parameters, as a rough approximation of inference cost (\cref{sec:eval-setup}).
Among the 36 models with up to 2B active parameters, \model ranks first on the average of the four domains, with 57.0, 3.1 points above the next model (\cref{fig:teaser}).
It also ranks first on the multilingual average in the languages it supports, 3.4 points ahead.
It leads even though this group includes models with more than 3 times its active parameters.
The next two models, granite-4.0-h-tiny and Qwen3-1.7B, were trained on 22.5T and 36T tokens, about 4 and 6 times as many (\cref{tab:main-results}).
\model also outscores several models with 6 to 13 times its active parameters, such as Ministral-3-3B (53.8), Gemma-4-E4B (55.4) and Llama-3.1-8B (51.2).
Among the models we evaluated from European developers, it is the strongest, ahead of Apertus-v1.5-8B (55.8) and Domyn-Edge (55.4).
Although English accounts for less than half of its training data, \model ranks second in its group for English.
When the multilingual third of the anneal is replaced with English, the English average rises by only 0.3 points and the multilingual average falls by 2.7 (\cref{sec:abl-multilingual}).
An expert routing analysis reveals that experts are shared across languages, which is consistent with this ablation hardly deteriorating performance on English tasks (\cref{sec:abl-routing}).

These results come at a small inference cost.
\model needs the compute per token of a small dense model, although all its parameters must be held in memory.
For a single user on one A100 GPU, without quantization, it generates about 1{,}100 characters per second (\cref{fig:pareto-speed}).
This speed matches that of dense models with about 1B parameters and is at least 1.8 times that of every model that scores higher on average.

We release the weights of \model under the CC BY-NC-SA 4.0 license\footnote{\url{https://creativecommons.org/licenses/by-nc-sa/4.0/}}, which allows research and other non-commercial use.
We will also release T-HellaSwag, a machine translation of the commonsense reasoning benchmark HellaSwag~\citep{zellers2019hellaswag} into 110 language variants, of higher quality than the commonly used Multilingual HellaSwag~\citep{lai2023okapi} (\cref{app:t-hellaswag}).

\Cref{sec:data} describes the pretraining data, \cref{sec:architecture} the architecture and \cref{sec:pretraining} training operations.
\Cref{sec:evaluation} reports evaluation methods and results. \cref{sec:ml-analysis} analyses the benefits of our data choices on multilingual and English tasks. The Appendix contains additional technical detail aimed at sharing our insights with the open LLM community.

\section{Pretraining Data}
\label{sec:data}

Unlike general-purpose models, \model is not trained mostly on English: English accounts for only 41\% of its 6T training tokens.\footnote{The English domain of \cref{tab:data-domains}. Code, math and SFT-like data, mostly in English, are counted as separate domains.}
The 44 other languages account for at least three times the share of general-purpose open base models that disclose them (\cref{tab:ml-share}).
In absolute terms, \model processes more multilingual tokens than Llama~3, SmolLM3, or Nemotron Nano~2, even though they train on 11.2T to 20T tokens.
Its share is within the range of the models for European languages in \cref{tab:ml-share}, all of which have at least 1.7B active parameters, against 0.6B for \model.
Smaller multilingual models exist, such as EuroMoE-2.6B-A0.6B and Apertus-v1.1-0.5B, but score far below \model (\cref{tab:app-full-results}).

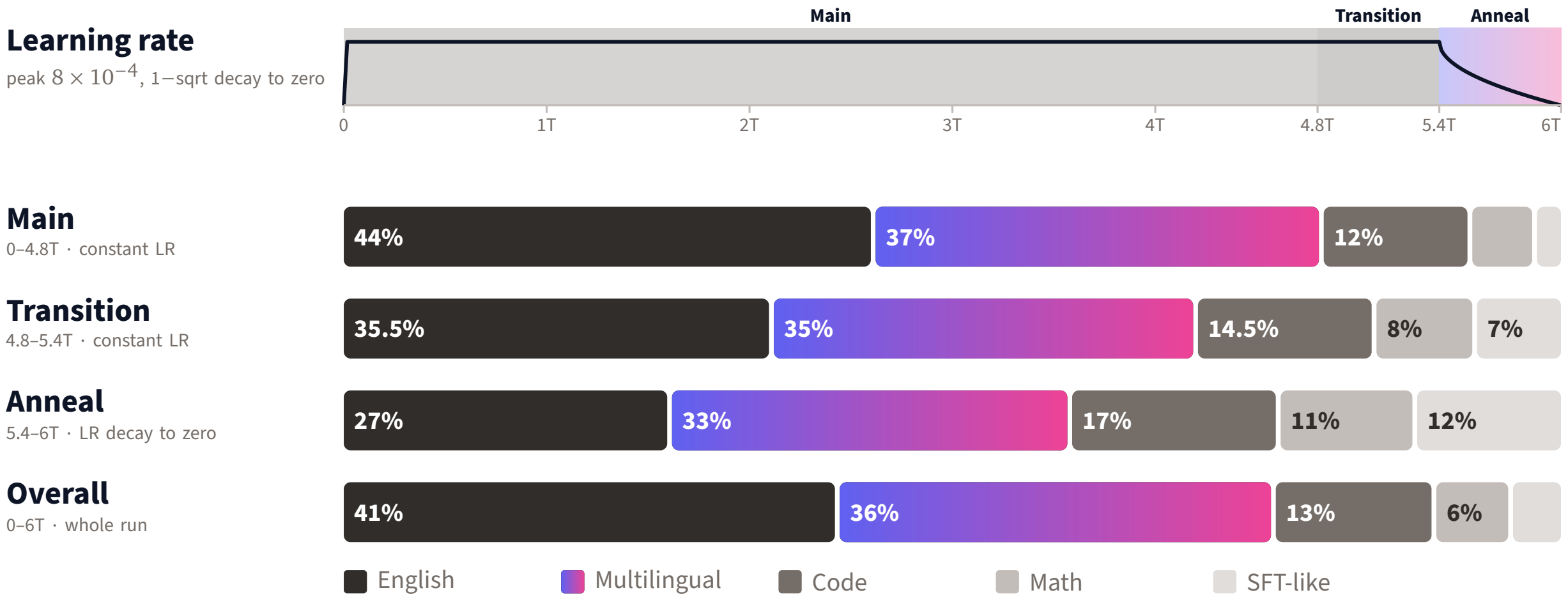
\begin{figure}[t]
  \centering
  \begin{tikzpicture}[font=\sffamily]
  \def\lrbarx{3.5}  %
  \def\lrbarw{12.6} %
  \def\lrgap{0.05}  %
  \def\lrbarh{0.62} %
  \pgfmathsetmacro\lrxs{\lrbarw/6} %

  \def\lrty{1.05}   %
  \def\lrth{0.80}   %
  \def\lrpk{0.82}   %
  \fill[ordinal1!35] (\lrbarx, \lrty) rectangle (\lrbarx+4.8*\lrxs, \lrty+\lrth);
  \fill[ordinal2!35] (\lrbarx+4.8*\lrxs, \lrty) rectangle (\lrbarx+5.4*\lrxs, \lrty+\lrth);
  \fill[gradtint=35] (\lrbarx+5.4*\lrxs, \lrty) rectangle (\lrbarx+6*\lrxs, \lrty+\lrth);
  \draw[brandink, line width=1.1pt, line join=round]
    (\lrbarx, \lrty) -- (\lrbarx+0.017*\lrxs, \lrty+\lrpk*\lrth) -- (\lrbarx+5.4*\lrxs, \lrty+\lrpk*\lrth)
    -- plot[domain=5.4:6, samples=61] ({\lrbarx+\x*\lrxs}, {\lrty+\lrpk*\lrth*(1-sqrt(max(\x-5.4,0)/0.6))});
  \draw[chartaxis, line width=0.6pt] (\lrbarx, \lrty) -- (\lrbarx+\lrbarw, \lrty);
  \foreach \t/\lab in {0/0, 1/1T, 2/2T, 3/3T, 4/4T, 4.8/4.8T, 5.4/5.4T, 6/6T} {
    \draw[chartaxis, line width=0.6pt] (\lrbarx+\t*\lrxs, \lrty) -- ++(0,-0.08);
  }
  \foreach \t/\lab in {0/0, 1/1T, 2/2T, 3/3T, 4/4T, 4.8/4.8T, 5.4/5.4T} {
    \node[anchor=north, font=\sffamily\tiny, text=chartsecondary, inner sep=1.5pt]
      at (\lrbarx+\t*\lrxs, \lrty-0.08) {\lab};
  }
  \node[anchor=north east, font=\sffamily\tiny, text=chartsecondary, inner sep=1.5pt]
    at (\lrbarx+\lrbarw+0.06, \lrty-0.08) {6T};
  \foreach \a/\b/\name in {0/4.8/Main, 4.8/5.4/Transition, 5.4/6/Anneal} {
    \node[anchor=south, font=\sffamily\tiny\bfseries, text=brandink, inner sep=1.5pt]
      at (\lrbarx+0.5*\a*\lrxs+0.5*\b*\lrxs, \lrty+\lrth) {\name};
  }
  \node[anchor=north west, inner sep=0pt, font=\sffamily\small\bfseries, text=brandink]
    at (0, \lrty+\lrth+0.02) {Learning rate};
  \node[anchor=north west, inner sep=0pt, font=\sffamily\tiny, text=chartsecondary, text width=3.3cm]
    at (0, \lrty+\lrth-0.36) {peak $8 \times 10^{-4}$, 1$-$sqrt decay to zero};

  \def\lrstagebar#1#2{%
    \gdef\lrcursor{\lrbarx}%
    \pgfmathsetmacro\lravail{\lrbarw-4*\lrgap}%
    \foreach \s/\f/\t in {#2} {%
      \pgfmathsetmacro\lrw{\s/100*\lravail}%
      \pgfmathsetmacro\lrend{\lrcursor+\lrw}%
      \fill[\f, rounded corners=2pt] (\lrcursor, #1) rectangle (\lrend, #1-\lrbarh);
      \pgfmathtruncatemacro\lrfits{\lrw>0.62}%
      \ifnum\lrfits=1
        \node[font=\sffamily\scriptsize\bfseries, text=\t, anchor=west, inner sep=0pt]
          at (\lrcursor+0.1, #1-\lrbarh/2) {\s\%};
      \fi
      \pgfmathsetmacro\lrnext{\lrend+\lrgap}%
      \xdef\lrcursor{\lrnext}%
    }}
  \foreach \y/\stage/\detail/\oc in {0/Main/{0--4.8T $\cdot$ constant LR}/stage1,
                                     -0.95/Transition/{4.8--5.4T $\cdot$ constant LR}/stage2,
                                     -1.90/Anneal/{5.4--6T $\cdot$ LR decay to zero}/stage3,
                                     -2.85/Overall/{0--6T $\cdot$ whole run}/stageall} {
    \node[anchor=north west, inner sep=0pt, font=\sffamily\small\bfseries, text=brandink] at (0, \y+0.02) {\stage};
    \node[anchor=north west, inner sep=0pt, font=\sffamily\tiny, text=chartsecondary, text width=3.0cm] at (0, \y-0.36) {\detail};
  }
  \lrstagebar{0}{44/domen/white, 37/domml/white, 12/domcode/white, 5/dommath/chartink, 2/domsft/chartink}
  \lrstagebar{-0.95}{35.5/domen/white, 35/domml/white, 14.5/domcode/white, 8/dommath/chartink, 7/domsft/chartink}
  \lrstagebar{-1.90}{27/domen/white, 33/domml/white, 17/domcode/white, 11/dommath/chartink, 12/domsft/chartink}
  \lrstagebar{-2.85}{41/domen/white, 36/domml/white, 13/domcode/white, 6/dommath/chartink, 4/domsft/chartink}
  \foreach \i/\c/\name in {0/domen/English, 1/domml/Multilingual, 2/domcode/Code, 3/dommath/Math, 4/domsft/SFT-like} {
    \fill[\c, rounded corners=1pt] (\lrbarx+\i*2.25, -4.00) rectangle ++(0.24, 0.24);
    \node[anchor=west, inner sep=0pt, font=\sffamily\scriptsize, text=chartsecondary] at (\lrbarx+\i*2.25+0.34, -3.88) {\name};
  }
\end{tikzpicture}
  \caption{Learning-rate schedule over the 6T training tokens, with the three stages shaded (top), and the domain shares of each stage (bottom); all values are in \cref{tab:app-groups}. Code, math and SFT-like data grow from 19\% to 40\% of the tokens, while multilingual text stays at about a third and outweighs English in the anneal.}
  \label{fig:stage-mixture}
\end{figure}

All data come from openly released datasets (\cref{tab:data-domains,sec:data-sources}), which Translated's internal pipeline deduplicates and filters into a training pool of 13.8T tokens (\cref{sec:data-filtering}).
We sample 6T tokens from this pool in three stages (main, transition and anneal), moving toward high-quality data (\cref{fig:stage-mixture}).
Explicit rules set the mixture of each stage (\cref{tab:mixture-design,sec:data-mixture}).
A fourth stage of 100B tokens then extends the context from 4,096 to 128k tokens (\cref{sec:data-long-context}).

\begin{table}[t]
  \centering
  \small
  \caption{Share of multilingual text (natural language other than English, without parallel data) in the pretraining data of open base models, based on technical reports. Active (total) parameters: total parameters in parentheses for mixtures of experts. Qwen3~\citep{yang2025qwen3}, Gemma~3~\citep{gemmateam2025gemma3} and DeepSeek-V3~\citep{deepseekai2024v3} do not disclose the share. The multilingual share of \model is three times the largest share among the general-purpose models, with fewer active parameters than any model for European languages.}
  \label{tab:ml-share}
  \begin{tabularx}{\linewidth}{@{}Lrrrr@{}}
    \toprule
    \textbf{Model} & \textbf{\begin{tabular}[b]{@{}r@{}}Active (total)\\parameters\end{tabular}} & \textbf{\begin{tabular}[b]{@{}r@{}}Training\\tokens\end{tabular}} & \textbf{\begin{tabular}[b]{@{}r@{}}Multilingual\\share\end{tabular}} & \textbf{\begin{tabular}[b]{@{}r@{}}Multilingual\\tokens\end{tabular}} \\
    \midrule
    \multicolumn{5}{@{}l}{\textit{General-purpose models}} \\
    SmolLM2~\citep{allal2025smollm2}                & 1.7B           & 11T   & 0\%          & -- \\
    Olmo~3~\citep{olmo2025olmo3}                    & 7B, 32B        & 5.9T  & 0\%          & -- \\
    Nemotron Nano~2~\citep{nvidia2025nemotronnano2} & 12B            & 20T   & 5\%          & 1.0T \\
    Llama~3~\citep{grattafiori2024llama3}           & 8B--405B       & 15.6T & 8\%          & 1.2T \\
    SmolLM3~\citep{hf2025smollm3}                   & 3B             & 11.2T & 12\%         & 1.3T \\
    \midrule
    \multicolumn{5}{@{}l}{\textit{Models for European languages}} \\
    Kolibri~\citep{alephalpha2026kolibri}           & 3.5B (78B)     & 20T   & 21\%         & 4.3T \\
    EuroLLM~\citep{martins2024eurollm,martins2025eurollm9b} & 1.7B, 9B      & 4T    & 37\%         & 1.5T \\
    Apertus~\citep{apertus2025}                     & 8B, 70B        & 15T   & 40\%         & 6.0T \\
    Teuken-7B~\citep{ali2024teuken}                 & 7B             & 6T    & 51\%         & 3.0T \\
    \midrule
    \ourrow \model                                  & 0.6B (7.2B)    & 6T    & 36\%         & 2.18T \\
    \bottomrule
  \end{tabularx}
\end{table}

\begin{table}[t]
  \centering
  \small
  \caption{Data domains: main sources, tokens in the training pool and tokens read in training with their share of the 6T. SFT-like: question-answer, math and code examples in the format of supervised fine-tuning data. All sources, with their licenses, are listed in \cref{tab:data-sources}; tokens per data group and stage are in \cref{tab:app-groups}. Math is the only domain read more than once.}
  \label{tab:data-domains}
  \begin{tabularx}{\linewidth}{@{}lLrr@{}}
    \toprule
    \textbf{Domain} & \textbf{Main sources} & \textbf{Pool} & \textbf{Training} \\
    \midrule
    English      & Nemotron-CC and its synthetic data~\citep{su2024nemotroncc}, FineWeb-Edu~\citep{penedo2024fineweb}, FinePDFs~\citep{hf2025finepdfs}, FineWiki~\citep{hf2025finewiki} & 5.65T & 2.49T (41.4\%) \\
    Multilingual & FineWeb-2~\citep{penedo2025fineweb2}, FineWeb2-HQ~\citep{messmer2025fineweb2hq}, HPLT~3.0~\citep{oepen2025hplt3}, CulturaX~\citep{nguyen2024culturax}, FinePDFs, FineWiki, synthetic question-answer data & 6.31T & 2.18T (36.4\%) \\
    Code         & Nemotron-CC code pages, The Stack~v3~\citep{hf2026stackv3} (permissive files), Nemotron synthetic code & 0.90T & 0.77T (12.8\%) \\
    Math         & FineMath~\citep{allal2025smollm2}, MegaMath~\citep{zhou2025megamath}, Nemotron-CC-Math~\citep{mahabadi2026nemotronccmath} & 0.33T & 0.35T (5.9\%) \\
    SFT-like     & Nemotron pretraining collections~\citep{nvidia2025nemotronnano2,nvidia2026specializedv11,nvidia2026specializedv12} & 0.61T & 0.21T (3.5\%) \\
    \midrule
    Total        & & 13.8T & 6T \\
    \bottomrule
  \end{tabularx}
\end{table}

\subsection{Data Sources}
\label{sec:data-sources}

Besides web, PDF and Wikipedia text (\cref{tab:data-domains}), 26\% of the training tokens (1.58T) are synthetic, most of them from NVIDIA's Nemotron collections.
The largest part is English web text rewritten by language models; the rest is code, math, SFT-like data (examples in the format of supervised fine-tuning data) and multilingual question-answer data.
The multilingual question-answer data make up 13\% of the multilingual tokens and are mostly machine-translated from English into 16 of our languages.
\Cref{app:data} lists every data group with its sources and licenses (\cref{tab:data-sources}) and shows how deduplication and selection reduce the collected data to the pool (\cref{fig:data-pipeline}).

\begin{table}[t]
  \centering
  \small
  \caption{Design of the data mixture: each choice and its basis. Shares are given per stage, in the order main, transition, anneal (\cref{fig:stage-mixture}).}
  \label{tab:mixture-design}
  \begin{tabularx}{\linewidth}{@{}l>{\raggedright\arraybackslash}p{3.1cm}L@{}}
    \toprule
    \textbf{Decision} & \textbf{Choice} & \textbf{Basis} \\
    \midrule
    English : multilingual & 54:46 between English and multilingual text, excluding code, math and SFT-like data (main stage) & Most high-quality and synthetic data are English; in small proxy runs, shifting tokens between English and multilingual text hurt the side that lost them more than it helped the other \\
    \addlinespace[3pt]
    Code & 12\%, 14.5\%, 17\% & Code improves natural-language reasoning~\citep{aryabumi2025code}; SmolLM3 uses 12--24\%~\citep{hf2025smollm3}, Llama~3 17\%~\citep{grattafiori2024llama3} \\
    \addlinespace[3pt]
    Math & 5\%, 8\%, 11\% & Limited by supply: about 330B unique tokens \\
    \addlinespace[3pt]
    SFT-like data & 2\%, 7\%, 12\% & High-quality and question-answer data help most at the end of training~\citep{blakeney2024doesyourdata,grattafiori2024llama3,olmo2025olmo2,nvidia2025nemotronnano2} \\
    \addlinespace[3pt]
    Multilingual anneal & 33\%, mostly high-quality web and question-answer data & In a 100B-token continuation of a 1B model, translated question-answer data scored 10 points higher on Global MMLU than heuristically filtered Common Crawl text~\citep{nvidia2025nemotronnano2} \\
    \bottomrule
  \end{tabularx}
\end{table}

\subsection{Deduplication and Filtering}
\label{sec:data-filtering}

We remove exact copies over the whole pool at once, so that a text kept from one source is removed from every other.
We then remove near-duplicates with the MinHash~\citep{broder1997resemblance,lee2022dedup} implementation of Marin~\citep{marin2025}, within each language across all its sources, and within code.
Synthetic and rewritten data are not compared for near-duplicates with the pages they rephrase.
Deduplication keeps about 30T of the 50T collected tokens.
Most duplicates occur \emph{across} sources, so per-source deduplication would miss them: FineWeb-Edu, for instance, keeps only 15\% of its tokens after deduplication against the other sources.

For quality, we use the model-based quality signals released with the main corpora.
In English web data, high quality means the two highest classes of Nemotron-CC and FineWeb-Edu.
The medium class of Nemotron-CC is used only before the anneal.
In multilingual web data, high quality means FineWeb2-HQ, the pages that a model-based filter ranks highest in each language.
Math is split in the same way by the quality tiers of its sources.
On top of these signals, Translated's internal pipeline applies further filters of its own, built on the company's experience with multilingual data.

The selected data, tokenized with the Qwen3 tokenizer, form the training pool (\cref{tab:app-groups}).
Besides all the high-quality data, the pool holds more plain web text (web data below the high-quality level) than training needs.
Training reads only about a fifth of this plain web text.

\paragraph{Decontamination.}
\label{sec:data-decontam}
We scanned the whole training pool for the items of our evaluation benchmarks.
The scan follows \citet{brown2020gpt3}: it matches long word n-grams (character n-grams in scripts written without spaces) and flags the documents that share a large part of a benchmark item.
Flagged documents make up about 0.03\% of the pool's tokens.
Benchmarks whose items circulate widely on the web, mostly in English, math and code, are the most exposed, while the benchmarks of our multilingual domain are much less so.
Of the first three stages, only the anneal is decontaminated, as in Kolibri~\citep{alephalpha2026kolibri}: it skips training sequences that contain a flagged document without changing the mixture.
The tests in \cref{sec:contamination} show no evidence that \model memorized the contaminated items: its scores are consistent with generalization and not inflated by contamination.

\subsection{Data Mixture and Curriculum}
\label{sec:data-mixture}

We designed the mixture to achieve competence in 44 languages besides English, with priority given to the official EU languages.
It also had to keep English knowledge and reasoning on par with English-centric models of the same size, while adding basic code and math abilities.
We set it with explicit rules rather than by optimizing mixture weights on proxy models.
Each rule rests on published evidence, data availability, or our priorities among languages (\cref{tab:mixture-design}).
\Cref{fig:stage-mixture} shows the resulting mixture of each stage.

Of the four stages of pretraining, the first three read 6T tokens in 4{,}096-token sequences.
The \emph{main} stage, 4.8T tokens at constant learning rate, uses the broadest mixture.
The \emph{transition} stage, with 0.6T tokens still at a constant learning rate, moves the mixture halfway toward the anneal mixture.
The \emph{anneal} of the last 0.6T tokens decays the learning rate to zero.
As in earlier work, the learning rate decays only at the end~\citep{hu2024minicpm,hagele2024scaling} and training ends on high-quality data~\citep{feng2024twophase}.
A sudden change in data can cause a model to forget what it learned earlier, and keeping part of the earlier mixture reduces this forgetting~\citep{ibrahim2024simple}.
The transition stage therefore splits the change into two smaller steps, as the two-blend schedule of \citet{parmar2024reuse} does in continued pretraining.
Over the three stages, code, math and SFT-like data grow from 19\% to 40\% of the tokens, and the anneal keeps only high-quality English web data.
The multilingual share stays at about a third while English falls from 44\% to 27\%, so that in the anneal multilingual text is the largest domain, ahead of English (\cref{fig:stage-mixture}).
Of the text in English and the other languages, without code, math and SFT-like data, the other languages make up 46\% in the main stage, 50\% in the transition and 55\% in the anneal.
Its content also shifts toward high-quality sources: plain web text falls from 41\% of the multilingual tokens to 6.5\%, while high-quality web and question-answer data rise from 39\% to 69\%.

We include every language with at least 20B tokens after deduplication.
Irish and Maltese have far fewer speakers, but we include them so that all 24 official languages of the European Union are covered.
This gives 44 languages besides English (\cref{tab:app-languages}).
Sampling languages by size would give Russian and Chinese a third of the multilingual tokens.
Flatter sampling trades quality in high-resource languages for quality in low-resource ones~\citep{conneau2020xlmr}.
We therefore sample languages more evenly than by size, with extra weight given to the official EU languages and to a selection of widely spoken languages outside the EU.
Compared with size-based sampling, these weights raise the share of official EU languages from 46\% to 61\% of multilingual tokens, mostly at the expense of Russian, Chinese and Japanese (\cref{fig:languages}).

Because mixtures of experts are sensitive to repeated data~\citep{jha2026datascarcity}, we cap the epochs of each source within the 6T budget, as in UniMax~\citep{chung2023unimax}.
No source is read more than twice except Wikipedia, and plain web text is repeated only in Irish and Maltese.
In all, 97\% of the training tokens are the first read of a document (\cref{tab:app-groups}).

\paragraph{Long-context stage.}
\label{sec:data-long-context}
The fourth stage reads 100B tokens in two steps: 40B in 32k-token sequences, then 60B in 128k-token sequences (\cref{sec:context-extension}).
As in Kolibri~\citep{alephalpha2026kolibri}, both steps mix the anneal mixture as is, the same mixture reweighted toward long documents, and natural long documents from FinePDFs~\citep{hf2025finepdfs} and, in the 32k step only, FineWeb2-HQ~\citep{messmer2025fineweb2hq} (\cref{tab:ce-recipe}).
Both steps also give about 11\% of their tokens to math and code: Nemotron-CC-Math documents of up to 16k tokens~\citep{mahabadi2026nemotronccmath}, real (not synthetic) code and, in the 32k step only, SFT-like math documents of 16k to 32k tokens (\cref{tab:data-sources}).
Keeping the anneal mixture limits the loss of short-context quality~\citep{gao2025prolong,olmo2025olmo3}.
It therefore makes up half of the 128k step, which ends training.
Across the whole stage, 58\% of tokens come from documents longer than 4k tokens, and 12.3B tokens come from documents of at least 128k tokens.

\section{Architecture}
\label{sec:architecture}

The architecture of \model follows two goals: capacity for many languages at a small active budget, and a small inference state.
The mixture of experts (MoE) layers serve the first: in our small-scale comparisons, a fine-grained MoE outperformed dense models with more than twice its active parameters (\cref{sec:abl-architecture}).
The hybrid token mixing serves the second.
Only 7 of the 34 layers use global attention, so the memory held per sequence grows with context length about five times more slowly than in a full-attention model of the same shape.

\begin{figure}[!t]
  \centering
  \begin{tikzpicture}[
    font=\sffamily\scriptsize, >={Stealth[length=4pt, width=3.5pt]},
    blk/.style={rectangle, rounded corners=3pt, draw=chartsecondary, line width=0.5pt,
      minimum width=3.3cm, minimum height=0.62cm, align=center, text=brandink, inner sep=2pt},
    nrm/.style={blk, fill=white, minimum height=0.48cm},
    add/.style={circle, draw=chartsecondary, fill=white, line width=0.5pt, inner sep=0pt,
      minimum size=0.34cm, font=\sffamily\footnotesize, text=chartsecondary},
    flow/.style={->, draw=chartsecondary, line width=0.6pt},
    note/.style={text=chartmuted, align=left, font=\sffamily\scriptsize},
    lead/.style={draw=chartmuted, dash pattern=on 1.5pt off 1.5pt, line width=0.4pt},
    pnl/.style={rectangle, rounded corners=4pt, draw=brandrule, fill=brandtint!60, line width=0.5pt},
    mix/.style={rectangle, rounded corners=2.5pt, draw=chartsecondary, line width=0.5pt,
      minimum width=4.6cm, minimum height=0.82cm, align=left, text width=4.3cm, inner sep=3pt},
    exp/.style={rectangle, rounded corners=1.5pt, draw=chartsecondary, line width=0.4pt,
      minimum width=0.42cm, minimum height=0.42cm, inner sep=0pt},
    act/.style={bottom color=brandgradstart, top color=brandgradend},
    idle/.style={fill=chartgrid},
  ]
  \node[text=chartsecondary] (tok) at (0, 0) {Tokenized text};
  \node[blk, fill=series5!18] (emb) at (0, 0.85) {Token embedding layer};

  \fill[brandtint!70, rounded corners=5pt] (-2.15, 1.55) rectangle (2.15, 7.35);
  \draw[brandrule, rounded corners=5pt, line width=0.6pt] (-2.15, 1.55) rectangle (2.15, 7.35);
  \node[font=\sffamily\small\bfseries, text=brandink, anchor=north east] at (2.12, 7.3) {34$\times$};

  \node[nrm] (n1) at (0, 2.15) {RMSNorm};
  \node[blk, fill=white, minimum height=0.8cm, text depth=0.32cm] (mx) at (0, 3.25) {\textbf{Token mixer}};
  \foreach \c/\x in {series1/-0.45,series2/0,series3/0.45}
    \fill[\c!70, rounded corners=1pt] (\x-0.17, 2.98) rectangle (\x+0.17, 3.1);
  \node[add] (a1) at (0, 4.2) {$+$};
  \node[nrm] (n2) at (0, 4.85) {RMSNorm};
  \node[blk, fill=series4!16, minimum height=0.8cm] (moe) at (0, 5.95) {\textbf{Mixture of experts}};
  \node[add] (a2) at (0, 6.9) {$+$};

  \node[nrm] (fn) at (0, 8.0) {Final RMSNorm};
  \node[blk, fill=series5!18] (out) at (0, 8.85) {Linear output layer};
  \node[text=chartsecondary] (lg) at (0, 9.7) {Next-token logits};

  \draw[flow] (tok) -- (emb);
  \draw[flow] (emb) -- (n1);
  \draw[flow] (n1) -- (mx);
  \draw[flow] (mx) -- (a1);
  \draw[flow] (a1) -- (n2);
  \draw[flow] (n2) -- (moe);
  \draw[flow] (moe) -- (a2);
  \draw[flow] (a2) -- (fn);
  \draw[flow] (fn) -- (out);
  \draw[flow] (out) -- (lg);
  \draw[flow] (0, 1.72) -- (-1.9, 1.72) -- (-1.9, 4.2) -- (a1.west);
  \draw[flow] (0, 4.45) -- (-1.9, 4.45) |- (a2.west);

  \node[pnl, minimum width=5.0cm, minimum height=5.75cm, anchor=north east] (lp) at (-2.6, 7.3) {};
  \node[anchor=north west, text=brandink, font=\sffamily\scriptsize\bfseries]
    at ($(lp.north west)+(0.15,-0.08)$) {Token mixers};
  \node[mix, fill=series3!18, anchor=north] (mG) at ($(lp.north)+(0,-0.5)$)
    {\textbf{Global attention}\\[-1pt]{\color{chartsecondary}7 layers, no positional encoding}};
  \node[mix, fill=series2!18, below=0.14cm of mG] (mW)
    {\textbf{Sliding-window attention}\\[-1pt]{\color{chartsecondary}13 layers, 512 tokens, RoPE}};
  \node[mix, fill=series1!18, below=0.14cm of mW] (mD)
    {\textbf{Gated DeltaNet}\\[-1pt]{\color{chartsecondary}14 layers, recurrent state}};
  \node[note, below=0.06cm of mD, text width=4.6cm]
    {Attention: 16 query / 4 key-value heads, QK-norm};
  \node[anchor=south west, text=brandink, font=\sffamily\scriptsize\bfseries]
    at ($(lp.south west)+(0.15,0.8)$) {Layer pattern};
  \foreach \i/\c in {1/series1,2/series2,3/series1,4/series3,5/series2,6/series1,7/series2,8/series1,9/series3,10/series2,11/series1,12/series2,13/series1,14/series3,15/series2,16/series1,17/series2,18/series1,19/series3,20/series2,21/series1,22/series2,23/series1,24/series3,25/series2,26/series1,27/series2,28/series1,29/series3,30/series2,31/series1,32/series2,33/series1,34/series3}
    \fill[\c, rounded corners=0.5pt] ($(lp.south west)+(0.2+0.136*\i-0.136,0.46)$) rectangle ++(0.11,0.3);
  \node[note, anchor=north west, font=\sffamily\tiny] at ($(lp.south west)+(0.15,0.45)$) {layer 1};
  \node[note, anchor=north east, font=\sffamily\tiny] at ($(lp.south east)+(-0.1,0.45)$) {layer 34};
  \draw[lead] (mx.west) -- (-2.6, 3.25);

  \node[pnl, minimum width=5.0cm, minimum height=3.75cm, anchor=north west] (rp) at (2.6, 7.3) {};
  \node[anchor=north west, text=brandink, font=\sffamily\scriptsize\bfseries]
    at ($(rp.north west)+(0.15,-0.08)$) {Mixture of experts};
  \node[blk, fill=series4!16, minimum width=2.2cm, minimum height=0.48cm] (rt)
    at ($(rp.north west)+(1.4,-0.75)$) {Router, top-4};
  \node[note, anchor=west] at ($(rt.east)+(0.1,0)$) {fp32};
  \foreach \k/\f in {0/act,1/idle,2/act,3/idle,4/idle,5/act,6/idle,7/act} {
    \node[exp, \f] (e\k) at ($(rp.north west)+(0.45+0.5*\k,-1.65)$) {};
  }
  \node[text=chartsecondary] at ($(e7.east)+(0.3,0)$) {\dots};
  \node[note, anchor=north west] at ($(e0.south west)+(-0.05,-0.04)$) {128 routed experts, 4 active};
  \node[blk, fill=chartink, text=white, minimum width=1.9cm, minimum height=0.48cm, anchor=west] (sh)
    at ($(rp.north west)+(0.25,-2.75)$) {Shared expert};
  \node[note, anchor=west] at ($(sh.east)+(0.1,0)$) {every token};
  \node[note, anchor=south west] at ($(rp.south west)+(0.15,0.1)$) {All experts: SwiGLU, hidden size 512};
  \draw[lead] (moe.east) -- (2.6, 5.95);

  \draw[lead] (emb.east) -- (2.75, 0.85);
  \node[note, anchor=west] at (2.8, 0.85) {Vocabulary of 151{,}936 tokens\\embedding dimension 1{,}024};
  \draw[lead] (out.east) -- (2.75, 8.85);
  \node[note, anchor=west] at (2.8, 8.85) {Tied with the token embedding};
  \node[note, anchor=west] at (2.8, 2.6) {Context length 4{,}096 in pretraining,\\131{,}072 after extension\\7.2B parameters, 0.6B active};
\end{tikzpicture}
  \caption{Architecture of \model. Center: each of the 34 layers applies a token mixer and a mixture of experts block, each preceded by RMSNorm and wrapped in a residual connection. Left: the three token mixers and their order: four layers [Gated DeltaNet, sliding window, Gated DeltaNet, global] followed by six blocks of five [sliding window, Gated DeltaNet, sliding window, Gated DeltaNet, global], so that the last layer is global. Right: the mixture of experts block, with 4 of the 128 routed experts active per token and one shared expert.}
  \label{fig:architecture}
\end{figure}
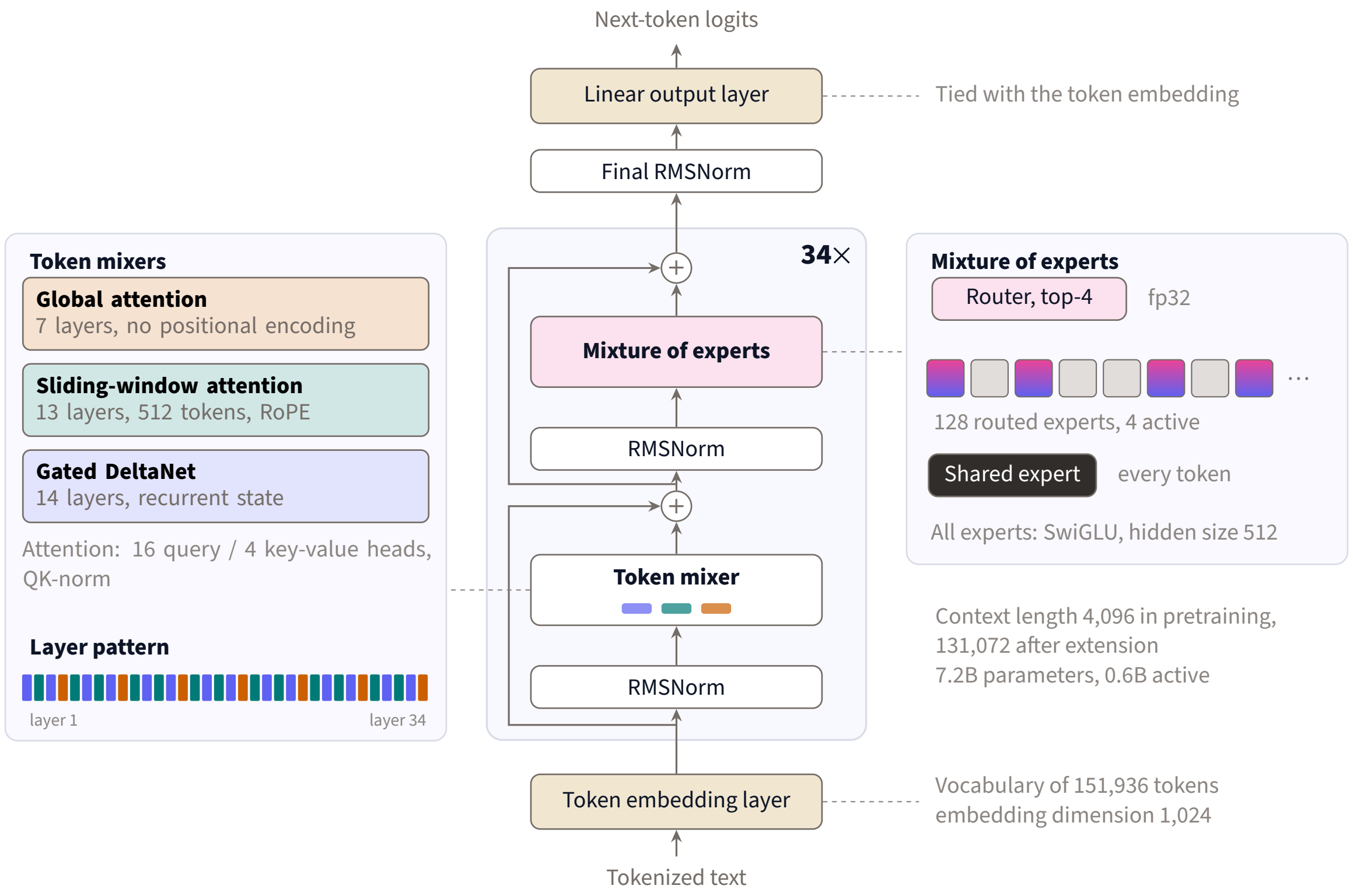

\model is a decoder-only Transformer with an MoE feed-forward block in every layer and three kinds of token mixer: Gated DeltaNet~\citep{yang2025gdn}, sliding-window attention and global attention.
It has 7.2B parameters in total, of which 0.6B are active for each token (\cref{tab:architecture}).
The routed experts hold 94\% of all parameters (6.8B).
Of the active parameters, 24\% are the tied input and output embedding, 41\% the four routed experts and the shared expert that each token passes through, and 35\% the token mixers, routers and normalization weights.
The remaining building blocks are standard: pre-normalization with RMSNorm~\citep{zhang2019rmsnorm}, SwiGLU feed-forward networks~\citep{shazeer2020glu} and no bias terms.
The model uses the Qwen3 tokenizer~\citep{yang2025qwen3}, with a vocabulary of 151{,}669 tokens (151{,}936 rows in the embedding table, padded).
\Cref{fig:architecture} shows one layer and the order of the token mixers.

\subsection{Hybrid Token Mixing}
\label{sec:arch-attention}

A Transformer with global attention in every layer caches the keys and values of all previous tokens.
Its memory per sequence therefore grows linearly with context length and dominates the cost of serving long contexts.
\model instead interleaves three token mixers with complementary strengths, in short blocks, each ending with a global layer (\cref{fig:architecture}).
A sliding-window layer is exact but sees only the last 512 tokens.
A Gated DeltaNet layer captures the full context via a compressed state but struggles with precise local comparisons~\citep{arora2024based,yang2024deltanet}.
Earlier hybrids therefore pair recurrent layers with local attention~\citep{de2024griffin,ren2024samba}.
The global layers enable exact retrieval at any distance, whereas the other two's memory does not grow with context.
Gated DeltaNet can also make training more efficient: hybrids that use it in most layers match the quality of Transformers with substantially fewer training tokens~\citep{merrill2026olmohybrid}.

\paragraph{Gated DeltaNet.}
Gated DeltaNet~\citep{yang2025gdn} is a linear-attention layer whose state is a fixed-size matrix per head, updated with the gated delta rule:
\begin{equation}
  \mathbf{S}_t = \alpha_t\, \mathbf{S}_{t-1} \left(\mathbf{I} - \beta_t\, \mathbf{k}_t \mathbf{k}_t^{\top}\right) + \beta_t\, \mathbf{v}_t \mathbf{k}_t^{\top},
  \qquad
  \mathbf{o}_t = \mathbf{S}_t\, \mathbf{q}_t .
  \label{eq:gdn}
\end{equation}
The decay $\alpha_t \in (0,1)$ and the write strength $\beta_t \in (0,1)$ depend on the input token and are computed per head.
The decay lets a head forget, and the delta rule~\citep{schlag2021linear,yang2024deltanet} overwrites the value stored under a key instead of adding to it.
Queries, keys and values pass through a causal depthwise convolution of width 4 followed by SiLU, and queries and keys are then L2-normalized.
The output of each head is normalized with RMSNorm and gated with SiLU.
The layer has no positional encoding: the recurrence itself carries the order of the tokens.

\paragraph{Sliding-window attention.}
Sliding-window attention~\citep{beltagy2020longformer,jiang2023mistral} restricts the attention of each token to the 512 most recent tokens, including itself.
These layers use rotary position embeddings with base 10{,}000 over the full head dimension.

\paragraph{Global attention.}
The global layers attend over the whole context and use no positional encoding~\citep{kazemnejad2023nope}, since the Gated DeltaNet and sliding-window layers below them already encode position.
All attention layers, windowed and global, use grouped-query attention~\citep{ainslie2023gqa} with 16 query and 4 key-value heads of dimension 128, and QK-norm.

\paragraph{Quality and memory.}
We compared the token mixers in ablations at 100B tokens against a baseline with global attention in every layer (\cref{sec:abl-hybrid}).
The hybrid achieved a lower held-out loss than the baseline, very close to that of a variant with Gated DeltaNet in all 26 non-global layers, whereas sliding windows alone did not lower it.
The hybrid also had the largest mean benchmark gain over the baseline, matching an oracle that, for each benchmark, selects the better of the Gated DeltaNet and sliding-window variants.
After a context extension to 32k tokens, the hybrid was also the best variant on RULER and ONERULER at that length, 4 to 8 points above both single-mixer variants (\cref{tab:abl-hybrid-long}).
As computed from the architecture, the memory used per sequence during generation is 468\,MiB at 32k tokens and 1.77\,GiB at 128k tokens.
A model of the same shape with global attention in every layer would hold 2.13\,GiB and 8.50\,GiB.

\subsection{Mixture of Experts Layers}
\label{sec:arch-moe}

Every layer of \model has an MoE block with 128 routed experts and 1 shared expert~\citep{dai2024deepseekmoe}, all of which are SwiGLU networks with a hidden size of 512.
A linear router computes one logit per routed expert, in fp32.
Each token is routed to its top-4 experts by router logit, and their outputs are weighted by a softmax over these 4 logits.
The shared expert processes every token and its output is added without a gate.
Routing is dropless~\citep{gale2023megablocks}: no expert has a capacity limit, and no token is dropped.
A load-balancing loss~\citep{fedus2022switch} with coefficient $10^{-3}$ pushes the router toward an even distribution of expert use.
The experts are fine-grained: their hidden size is half the model width, and a token activates $1/32$ of the routed parameters.
We chose the number of experts and this sparsity through ablations (\cref{sec:abl-architecture}).
\Cref{sec:abl-routing} analyzes how the trained model shares its experts across languages.

\section{Training}
\label{sec:pretraining}

All four stages of pretraining use a hybrid optimizer: Muon for the weight matrices of the layers and AdamW for the other parameters (\cref{sec:pt-recipe}).
In the first three stages (\cref{sec:data-mixture}), the learning rate remains constant until the anneal, then decays to zero (\cref{tab:pt-hparams}).
The fourth stage extends the context to 128k tokens with its own learning-rate schedule (\cref{sec:context-extension}).

\subsection{Pretraining Recipe}
\label{sec:pt-recipe}

\paragraph{Optimizer.}

Muon~\citep{jordan2024muon} updates the layer weight matrices, experts included, and AdamW the other parameters; all of them share one learning rate and one weight decay (\cref{tab:pt-hparams}, details in \cref{app:training}).
In our ablations, AdamW needed 1.5 to 1.9 times as many tokens as Muon to reach the same training loss (\cref{sec:abl-optimizer}).

\begin{table}[!b]
  \centering
  \small
  \caption{Architecture of \model. Parameter counts include the embedding, counted once.}
  \label{tab:architecture}
  \begin{tabular}{@{}ll@{}}
    \toprule
    \textbf{Hyperparameter} & \textbf{Value} \\
    \midrule
    Layers                                   & 34 \\
    \quad Gated DeltaNet                     & 14 \\
    \quad Sliding-window attention           & 13 (window of 512 tokens) \\
    \quad Global attention                   & 7 (no positional encoding) \\
    Hidden size                              & 1{,}024 \\
    Attention heads (query / key-value)      & 16 / 4, dimension 128 \\
    Gated DeltaNet heads (query-key / value) & 8 / 16, dimension 128 \\
    Routed experts / active per token        & 128 / 4 \\
    Shared experts                           & 1 \\
    Expert hidden size                       & 512 \\
    Active / total parameters                & 0.6B / 7.2B \\
    Vocabulary size                          & 151{,}669 (Qwen3 tokenizer; 151{,}936 embedding rows) \\
    Tied input / output embeddings           & Yes \\
    Context length (stages 1--3 / stage 4)  & 4{,}096 / 131{,}072 \\ %
    \bottomrule
  \end{tabular}
\end{table}

\paragraph{Stability measures.}
\label{sec:arch-stability}

Large logits are a common source of instability in bf16 training. \model bounds logits after each softmax.
Besides the query and key normalization~\citep{henry2020qknorm} and the fp32 router of \cref{sec:architecture}, it adds an output z-loss with coefficient $10^{-4}$~\citep{chowdhery2023palm} and a router z-loss of $10^{-3}$~\citep{zoph2022stmoe}.

\begin{table}[!t]
  \centering
  \small
  \caption{Hyperparameters of the first three stages of pretraining (6T tokens).}
  \label{tab:pt-hparams}
  \begin{tabular}{@{}ll@{}}
    \toprule
    \textbf{Hyperparameter} & \textbf{Value} \\
    \midrule
    Optimizer                       & Muon + AdamW \\
    Peak learning rate              & $8 \times 10^{-4}$ \\
    Warmup                          & 2{,}000 steps (17B tokens) \\
    Schedule                        & constant to 90\% of training, then 1$-$sqrt decay to 0 \\
    Global batch size               & 2{,}048 sequences ($\approx$8.4M tokens) \\
    Sequence length                 & 4{,}096 \\
    Training steps / tokens         & 715{,}256 / 6T \\
    Output z-loss                   & $10^{-4}$ \\
    AdamW $\beta_1$, $\beta_2$, $\epsilon$ & 0.9, 0.95, $10^{-8}$ \\
    Muon momentum                   & 0.95 (Nesterov), 5 Newton--Schulz steps \\
    Initialization std              & 0.02 \\
    Weight decay, gradient clipping & 0.1, 1.0 \\
    Precision                       & bf16 \\
    \bottomrule
  \end{tabular}
\end{table}

\paragraph{Long-context stage.}
\label{sec:context-extension}
The long-context stage uses 100B tokens (\cref{sec:data-long-context}), with the settings of \cref{tab:ce-recipe} (\cref{app:training}) taking over the last anneal optimizer state.
These 100B tokens are 1.7\% of the tokens of the first three stages, in line with the 0.8\% to 1.8\% of Olmo~3, Kolibri, SmolLM3 and Apertus~\citep{olmo2025olmo3,alephalpha2026kolibri,hf2025smollm3,apertus2025}.
The learning rate warms up to $8 \times 10^{-5}$, a tenth of the peak of the first three stages, and decays to zero over the last 15B tokens.
The position encoding is unchanged: the global attention layers use none~\citep{yang2025ropetonope}, and the sliding-window layers never attend beyond 512 tokens (\cref{sec:arch-attention}).
Unlike in the first three stages, documents packed into one sequence do not interact: attention is masked across documents and the Gated DeltaNet state restarts at each document.
Llama~3 found this masking important in continued pretraining on very long sequences~\citep{grattafiori2024llama3}.
\Cref{sec:ce-eval} evaluates the result.

\subsection{Training Dynamics}
\label{sec:pt-stability}
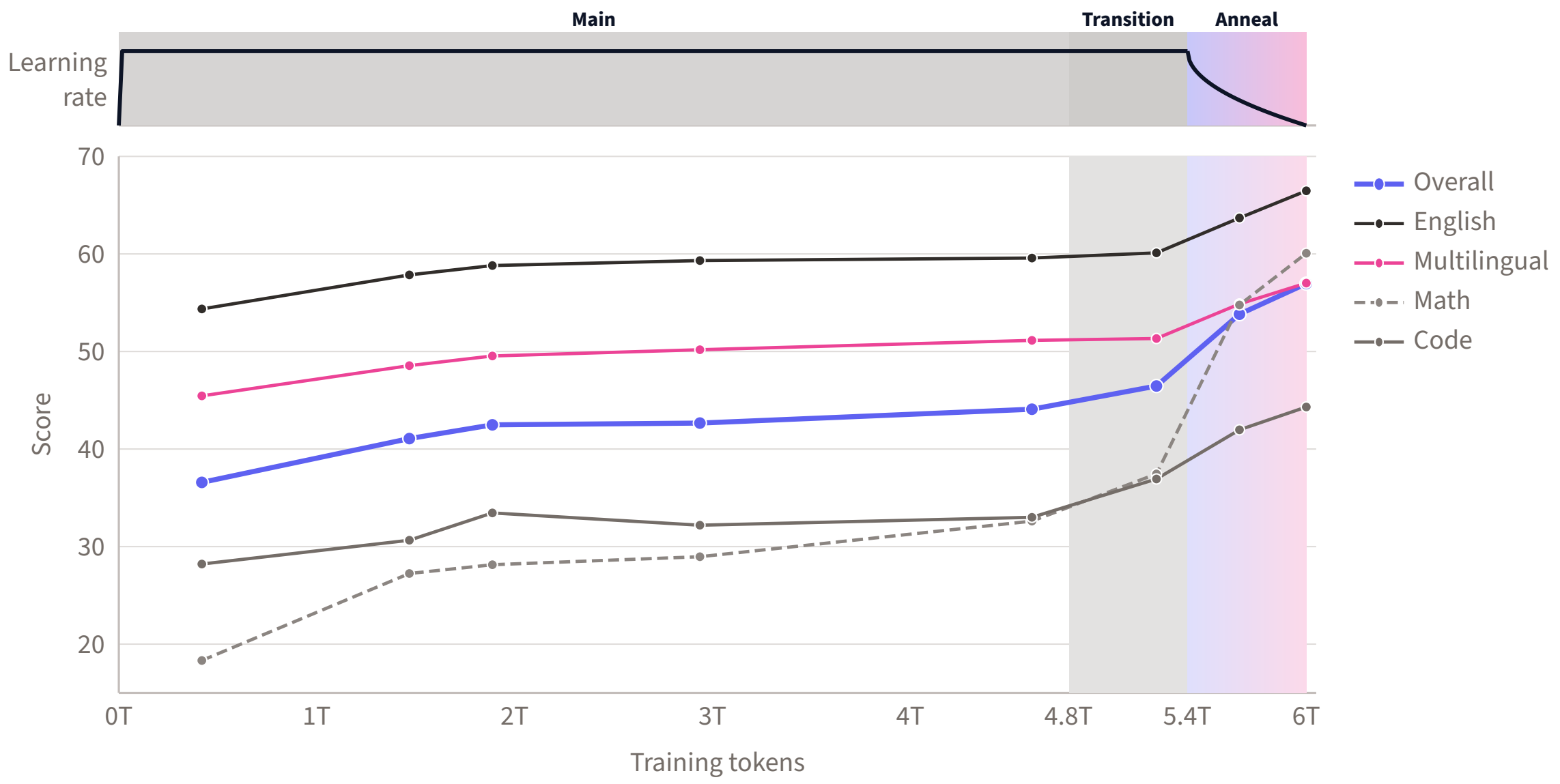
\begin{figure}[t]
  \centering
  \begin{tikzpicture}
  \begin{groupplot}[lrchart,
      group style={group size=1 by 2, vertical sep=0.3cm,
        xticklabels at=edge bottom, xlabels at=edge bottom},
      scale only axis, width=11.6cm,
      xmin=0, xmax=6.05, xtick={0,1,2,3,4,4.8,5.4,6},
      xticklabel={\pgfmathprintnumber{\tick}T},
      xlabel={Training tokens},
      table/col sep=comma,
    ]
    \nextgroupplot[height=0.9cm, ymin=0, ymax=10, axis y line=none, ytick=\empty, clip=false]
    \fill[ordinal1!35] (axis cs:0,0) rectangle (axis cs:4.8,10);
    \fill[ordinal2!35] (axis cs:4.8,0) rectangle (axis cs:5.4,10);
    \fill[gradtint=35] (axis cs:5.4,0) rectangle (axis cs:6.0,10);
    \addplot[brandink, line width=1.1pt, line join=round] coordinates {(0,0) (0.017,8) (5.4,8)};
    \addplot[brandink, line width=1.1pt, domain=5.4:6, samples=61] {8*(1-sqrt(max(x-5.4,0)/0.6))};
    \node[anchor=south, font=\sffamily\tiny\bfseries, text=brandink, inner sep=1.5pt] at (axis cs:2.4,10) {Main};
    \node[anchor=south, font=\sffamily\tiny\bfseries, text=brandink, inner sep=1.5pt] at (axis cs:5.1,10) {Transition};
    \node[anchor=south, font=\sffamily\tiny\bfseries, text=brandink, inner sep=1.5pt] at (axis cs:5.7,10) {Anneal};
    \node[anchor=east, align=right, font=\sffamily\scriptsize, text=chartsecondary, inner sep=3pt] at (axis cs:0,5) {Learning\\rate};

    \nextgroupplot[height=5.2cm, ymin=15, ymax=70, ytick={20,30,40,50,60,70}, ymajorgrids,
      yticklabel style={/pgf/number format/.cd, fixed, precision=0},
      ylabel={Score},
      legend style={at={(1.02,1)}, anchor=north west},
      legend image post style={xscale=0.8}]
    \fill[ordinal2!20] (axis cs:4.8,15) rectangle (axis cs:5.4,70);
    \fill[gradtint=20] (axis cs:5.4,15) rectangle (axis cs:6.0,70);
    \addplot[brandgradstart, line width=1.4pt, line join=round, mark=*, mark size=1.8pt,
      mark options={draw=white, line width=0.4pt, fill=brandgradstart}]
      table[x=tokens, y=overall] {figures/data/scores-over-training.csv};
    \addlegendentry{Overall}
    \addplot[domainen, line width=0.9pt, line join=round, mark=*, mark size=1.4pt,
      mark options={draw=white, line width=0.4pt, fill=domainen}]
      table[x=tokens, y=english] {figures/data/scores-over-training.csv};
    \addlegendentry{English}
    \addplot[domainml, line width=0.9pt, line join=round, mark=*, mark size=1.4pt,
      mark options={draw=white, line width=0.4pt, fill=domainml}]
      table[x=tokens, y=multilingual] {figures/data/scores-over-training.csv};
    \addlegendentry{Multilingual}
    \addplot[chartmuted, line width=0.9pt, dash pattern=on 3pt off 1.5pt, line join=round, mark=*, mark size=1.4pt,
      mark options={draw=white, line width=0.4pt, fill=chartmuted, solid}]
      table[x=tokens, y=math] {figures/data/scores-over-training.csv};
    \addlegendentry{Math}
    \addplot[domaincode, line width=0.9pt, line join=round, mark=*, mark size=1.4pt,
      mark options={draw=white, line width=0.4pt, fill=domaincode}]
      table[x=tokens, y=code] {figures/data/scores-over-training.csv};
    \addlegendentry{Code}
  \end{groupplot}
\end{tikzpicture}
  \caption{Scores of \model on the evaluation suite of \cref{sec:evaluation} at eight checkpoints of the first three stages: Overall and domain averages, with the learning-rate schedule and the training stages on top. Shading marks the transition (4.8--5.4T) and anneal (5.4--6T) stages; lines only join the evaluated checkpoints. Every domain improves fastest during the anneal.}
  \label{fig:scores-over-training}
\end{figure}

Training was stable throughout: the loss never exceeded its median over the preceding 1{,}000 steps by more than 3\%, and expert utilization remained nearly uniform (\cref{app:training}).
Because the training loss shifts with the data mixture at each stage boundary, we followed progress on a fixed held-out set and on the evaluation suite of \cref{sec:evaluation} (\cref{fig:scores-over-training}).
During the main stage, the Overall average (the mean of the four domain averages) rose gradually from 41.1 at 1.5T tokens to 44.1 at 4.6 T tokens.
The transition stage, which increases the share of code and math, mainly improved these two domains.
The anneal, which decays the learning rate on the highest-quality mixture, improved every domain.
From the last checkpoint before the anneal (5.2T tokens) to its end, the Overall average rose by 10.5 points and math by 22.6.
English and multilingual scores improved together at every evaluated checkpoint, both in the main stage and in the anneal.

\section{Evaluation}
\label{sec:evaluation}

We compare \model against 56 open base models across 26 benchmarks in four domains: English knowledge and reasoning, multilingual understanding, math and code.
The comparison tests whether a model trained on only 41\% English data remains competitive in English and leads in the other languages.
We evaluate the released \model, the checkpoint after the long-context stage (\cref{sec:context-extension}); \cref{sec:ce-eval} reports its long-context results.

\subsection{Evaluation Setup}
\label{sec:eval-setup}

\model is a base model, so we evaluate it as one, with few-shot or zero-shot prompts~\citep{brown2020gpt3,gu2024olmes} and no chat template.
We compare it only with other pretrained base models, never with instruction-tuned ones.
We score every model ourselves, with the same prompts, settings and answer extraction, using lm-evaluation-harness~\citep{biderman2024lmeval}.
Log-likelihood tasks run on the Hugging Face backend with log-softmax in fp32, and generative tasks on vLLM~\citep{kwon2023vllm}.
The two backends agree within 0.6 points on the tasks we checked.

\Cref{tab:benchmarks} in \cref{app:eval-details} lists the benchmarks and their settings.
We call the languages of the training data of \model its supported languages.
As is common for models trained on a defined set of languages~\citep{martins2024eurollm,ali2024teuken,gonzalezagirre2025salamandra}, we evaluate on the supported ones.
\Cref{app:per-language} reports the results per language.
One multilingual benchmark, T-HellaSwag, is our own translation of HellaSwag~\citep{zellers2019hellaswag} (\cref{sec:eval-t-hellaswag}).
Each domain average is the mean of its benchmarks, and the Overall average gives the four domains equal weight.
Code is scored by executing the generated programs.
HumanEval is not in the suite, because all of its items occur in the public datasets of our training pool (\cref{sec:data-decontam}).

\paragraph{Baselines.}
Following \citet{muennighoff2025olmoe}, we group the 56 baselines by active parameters, which approximate the cost of inference.
The three groups have up to 2B, from 2B to 4B and more than 4B active parameters.
The boundary between the last two groups is at 5B, so that the 4B models, with up to 4.7B active parameters, fall in the second.
Gemma-4-E2B and Gemma-4-E4B, whose 5.1B and 7.9B parameters include per-layer embeddings, are placed at their 2B and 4B effective sizes.
We checked that this grouping compares models of similar inference cost by measuring the generation speed of 45 of them on one GPU (\cref{app:inference}).
For a single user, every measured model in the two larger groups generates text 1.4 to 4.2 times more slowly than \model.
In its own group, 21 of the 28 other measured models are slower as well.
\Cref{tab:main-results} shows 27 of the 56 baselines, and \cref{app:full-results} reports all 56 models on every benchmark, including those named in the text but not in the table.

\subsection{T-HellaSwag}
\label{sec:eval-t-hellaswag}

T-HellaSwag is our higher-quality machine translation of the English commonsense reasoning benchmark HellaSwag~\citep{zellers2019hellaswag} into 110 language variants of 105 languages; we include it in our evaluation suite and will release it.
The variants are written in 25 scripts, and 46 of them use a script other than Latin.
Each variant holds 200 documents, 22{,}000 in all, selected for translation quality while keeping the difficulty of the English originals.
T-HellaSwag-full holds all 10{,}042 translated documents of each variant.
A multilingual version of HellaSwag already exists, Multilingual HellaSwag~\citep{lai2023okapi}, which covers 33 languages.
We compared the two translations across 50 documents per language using two LLM judges: Gemini~4 Argon~\citep{google2026gemini4argon} and Claude Opus~5.5~\citep{anthropic2026opus55}.
In the 32 languages they share, both judges preferred our translation in more than 98\% of the documents.
\Cref{app:t-hellaswag} describes how we translated and selected the documents, how we assessed the translation quality, and the limitations of the dataset.

\subsection{Main Results}
\label{sec:eval-results}

Among the 36 models with up to 2B active parameters (\cref{tab:main-results}), \model ranks first overall (57.0, 3.1 points above granite-4.0-h-tiny) and on the multilingual average (56.7, 3.4 points above Gemma-4-E2B).
It is also first on math, second on English and fifth on code.
Six of the 56 models score higher overall and six on the multilingual average, all in the two larger groups, with 4 to 14 times as many active parameters as \model (\cref{tab:app-full-results}).
Of these models, the four that report their pretraining tokens were trained on 18T to 36T, 3 to 6 times as many as \model.
\model also scores higher overall than several models with 6 to 13 times its active parameters, such as Ministral-3-3B, Gemma-4-E4B, Apertus-v1.5-8B and Llama-3.1-8B.
\Cref{fig:teaser} summarizes the same results.

\begin{figure}[!h]
  \centering
  \begin{tikzpicture}
  \begin{axis}[lrchart,
      width=0.92\linewidth, height=7.2cm,
      xmode=log, xmin=200, xmax=2700,
      xtick={250,500,1000,2000}, xticklabels={250,500,1{,}000,2{,}000},
      minor xtick={300,400,600,700,800,900,1500},
      xlabel={Characters generated per second for one user (log scale)},
      ymin=20, ymax=70, ytick={20,30,40,50,60,70},
      ylabel={Overall average},
      ymajorgrids, xmajorgrids,
      table/col sep=comma,
      clip=false,
      legend style={at={(0.01,0.03)}, anchor=south west, fill=white, fill opacity=0.9, text opacity=1, draw=none},
      legend image post style={scale=1.2},
    ]
    \addplot[chartsecondary, line width=0.9pt, line join=round, forget plot]
      table[x=cps, y=overall] {figures/data/pareto-speed-frontier.csv};
    \addplot[only marks, mark=*, mark size=1.6pt,
      mark options={draw=white, line width=0.4pt, fill=chartmuted},
      y filter/.expression={and(\thisrow{ours}==0, \thisrow{grp}==1) ? y : nan}]
      table[x=cps, y=overall] {figures/data/pareto-speed.csv};
    \addlegendentry{Up to 2B active}
    \addplot[only marks, mark=square*, mark size=1.5pt,
      mark options={draw=white, line width=0.4pt, fill=series2},
      y filter/.expression={\thisrow{grp}==2 ? y : nan}]
      table[x=cps, y=overall] {figures/data/pareto-speed.csv};
    \addlegendentry{2B to 4B active}
    \addplot[only marks, mark=triangle*, mark size=2.0pt,
      mark options={draw=white, line width=0.4pt, fill=series3},
      y filter/.expression={\thisrow{grp}==3 ? y : nan}]
      table[x=cps, y=overall] {figures/data/pareto-speed.csv};
    \addlegendentry{More than 4B active}
    \coordinate (oursv) at (axis cs:1088,57.0);
    \fill[white] (oursv) circle (7.5pt);
    \fill[even odd rule, left color=brandgradstart, right color=brandgradend] (oursv) circle (7.5pt) (oursv) circle (6.6pt);
    \begin{scope}
      \clip ($(oursv)+(-5pt,-5pt)$) rectangle ($(oursv)+(5pt,5pt)$);
      \node[anchor=west, inner sep=0pt] at ($(oursv)+(-5pt,0pt)$) {\TranslatedLogo[series1]{10pt}};
    \end{scope}
    \node[font=\sffamily\scriptsize\bfseries, text=brandaccent, anchor=south west, inner sep=1.5pt]
      (ourslabel) at ($(oursv)+(7pt,1pt)$) {\model};
    \node[font=\sffamily\tiny, text=chartsecondary, anchor=north west, inner sep=1.5pt]
      at (ourslabel.south west) {7.2B-A0.6B};
    \begin{scope}[every node/.style={font=\sffamily\tiny, text=chartsecondary, inner sep=1.5pt}]
      \node[anchor=south east] at (axis cs:2243,24.2) {Apertus-v1.1-0.5B};
      \node[anchor=south west] at (axis cs:1567,42.6) {Qwen3-0.6B};
      \node[anchor=south] at (axis cs:596,57.5) {MiniCPM5-2B};
      \node[anchor=west] at (axis cs:465,61.7) {Qwen3.5-4B};
      \node[anchor=south west] at (axis cs:440,63.7) {Qwen3-4B};
      \node[anchor=south] at (axis cs:281,65.9) {Qwen3-8B};
      \node[anchor=north east] at (axis cs:427,63.1) {Hunyuan-4B};
      \node[anchor=north] at (axis cs:300,62.5) {Qwen2.5-7B};
      \node[anchor=north west] at (axis cs:872,53.8) {Qwen3-1.7B};
      \node[anchor=east] at (axis cs:745,53.9) {granite-4.0-h-tiny};
      \node[anchor=east] at (axis cs:432,53.2) {Falcon-H1-1.5B-Deep};
      \node[anchor=north] at (axis cs:573,50.7) {Trinity-Nano};
      \node[anchor=east] at (axis cs:995,50.8) {LFM2.5-8B-A1B};
      \node[anchor=west] at (axis cs:298,51.2) {Llama-3.1-8B};
      \node[anchor=south] at (axis cs:383,55.4) {Gemma-4-E4B};
      \node[anchor=north] at (axis cs:655,41.9) {Gemma-4-E2B};
      \node[anchor=north west, xshift=-6pt] at (axis cs:904,48.9) {Qwen3.5-2B};
      \node[anchor=west] at (axis cs:538,52.0) {SmolLM3-3B};
      \node[anchor=east] at (axis cs:294,55.8) {Apertus-v1.5-8B};
    \end{scope}
  \end{axis}
\end{tikzpicture}
  \caption{Overall average against generation speed for a single user, in characters per second, for \model and the 45 open base models we could measure (\cref{app:inference}). Marks indicate the groups in \cref{tab:main-results}; the line is the Pareto frontier for the drawn models. Every model with a higher Overall average than \model generates text at least 1.8 times more slowly.}
  \label{fig:pareto-speed}
\end{figure}
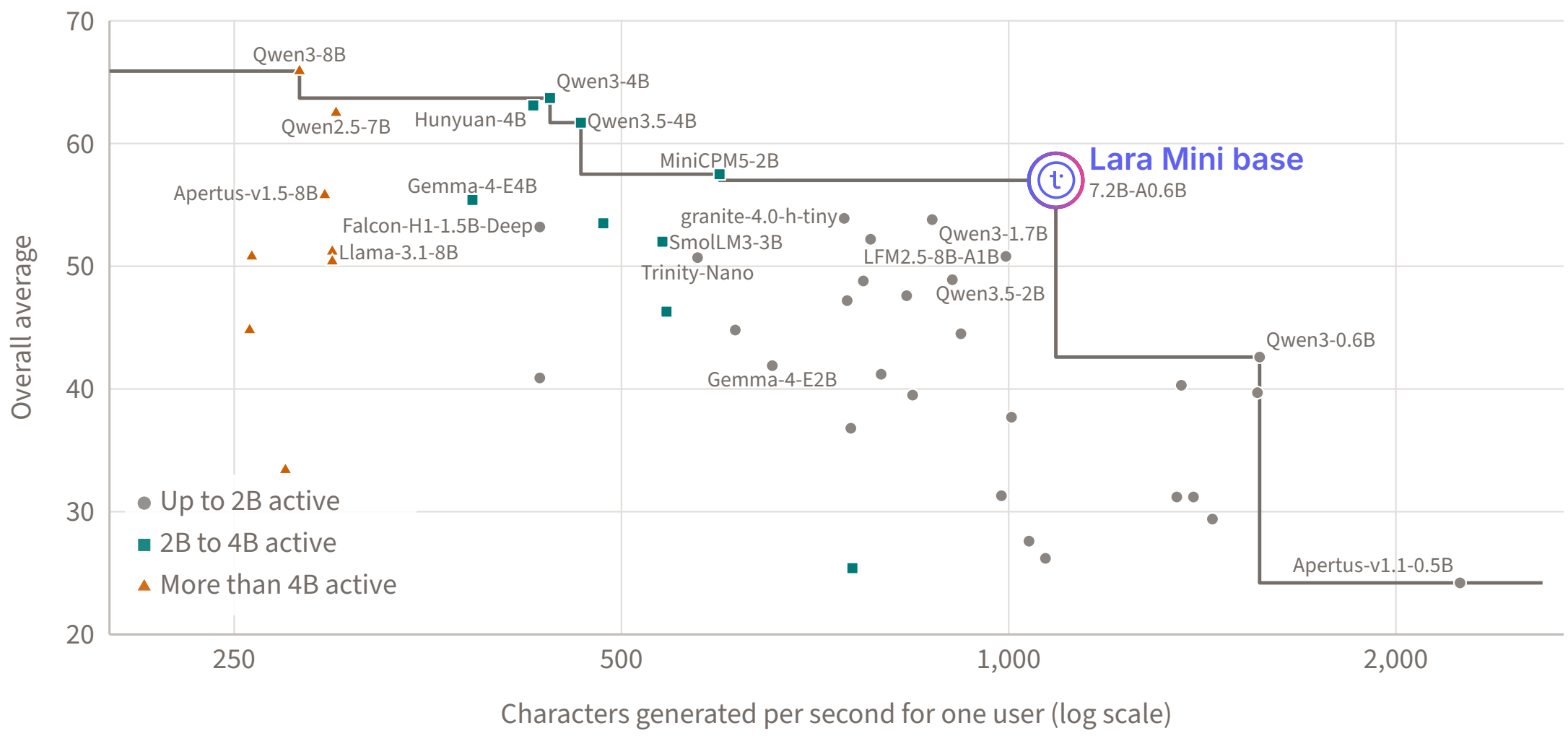

\paragraph{Efficiency.}
\Cref{fig:pareto-speed} plots the Overall average of each model against the speed at which it generates text for a single user.
Speed is measured in characters per second, so that models with tokenizers of different densities can be compared on equal footing.
\model lies on the Pareto frontier: the seven faster models score at most 42.6, and every model that scores higher generates text at least 1.8 times more slowly.
The closest of them in Overall average, MiniCPM5-2B, is 0.5 points higher but 8.7 points lower on the multilingual average.
The speed advantage of \model is largest for a single user and on long prompts, and narrows when many users share the GPU (\cref{app:inference}).

\begin{table}[!t]
  \centering
  \small
  \caption{Domain averages of \model and open base models, grouped by active parameters as in \citet{muennighoff2025olmoe}: up to 2B, from 2B to 4B and more than 4B. Active, Total: active and total parameters. Tokens: pretraining tokens reported by each developer, over all pretraining stages (--: not reported; $^\dagger$: distilled from a larger model during pretraining, wholly or in part); for \model, the 6T of the first three stages (\cref{sec:data}), as in \cref{tab:ml-share}; the long-context stage adds 100B. Multilingual is averaged over the languages supported by \model; Overall is the mean of the four domain averages. Best result in each group in bold, second best underlined. Gemma-4-E2B and Gemma-4-E4B are listed at their 2B and 4B effective sizes, and their totals include per-layer embeddings. All 56 models we evaluated are in \cref{tab:app-full-results}. \model has the highest Overall, multilingual and math averages of the models with up to 2B active parameters.}
  \label{tab:main-results}
  \setlength{\tabcolsep}{5pt} %
  \begin{tabular}{@{}lrrrrrrrr@{}}
    \toprule
    & \multicolumn{2}{c}{\textbf{Parameters}} & & \multicolumn{4}{c}{\textbf{Domain average}} & \\
    \cmidrule(lr){2-3}\cmidrule(lr){5-8}
    \textbf{Model} & \textbf{Active} & \textbf{Total} & \textbf{Tokens} & \textbf{English} & \textbf{Multilingual} & \textbf{Math} & \textbf{Code} & \textbf{Overall} \\
    \midrule
    \multicolumn{9}{@{}l}{\textit{Models with more than 4B active parameters}} \\
    Llama-3.1-8B & 8.0B & 8.0B & 15T & 69.6 & 55.2 & 36.6 & 43.6 & 51.2 \\
    Apertus-v1.5-8B & 8.0B & 8.0B & 19T & 69.0 & \textbf{59.2} & 50.6 & 44.3 & 55.8 \\
    Qwen2.5-7B & 7.6B & 7.6B & 18T & \textbf{72.3} & \underline{57.3} & \underline{64.9} & \underline{55.5} & \underline{62.5} \\
    Qwen3-8B & 8.2B & 8.2B & 36T & \underline{71.9} & \textbf{59.2} & \textbf{71.3} & \textbf{61.0} & \textbf{65.9} \\
    \midrule
    \multicolumn{9}{@{}l}{\textit{Models with 2B to 4B active parameters}} \\
    SmolLM3-3B & 3.1B & 3.1B & 11.2T & 64.7 & 48.7 & 51.0 & 43.6 & 52.0 \\
    Nemotron-Flash-3B & 2.7B & 2.7B & 4.5T & 65.6 & 42.0 & 59.7 & 44.5 & 52.9 \\
    Ministral-3-3B & 3.4B & 3.4B & 1--3T$^\dagger$ & 67.7 & 55.4 & 49.9 & 42.3 & 53.8 \\
    Domyn-Edge & 3.9B & 3.9B & 13T & 68.2 & 52.0 & 54.1 & 47.3 & 55.4 \\
    Gemma-4-E4B & 4.0B & 7.9B & -- & 70.1 & \textbf{61.1} & 44.4 & 46.0 & 55.4 \\
    MiniCPM5-2B & 2.5B & 2.5B & -- & 65.3 & 48.0 & 66.0 & 50.8 & 57.5 \\
    Qwen3.5-4B & 4.7B & 4.7B & -- & \underline{70.5} & \underline{58.0} & 66.1 & 52.3 & 61.7 \\
    Hunyuan-4B & 4.2B & 4.2B & -- & 70.4 & 51.6 & \textbf{72.0} & \textbf{58.6} & \underline{63.1} \\
    Qwen3-4B & 4.0B & 4.0B & 36T & \textbf{71.2} & 57.5 & \underline{69.2} & \underline{57.0} & \textbf{63.7} \\
    \midrule
    \multicolumn{9}{@{}l}{\textit{Models with up to 2B active parameters}} \\
    Llama-3.2-1B & 1.2B & 1.2B & 9T$^\dagger$ & 51.9 & 39.7 & 5.0 & 28.0 & 31.2 \\
    Apertus-v1.1-1.5B & 1.5B & 1.5B & 1.7T$^\dagger$ & 53.6 & 44.8 & 5.9 & 20.6 & 31.2 \\
    OLMoE-1B-7B & 1.3B & 6.9B & 5.1T & 61.9 & 40.4 & 27.2 & 21.1 & 37.7 \\
    Gemma-4-E2B & 2.0B & 5.1B & -- & 63.3 & \underline{53.3} & 18.0 & 32.8 & 41.9 \\
    Qwen3-0.6B & 0.6B & 0.6B & 36T & 55.3 & 43.1 & 37.2 & 34.8 & 42.6 \\
    Qwen2.5-1.5B & 1.5B & 1.5B & 18T & 62.6 & 46.8 & 43.0 & 38.1 & 47.6 \\
    Qwen3.5-2B & 1.9B & 1.9B & -- & 63.3 & 50.0 & 45.6 & 36.7 & 48.9 \\
    LFM3-1.7B & 1.7B & 1.7B & 22T & 60.0 & 46.7 & 52.2 & 38.0 & 49.2 \\
    Trinity-Nano & 1.0B & 6.1B & 10T & \textbf{68.4} & 47.4 & 45.0 & 42.0 & 50.7 \\
    LFM2.5-8B-A1B & 1.7B & 8.5B & 38T$^\dagger$ & 65.7 & 48.6 & 52.3 & 36.6 & 50.8 \\
    Hunyuan-1.8B & 1.8B & 1.8B & -- & 62.7 & 44.3 & 52.0 & \textbf{49.7} & 52.2 \\
    Falcon-H1-1.5B-Deep & 1.6B & 1.6B & 3T & 65.3 & 45.2 & \underline{55.2} & 47.2 & 53.2 \\
    Qwen3-1.7B & 1.7B & 1.7B & 36T & 63.8 & 50.3 & 54.6 & 46.6 & 53.8 \\
    granite-4.0-h-tiny & 1.0B & 6.9B & 22.5T & 66.2 & 51.8 & 49.4 & \underline{48.4} & \underline{53.9} \\
    \ourrow \model & 0.6B & 7.2B & 6T & \underline{66.4} & \textbf{56.7} & \textbf{59.9} & 45.1 & \textbf{57.0} \\
    \bottomrule
  \end{tabular}
\end{table}

\subsection{English}
\label{sec:eval-english}

\Cref{tab:eval-domains} reports every benchmark for \model, the best models of its group and the best model of each larger group, and \cref{tab:app-english} lists the English benchmarks for all models.
Its English average (66.4) is the second best of its group, 2.0 points below Trinity-Nano (68.4).
Its English average lies between those of the 3B dense models SmolLM3-3B (64.7) and Ministral-3-3B (67.7).

\begin{table}[!h]
  \centering
  \footnotesize
  \setlength{\tabcolsep}{3.4pt}
  \newcommand{\hd}[1]{\textbf{\begin{tabular}[b]{@{}c@{}}#1\end{tabular}}}
  \caption{Per-benchmark results for \model, the five other models with up to 2B active parameters that score highest on the Overall average, and the best model of each larger group (\cref{tab:main-results}). Multilingual benchmarks are averaged over the languages supported by \model. Best result among the six models with up to 2B active parameters in bold, second best underlined. NP, P: non-parallel and parallel. Settings in \cref{tab:benchmarks}; all models in \cref{tab:app-english,tab:app-multilingual,tab:app-math-code}.}
  \label{tab:eval-domains}
  \begin{tabular}{@{}lrrrrr>{\columncolor{brandaccent!18}}r>{\color{chartsecondary}}r>{\color{chartsecondary}}r@{}}
    \toprule
    & \multicolumn{6}{c}{\textbf{Up to 2B active parameters}} & \multicolumn{2}{c@{}}{\textbf{Larger models}} \\
    \cmidrule(lr){2-7}\cmidrule(l){8-9}
    \textbf{Benchmark} & \hd{LFM2.5\\8B-A1B} & \hd{Hunyuan\\1.8B} & \hd{Falcon-H1\\1.5B-Deep} & \hd{Qwen3\\1.7B} & \hd{granite-4.0\\h-tiny} & \multicolumn{1}{>{\columncolor{brandaccent!18}}c}{\hd{\modelfont{Lara Mini}\\\modelfont{base}}} & \hd{Qwen3\\4B\\{\normalfont\itshape 2--4B}} & \hd{Qwen3\\8B\\{\normalfont\itshape $>$4B}} \\
    \midrule
    \multicolumn{9}{@{}l}{\textit{English}} \\
    NLI & 52.3 & 50.7 & \textbf{55.9} & \underline{54.2} & \underline{54.2} & 52.7 & 54.4 & 46.8 \\
    PAWS & 62.9 & \textbf{71.0} & \underline{68.0} & 61.2 & 63.3 & 65.3 & 68.6 & 55.5 \\
    Belebele & 69.1 & 75.4 & 70.9 & \underline{75.9} & 69.9 & \textbf{76.0} & 89.3 & 91.3 \\
    COPA & \textbf{85.4} & 78.6 & 82.2 & 82.4 & \textbf{85.4} & \underline{84.8} & 85.8 & 89.0 \\
    PIQA & 77.7 & 72.9 & 76.4 & 75.5 & \underline{79.5} & \textbf{80.0} & 78.0 & 79.5 \\
    HellaSwag & 74.0 & 62.2 & 69.8 & 67.2 & \textbf{77.7} & \underline{74.8} & 75.7 & 79.6 \\
    WinoGrande & 67.6 & 66.1 & 67.3 & 65.4 & \textbf{73.5} & \underline{69.6} & 72.0 & 76.6 \\
    ARC-C & \textbf{64.9} & 50.1 & 55.7 & 55.7 & \underline{61.7} & 58.6 & 64.4 & 67.6 \\
    MMLU & 63.2 & 62.9 & \textbf{66.6} & 62.6 & \underline{65.5} & 63.3 & 73.0 & 76.8 \\
    MMLU-Pro & \underline{39.8} & 37.5 & \textbf{40.1} & 37.4 & 31.6 & 39.0 & 51.2 & 56.1 \\
    \cmidrule(l){1-9}
    English average & 65.7 & 62.7 & 65.3 & 63.8 & \underline{66.2} & \textbf{66.4} & 71.2 & 71.9 \\
    \midrule
    \multicolumn{9}{@{}l}{\textit{Multilingual}} \\
    XNLI & 41.1 & 38.3 & 40.4 & 42.0 & \underline{42.8} & \textbf{44.9} & 43.3 & 39.1 \\
    PAWS-X & 54.4 & \textbf{58.5} & 56.8 & 55.7 & \underline{57.0} & 56.9 & 57.7 & 51.3 \\
    Belebele & 40.5 & 45.1 & 40.3 & \textbf{61.8} & 51.1 & \underline{59.3} & 78.5 & 82.3 \\
    XCOPA & \underline{62.4} & 55.3 & 54.8 & 62.2 & 61.5 & \textbf{68.1} & 66.6 & 69.4 \\
    Global PIQA (NP) & 61.8 & 54.5 & 56.8 & 58.1 & \underline{63.1} & \textbf{72.6} & 63.0 & 67.1 \\
    Global PIQA (P) & 29.0 & 23.5 & 26.5 & 28.4 & \underline{30.3} & \textbf{34.4} & 31.1 & 33.2 \\
    T-HellaSwag & 46.6 & 35.1 & 39.3 & 42.0 & \underline{51.3} & \textbf{58.4} & 51.2 & 56.8 \\
    INCLUDE & 48.2 & 43.4 & 42.5 & 51.4 & \underline{53.4} & \textbf{58.7} & 62.6 & 66.8 \\
    Global MMLU & 53.7 & 44.6 & 49.1 & 50.8 & \underline{55.4} & \textbf{56.5} & 63.2 & 67.2 \\
    \cmidrule(l){1-9}
    Multilingual average & 48.6 & 44.3 & 45.2 & 50.3 & \underline{51.8} & \textbf{56.7} & 57.5 & 59.2 \\
    \midrule
    \multicolumn{9}{@{}l}{\textit{Math}} \\
    GSM8K & 67.6 & 71.6 & \textbf{72.3} & \underline{71.8} & 59.1 & 67.2 & 87.8 & 86.7 \\
    MATH & 45.8 & 42.3 & \underline{49.2} & 41.0 & 45.4 & \textbf{60.9} & 52.8 & 55.3 \\
    MGSM & 43.3 & 42.1 & 44.3 & \underline{50.9} & 43.6 & \textbf{51.7} & 66.9 & 71.9 \\
    \cmidrule(l){1-9}
    Math average & 52.3 & 52.0 & \underline{55.2} & 54.6 & 49.4 & \textbf{59.9} & 69.2 & 71.3 \\
    \midrule
    \multicolumn{9}{@{}l}{\textit{Code}} \\
    MBPP & 40.4 & \textbf{58.8} & 56.8 & 55.4 & \underline{58.4} & 53.8 & 67.0 & 69.6 \\
    MBPP+ & 48.4 & \textbf{62.2} & 58.5 & \underline{59.8} & 58.5 & 58.2 & 68.0 & 70.4 \\
    CRUXEval-I & 28.4 & \textbf{39.6} & 35.8 & 35.9 & \underline{36.6} & 34.4 & 45.1 & 52.4 \\
    CRUXEval-O & 29.1 & \underline{38.4} & 37.8 & 35.1 & \textbf{40.0} & 33.9 & 48.0 & 51.6 \\
    \cmidrule(l){1-9}
    Code average & 36.6 & \textbf{49.7} & 47.2 & 46.6 & \underline{48.4} & 45.1 & 57.0 & 61.0 \\
    \bottomrule
  \end{tabular}
\end{table}

\subsection{Multilingual}
\label{sec:eval-multilingual}

\Cref{tab:eval-domains} reports the multilingual benchmarks on the languages supported by \model, and \cref{tab:app-multilingual} lists them for all models.
\model ranks first in its group on seven of the nine benchmarks, and on XCOPA is tied with Gemma-4-E2B.
Its lead is largest on INCLUDE (58.7, 5.3 points above granite-4.0-h-tiny).
It ranks lower on the other two benchmarks, Belebele and PAWS-X.
On Belebele, Qwen3.5-2B and Qwen3-1.7B score up to 2.6 points higher.
On PAWS-X, a paraphrase task in six high-resource languages, \model is 1.6 points behind the best models of its group, Hunyuan-1.8B and Trinity-Nano.
On T-HellaSwag, \model is first in its group on its supported languages (58.4), and only three of the 56 models, all with more active parameters, score higher.
Averaged over all the languages of the benchmarks, including those it does not support, \model still ranks first in its group, at 49.4 (\cref{tab:app-multilingual-all}).
The second, Gemma-4-E2B (49.0), supports more than 140 languages~\citep{gemmateam2026gemma4}, against the 45 of \model.

\subsection{Math and Code}
\label{sec:eval-math-code}

\Cref{tab:eval-domains} also reports math and code.
\model has the best math average in its group, 4.7 points above Falcon-H1-1.5B-Deep.
Its math average is also higher than that of every model with more active parameters except MiniCPM5-2B and five Qwen and Hunyuan models with 4B to 8B active parameters.
Its code average ranks fifth in its group.

\subsection{Long-Context Results}
\label{sec:ce-eval}
We evaluate \model after the long-context stage (\cref{sec:context-extension}) at 32k and 128k tokens on RULER~\citep{hsieh2024ruler} (13 synthetic English tasks), ONERULER~\citep{kim2025oneruler} (retrieval tasks in 26 languages) and HELMET~\citep{yen2025helmet} (mostly application-centric English tasks).
Since \model is a base model, it continues generating text after its answer.
We therefore keep only the tasks whose scores do not depend on that text: four retrieval tasks for ONERULER and four categories for HELMET (\cref{app:lc-eval}).

Before the long-context stage, \model scores at most 15.0 on the three benchmarks at 32k and 128k tokens (\cref{tab:ce-results}).
After it, \model leads the five open models of \cref{tab:ce-results} on ONERULER, the only multilingual benchmark of the three, at 32k tokens (90.1), and is second at 128k, 3.8 points behind Gemma-4-E2B.
On RULER and HELMET, it trails only Qwen3.5-2B and Gemma-4-E2B, which have about three times its active parameters.
The long-context stage barely changes short-context scores: on the benchmarks in \cref{tab:main-results}, the Overall average changes by less than 0.1 points, and no domain average by more than 0.8 points.

\begin{table}[!h]
  \centering
  \small
  \caption{Long-context scores of \model at the end of the anneal and after the long-context stage (the released model), and of open base models, all scored with the same protocol (\cref{app:lc-eval}): mean of the 13 RULER tasks, of four ONERULER retrieval tasks in 26 languages and of four HELMET categories. Open models: five models with up to 2B active parameters that support 128k tokens, including the strongest of the group overall (granite-4.0-h-tiny) and on the multilingual average (Gemma-4-E2B). Active, Total: active and total parameters. Gemma-4-E2B at its 2B effective size; best in bold, second underlined. After the long-context stage, \model ranks first on ONERULER at 32k tokens and second at 128k tokens.}
  \label{tab:ce-results}
  \begin{tabular}{@{}lrrrrrrrr@{}}
    \toprule
                          & \multicolumn{2}{c}{\textbf{Parameters}} & \multicolumn{2}{c}{\textbf{RULER}} & \multicolumn{2}{c}{\textbf{ONERULER}} & \multicolumn{2}{c@{}}{\textbf{HELMET}} \\
    \cmidrule(lr){2-3} \cmidrule(lr){4-5} \cmidrule(lr){6-7} \cmidrule(l){8-9}
    \textbf{Model}       & \textbf{Active} & \textbf{Total} & \textbf{32k} & \textbf{128k} & \textbf{32k} & \textbf{128k} & \textbf{32k} & \textbf{128k} \\
    \midrule
    End of the anneal     & 0.6B & 7.2B & 15.0 & 10.9 & 0.4 & 0.0 & 9.3 & 8.9 \\
    \ourrow \model           & 0.6B & 7.2B & 76.6 & 63.7 & \textbf{90.1} & \underline{80.9} & 63.0 & 54.7 \\
    \midrule
    Trinity-Nano          & 1.0B  & 6.1B  & 71.0 & 61.8 & 73.9 & 62.1 & 54.9 & 51.8 \\
    granite-4.0-h-tiny    & 1.0B  & 6.9B  & 29.5 & 26.4 & 19.5 & 14.1 & 41.0 & 36.3 \\
    Falcon-H1-1.5B-Deep   & 1.6B  & 1.6B  & 59.7 & 38.7 & 60.5 & 38.8 & 45.5 & 34.4 \\
    Qwen3.5-2B            & 1.9B  & 1.9B  & \textbf{83.7} & \textbf{73.8} & 87.7 & 77.4 & \underline{64.0} & \underline{56.8} \\
    Gemma-4-E2B           & 2.0B  & 5.1B  & \underline{79.2} & \underline{71.0} & \underline{89.3} & \textbf{84.7} & \textbf{66.0} & \textbf{58.6} \\
    \bottomrule
  \end{tabular}
\end{table}

\section{Multilingual Data and Model Capacity}
\label{sec:ml-analysis}
The introduction raised a common concern about multilingual pretraining: at small scales, languages compete for the model's capacity~\citep {conneau2020xlmr,chang2024multilinguality}.
If they did in \model, two signs should be visible.
First, its multilingual data would cost English: replacing them with English would clearly raise English scores.
Second, its mixture of experts would serve the languages by dividing its experts among them, so that each language uses only part of the model.

\paragraph{Development suite.}
We scored the anneals of \cref{sec:abl-multilingual} and the short runs of \cref{sec:abl-architecture,sec:abl-hybrid} on a development suite.
Its 17 benchmarks are the English and multilingual benchmarks in \cref{tab:benchmarks}, excluding ARC-Challenge and MMLU-Pro.
MMLU-Pro remained at chance level (10\%) in the short runs trained from scratch, and we exclude it from all development scores so that they all average to the same benchmarks.
The development suite uses its own prompts and, unless otherwise stated, averages across all languages within each benchmark, so its scores are not comparable to those in \cref{sec:evaluation}.

\subsection{Multilingual Share}
\label{sec:abl-multilingual}

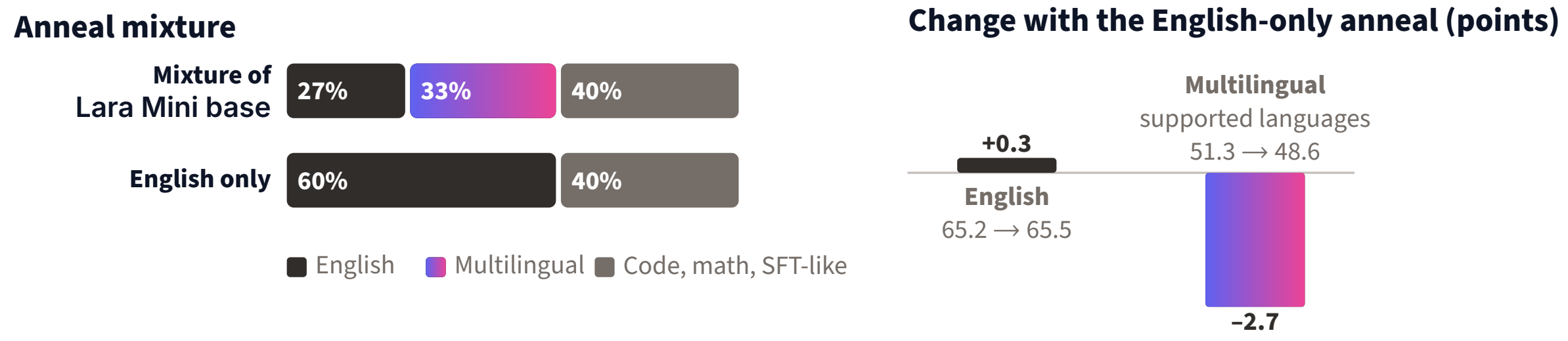
\begin{figure}[!h]
  \centering
  \begin{tikzpicture}[font=\sffamily]
  \def\lrbx{2.75}   %
  \def\lrbw{4.6}    %
  \def\lrbh{0.55}   %
  \def\lrgap{0.05}
  \node[anchor=south west, inner sep=0pt, font=\sffamily\small\bfseries, text=brandink] at (0, 1.25) {Anneal mixture};
  \def\lrmixbar#1#2{%
    \gdef\lrcursor{\lrbx}%
    \foreach \s/\f/\t in {#2} {%
      \pgfmathsetmacro\lrw{\s/100*\lrbw}%
      \pgfmathsetmacro\lrend{\lrcursor+\lrw-\lrgap}%
      \fill[\f, rounded corners=2pt] (\lrcursor, #1) rectangle (\lrend, #1-\lrbh);
      \node[font=\sffamily\scriptsize\bfseries, text=\t, anchor=west, inner sep=0pt]
        at (\lrcursor+0.1, #1-\lrbh/2) {\s\%};
      \pgfmathsetmacro\lrnext{\lrend+\lrgap}%
      \xdef\lrcursor{\lrnext}%
    }}
  \node[anchor=east, inner sep=0pt, font=\sffamily\scriptsize\bfseries, text=brandink, align=right]
    at (\lrbx-0.15, 0.725) {Mixture of\\\model};
  \node[anchor=east, inner sep=0pt, font=\sffamily\scriptsize\bfseries, text=brandink, align=right]
    at (\lrbx-0.15, -0.175) {English only};
  \lrmixbar{1.0}{27/domen/white, 33/domml/white, 40/domcode/white}
  \lrmixbar{0.1}{60/domen/white, 40/domcode/white}
  \foreach \x/\c/\lab in {0/domen/English, 1.4/domml/Multilingual, 3.1/domcode/{Code, math, SFT-like}} {
    \fill[\c, rounded corners=1pt] (\lrbx+\x, -0.95) rectangle ++(0.2, -0.2);
    \node[anchor=west, inner sep=0pt, font=\sffamily\scriptsize, text=chartsecondary] at (\lrbx+\x+0.28, -1.05) {\lab};
  }

  \def\lrzy{-0.1}   %
  \def\lrsc{0.5}    %
  \def\lrdw{1.0}    %
  \node[anchor=south west, inner sep=0pt, font=\sffamily\small\bfseries, text=brandink] at (9.0, 1.25)
    {Change with the English-only anneal (points)};
  \draw[chartaxis, line width=0.6pt] (9.0, \lrzy) -- (13.5, \lrzy);
  \fill[domen, rounded corners=1pt] (9.5, \lrzy) rectangle (9.5+\lrdw, \lrzy+0.3*\lrsc);
  \node[anchor=south, inner sep=1.5pt, font=\sffamily\scriptsize\bfseries, text=chartink] at (10.0, \lrzy+0.3*\lrsc) {+0.3};
  \node[anchor=north, inner sep=0pt, font=\sffamily\scriptsize, text=chartsecondary, align=center] at (10.0, \lrzy-0.12)
    {\textbf{English}\\65.2 $\rightarrow$ 65.5};
  \fill[domml, rounded corners=1pt] (12.0, \lrzy) rectangle (12.0+\lrdw, \lrzy-2.7*\lrsc);
  \node[anchor=north, inner sep=1.5pt, font=\sffamily\scriptsize\bfseries, text=chartink] at (12.5, \lrzy-2.7*\lrsc) {\textendash 2.7};
  \node[anchor=south, inner sep=0pt, font=\sffamily\scriptsize, text=chartsecondary, align=center] at (12.5, \lrzy+0.12)
    {\textbf{Multilingual}\\supported languages\\51.3 $\rightarrow$ 48.6};
\end{tikzpicture}
  \caption{Multilingual share of the anneal. Left: mixtures of the two anneals. Right: change in the mean score on the development suite when the multilingual data are replaced with English, with the scores of both anneals under each bar. English averages eight benchmarks; multilingual averages nine across the languages supported by \model. Replacing the multilingual data with English barely raises the English average and clearly lowers the multilingual one.}
  \label{fig:abl-multilingual}
\end{figure}

To test the first sign, we branched two anneals of 100B tokens from the training run of \model at the end of its transition stage (5.4T tokens).
Both decayed the learning rate to zero and differed only in their mixture.
The first used the anneal mixture of \model, with 27\% English, 33\% multilingual and 40\% code, math and SFT-like data.
The second replaced all multilingual data with English, at 60\% English and the same 40\% of code, math and SFT-like data.
We scored both on the development suite, averaging the multilingual benchmarks over the supported languages as in \cref{sec:evaluation} (\cref{fig:abl-multilingual}).

With more than twice as much English data in the anneal, the English average rose by only 0.3 points, about a tenth of the 2.7 points that the multilingual average lost on the supported languages (\cref{fig:abl-multilingual}).
Without multilingual data, the score was lower on eight of the nine multilingual benchmarks, by up to 4.6 points on commonsense and knowledge benchmarks such as Global PIQA, XCOPA, T-HellaSwag and INCLUDE.
The multilingual data cost math and code less than half a point.
On the math and code benchmarks of \cref{sec:evaluation}, the anneal with multilingual data scored 53.8 and 43.6, and the one without it 54.2 and 43.7.
A likely reason for the small English gain is that, by the end of the transition stage, the model had already read more than 2T English tokens, so that additional English in the anneal added little.
The test covers only the anneal: it does not measure the effect of the multilingual share during the main stage.

\begin{takeaway}
In the anneal, replacing the multilingual third of the mixture with English raises the English average by 0.3 points and lowers the multilingual average by 2.7 points.
\end{takeaway}

\subsection{Expert Routing across Languages}
\label{sec:abl-routing}
\begin{figure}[!h]
  \centering
  \begin{tikzpicture}
  \begin{axis}[lrchart,
      width=0.75\linewidth, height=5.0cm,
      xmin=0.5, xmax=34.5, xtick={1,5,10,15,20,25,30,34},
      ymin=0, ymax=0.4, ytick={0,0.1,0.2,0.3,0.4},
      yticklabel style={/pgf/number format/.cd, fixed, fixed zerofill, precision=1},
      ymajorgrids,
      xlabel={Layer},
      ylabel={Divergence from mean routing (bits)},
      table/col sep=comma,
      legend style={at={(0.5,1.03)}, anchor=south, legend columns=3,
        /tikz/every even column/.append style={column sep=6pt}},
    ]
    \fill[chartsecondary!12] (axis cs:9.5,0) rectangle (axis cs:17.5,0.4);
    \node[anchor=north, font=\sffamily\scriptsize, text=chartsecondary, inner sep=2pt] at (axis cs:13.5,0.39) {shared};
    \addplot[series1, line width=1.0pt, line join=round, mark=*, mark size=1.1pt]
      table[x=layer, y=jsd_mean] {figures/data/expert-routing.csv};
    \addlegendentry{Mean of the supported languages}
    \addplot[series1, line width=0.4pt, dashed, forget plot]
      table[x=layer, y=jsd_min] {figures/data/expert-routing.csv};
    \addplot[series1, line width=0.4pt, dashed]
      table[x=layer, y=jsd_max] {figures/data/expert-routing.csv};
    \addlegendentry{Range over languages}
    \addplot[chartink, line width=1.0pt, line join=round]
      table[x=layer, y=jsd_english] {figures/data/expert-routing.csv};
    \addlegendentry{English}
  \end{axis}
\end{tikzpicture}
  \caption{Expert routing by language in the released model: for each layer, the Jensen--Shannon divergence between a language's distribution over the 128 routed experts and the mean over the 43 supported languages covered by Belebele, on the same text in every language. Languages route differently in the first and last layers and almost identically in the middle layers.}
  \label{fig:expert-routing}
\end{figure}
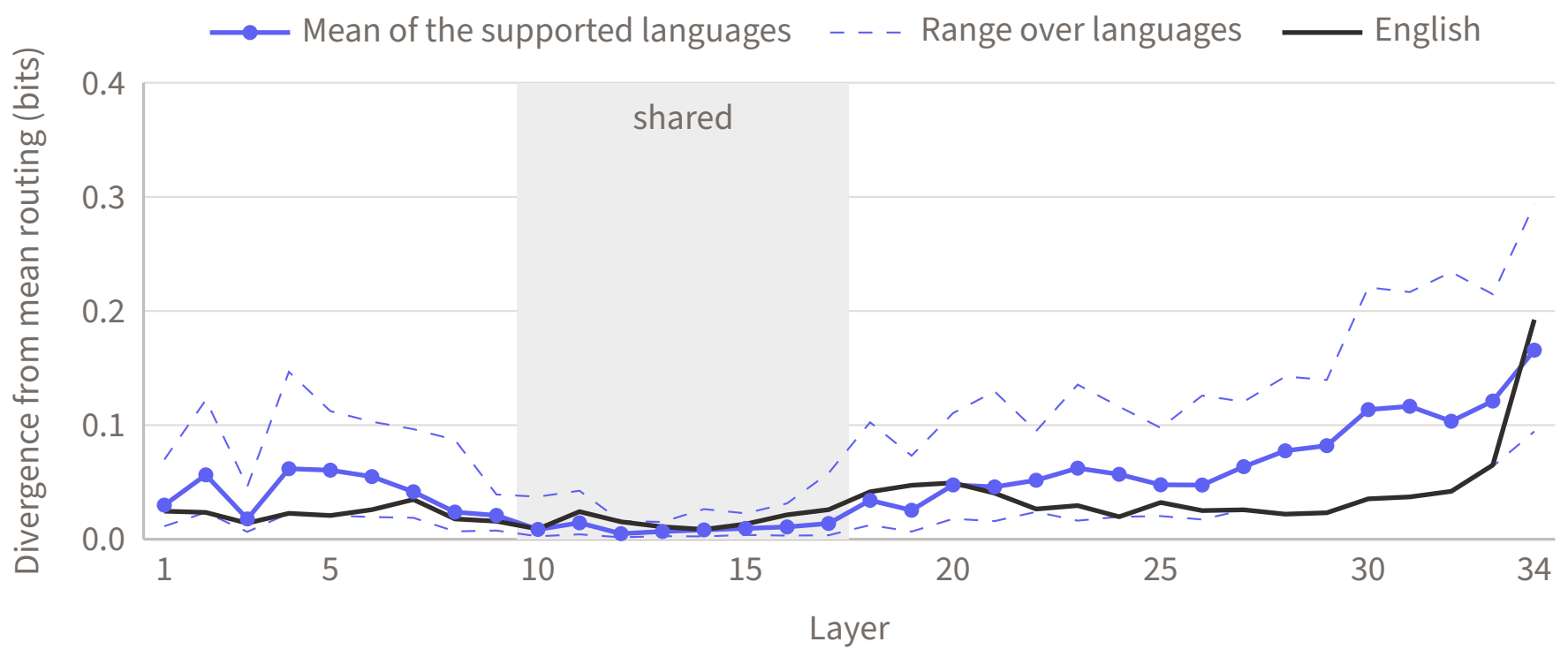

To test the second sign, we ran the released \model on the Belebele passages~\citep {bandarkar2024belebele}, the same text in each supported language, and recorded the experts to which each token was routed.
In each layer, we measured the Jensen--Shannon divergence between each language's distribution over the routed experts and the mean distribution across all 43 languages (\cref{fig:expert-routing}).

Routing differs between languages in the first and last layers and is almost identical in the middle of the network, with an average divergence of at most 0.015 bits in layers 10 to 17.
In layers 2 to 7, 31 of the 37 languages with a relative among the 43 route most like that relative, for example, Catalan like Spanish, Czech like Slovak, and Estonian like Finnish.
The grouping follows the language rather than the script: Serbian, written in Cyrillic, routes most like Croatian, and Maltese, written in Latin script, most like Arabic.
Languages, therefore, share experts rather than divide them, which is consistent with the small cost of multilingual data for English (\cref{sec:abl-multilingual}).
Together, the two analyses find little of the competition that the concern predicts, but they do not show whether the mixture of experts or the data selection is responsible.

These findings are consistent with the hypothesis that multilingual learning benefits from shared representations rather than requiring separate capacity for each language.
This resembles the positive transfer observed in multi-task learning~\citep{caruana1997multitask} and multilingual neural machine translation~\citep{firat2016multiway}.
One possible explanation is that different languages provide complementary constraints on representations of underlying concepts, allowing the model to generalize across linguistic boundaries.
However, shared expert routing does not directly establish shared semantic representations, and our experiments do not demonstrate that multilingual training improves general reasoning or English performance.

\begin{takeaway}
\model does not divide its experts among languages: related languages share experts in the first layers, and the middle layers route all languages alike.
\end{takeaway}

\section{Conclusion}
\label{sec:conclusion}

We presented \model, a mixture of experts base model with 7.2B total and 0.6B active parameters.
It was pretrained on 6T tokens, only 41\% of which were in English, and in the anneal, text from the 44 other languages is the largest domain in the mixture.
Its context was then extended to 128k tokens.
Among the 36 open base models with up to 2B active parameters, \model ranks first on the Overall average, on math and on the multilingual average in the languages it supports, and second on English.
It achieves this at a small inference cost: for a single user, it generates text at least 1.8 times faster than every model with a higher Overall average.
It is also the first among the models we compared on the multilingual long-context benchmark, ONERULER, at 32k tokens (\cref{sec:ce-eval}).
The concern that languages compete for the capacity of a small model found little support in \model.
Replacing its multilingual data with English in the anneal raised the English average by only 0.3 points and lowered the multilingual average by 2.7 (\cref{sec:abl-multilingual}).
Its languages also share the same experts rather than dividing them (\cref{sec:abl-routing}).

Beyond the absence of a substantial capacity penalty, our results motivate a broader hypothesis: linguistic diversity may contribute to general-purpose intelligence by exposing a model to complementary descriptions of the world.
This interpretation connects multilingual pretraining to established ideas in multi-task learning and representation learning.
Testing whether language diversity directly improves general reasoning remains an important direction for future research.

\paragraph{Release.}
We release the \href{https://huggingface.co/translated/Lara-4-Mini-Base}{weights of \modelbase} under the CC BY-NC-SA 4.0 license, and will release T-HellaSwag (\cref{app:t-hellaswag}).
The data pipeline and the training pool remain internal, but \cref{sec:data,app:data} describe the mixture in full.

\paragraph{Limitations.}
\model is a base model: it is not tuned to follow instructions or for safety, and should not be deployed as an assistant without further training.
Its quality varies across languages.
Among the languages it supports, it scores lowest in some with limited training data, such as Azerbaijani and Bengali (\cref{app:per-language}).
Our analyses show that the multilingual data cost English little in the anneal.
They do not measure the effect of these data during the main stage, nor show which of our choices keeps the cost small.

\paragraph{Future work.}
We plan to post-train \model into an instruction-following model, extend the training to larger budgets and more languages, and further study how the multilingual share of the mixture affects each language and English.
We also plan controlled experiments that vary linguistic diversity during the main pretraining stage, while keeping compute, total tokens, and data quality comparable, and measure the effects on English, math, code, and multilingual performance.

\clearpage
\section*{Acknowledgments}
\addcontentsline{toc}{section}{Acknowledgments}

Our research is supported by the DVPS project, funded by the European Union's Horizon Europe programme under grant agreement No.~101213369, and by CINECA.

\FloatBarrier %
\bibliographystyle{plainnat}
\bibliography{bibliography}

\clearpage
\appendix
\section*{Appendix}
\addcontentsline{toc}{section}{Appendix}
\Cref{app:data} details the data mixture: the sources and licenses for each data group, the tokens in each group in the pool and at each stage, and the share of each language.
\Cref{app:training} completes the training recipe with the optimizer, the choice of learning rate, the stability of the run, its parallelism, and the settings for the long-context stage.
\Cref{app:ablations} reports the small-scale ablations behind the architecture and the optimizer.
\Cref{app:eval-details} gives the evaluation protocol, the results of all 56 models on every benchmark, the results of \model in each language and the protocol of the long-context evaluation.
\Cref{app:inference} describes how we measured inference speed and reports it for \model and 45 of the compared models.
\Cref{app:t-hellaswag} describes T-HellaSwag, the translated benchmark we will release, and \cref{app:model-card} gives the model card.

\section{Data Mixture in Detail}
\label{app:data}

Within each domain, the training pool is divided into data groups by kind of data and by quality.

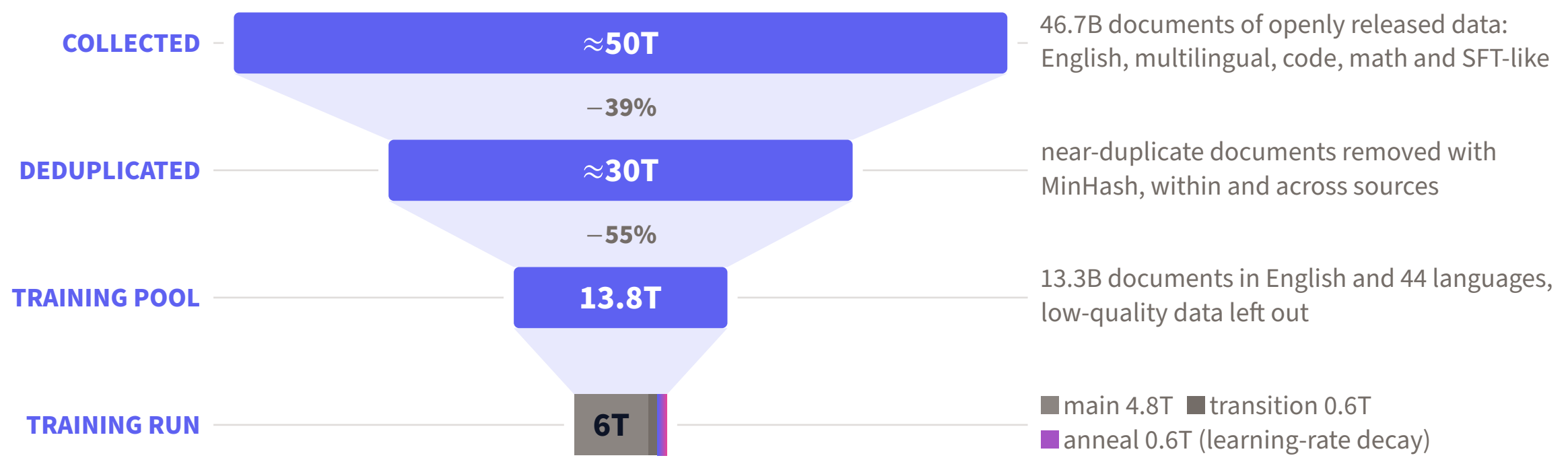
\begin{figure}[htp]
  \centering
  \begin{tikzpicture}[font=\sffamily,
    lrname/.style={anchor=east, font=\sffamily\scriptsize\bfseries, text=brandaccent, inner sep=2pt},
    lrdesc/.style={anchor=west, font=\sffamily\scriptsize, text=chartsecondary, text width=5.0cm,
      align=flush left, inner sep=2pt},
    lrnum/.style={font=\sffamily\small\bfseries, inner sep=1pt},
    lrdrop/.style={font=\sffamily\scriptsize\bfseries, text=chartsecondary, inner sep=1pt},
    lrguide/.style={draw=chartgrid, line width=0.5pt},
  ]
  \def\lrk{0.152}  %
  \def\lrx{4.05}   %
  \pgfmathsetmacro\lra{0.5*\lrk*50}     %
  \pgfmathsetmacro\lrb{0.5*\lrk*30}     %
  \pgfmathsetmacro\lrc{0.5*\lrk*13.81}  %
  \pgfmathsetmacro\lrd{0.5*\lrk*6}      %
  \fill[brandaccent!14] (-\lra,-0.3) -- (\lra,-0.3) -- (\lrb,-0.95) -- (-\lrb,-0.95) -- cycle;
  \fill[brandaccent!14] (-\lrb,-1.55) -- (\lrb,-1.55) -- (\lrc,-2.2) -- (-\lrc,-2.2) -- cycle;
  \fill[brandaccent!14] (-\lrc,-2.8) -- (\lrc,-2.8) -- (\lrd,-3.45) -- (-\lrd,-3.45) -- cycle;
  \node[lrdrop] at (0,-0.625) {$-$39\%};
  \node[lrdrop] at (0,-1.875) {$-$55\%};
  \fill[brandaccent, rounded corners=1.5pt] (-\lra,-0.3) rectangle (\lra,0.3);
  \fill[brandaccent, rounded corners=1.5pt] (-\lrb,-1.55) rectangle (\lrb,-0.95);
  \fill[brandaccent, rounded corners=1.5pt] (-\lrc,-2.8) rectangle (\lrc,-2.2);
  \fill[ordinal1] (-\lrd,-4.05) rectangle ({0.6*\lrd},-3.45);
  \fill[ordinal2] ({0.6*\lrd},-4.05) rectangle ({0.8*\lrd},-3.45);
  \fill[gradfill] ({0.8*\lrd},-4.05) rectangle (\lrd,-3.45);
  \node[lrnum, text=white] at (0,0) {$\approx$50T};
  \node[lrnum, text=white] at (0,-1.25) {$\approx$30T};
  \node[lrnum, text=white] at (0,-2.5) {13.8T};
  \node[lrnum, text=brandink] at ({-0.2*\lrd},-3.75) {6T};  %

  \draw[lrguide] (-\lrx+0.05,0) -- (-\lra-0.1,0)         (\lra+0.1,0) -- (\lrx-0.05,0);
  \draw[lrguide] (-\lrx+0.05,-1.25) -- (-\lrb-0.1,-1.25) (\lrb+0.1,-1.25) -- (\lrx-0.05,-1.25);
  \draw[lrguide] (-\lrx+0.05,-2.5) -- (-\lrc-0.1,-2.5)   (\lrc+0.1,-2.5) -- (\lrx-0.05,-2.5);
  \draw[lrguide] (-\lrx+0.05,-3.75) -- (-\lrd-0.1,-3.75) (\lrd+0.1,-3.75) -- (\lrx-0.05,-3.75);

  \node[lrname] at (-\lrx,0)     {COLLECTED};
  \node[lrname] at (-\lrx,-1.25) {DEDUPLICATED};
  \node[lrname] at (-\lrx,-2.5)  {TRAINING POOL};
  \node[lrname] at (-\lrx,-3.75) {TRAINING RUN};
  \node[lrdesc] at (\lrx,0)     {46.7B documents of openly released data: English, multilingual, code, math and SFT-like};
  \node[lrdesc] at (\lrx,-1.25) {near-duplicate documents removed with MinHash, within and across sources};
  \node[lrdesc] at (\lrx,-2.5)  {13.3B documents in English and 44 languages, low-quality data left out};
  \node[lrdesc] at (\lrx,-3.75) {{\color{ordinal1}\rule{5pt}{5pt}}\,main 4.8T\enspace {\color{ordinal2}\rule{5pt}{5pt}}\,transition 0.6T\\
    {\color{ordinal3}\rule{5pt}{5pt}}\,anneal 0.6T (learning-rate decay)};
\end{tikzpicture}
  \caption{From collected data to the 6T training run, with bar widths proportional to tokens; deduplication and selection are described in \cref{sec:data-filtering}. Deduplication removes about 40\% of the collected tokens.}
  \label{fig:data-pipeline}
\end{figure}

\begin{table}[htp]
  \centering
  \small
  \caption{Data groups of the training pool, by domain, with the licenses of their sources. NVIDIA DA: NVIDIA Data Agreement for Model Training; permissive: only files under permissive licenses. Tokens and epochs per group and stage are in \cref{tab:app-groups}.}
  \label{tab:data-sources}
  \begin{tabularx}{\linewidth}{@{}llL>{\raggedright\arraybackslash}p{2.6cm}@{}}
    \toprule
    \textbf{Domain} & \textbf{Group} & \textbf{Sources} & \textbf{Licenses} \\
    \midrule
    \multirow{5}{*}{English}
      & Web, high quality   & Nemotron-CC high and medium-high (v1 and v2), FineWeb-Edu & NVIDIA DA, ODC-BY \\
      & Web, medium quality & Nemotron-CC medium (v1 and v2) & NVIDIA DA \\
      & Synthetic           & Nemotron-CC high-quality synthetic, v2 synthetic and diverse QA & NVIDIA DA \\
      & PDFs                & FinePDFs (English) & ODC-BY \\
      & Wikipedia           & FineWiki (English) & CC BY-SA 4.0, GFDL \\
    \midrule
    \multirow{5}{*}{Multilingual}
      & Web                 & HPLT~3.0, FineWeb-2, CulturaX & CC0, ODC-BY \\
      & Web, high quality   & FineWeb2-HQ & ODC-BY \\
      & Synthetic           & Nemotron-CC v2 translated diverse QA, FineWeb2-HQ QA & NVIDIA DA, ODC-BY \\
      & PDFs                & FinePDFs (non-English) & ODC-BY \\
      & Wikipedia           & FineWiki & CC BY-SA 4.0, GFDL \\
    \midrule
    \multirow{2}{*}{Code}
      & Real                & Nemotron-CC code, The Stack~v3 (permissive licenses only) & NVIDIA DA, permissive \\
      & Synthetic           & Nemotron GitHub code (QA, review, rewriting, transpilation) & NVIDIA DA \\
    \midrule
    \multirow{2}{*}{Math}
      & High quality        & FineMath 4+, MegaMath-Web-Pro, Nemotron-CC-Math 4+ and its MIND synthetic math dialogues & ODC-BY, NVIDIA DA \\
      & Web                 & FineMath 3+, MegaMath, Nemotron-CC-Math 3 & ODC-BY, NVIDIA DA \\
    \midrule
    \multirow{3}{*}{SFT-like}
      & Question answering  & Nemotron SFT-General, STEM-SFT, Fact-Seeking, Multiple-Choice and other question-answer sets & NVIDIA DA, CC BY 4.0, CC BY-SA 4.0, GFDL \\
      & Math                & Nemotron SFT-Math, Math-Textbooks & NVIDIA DA, CC BY 4.0 \\
      & Code                & Nemotron SFT-Code, Code-Concepts, Scientific-Coding & NVIDIA DA, CC BY 4.0, CC BY-SA 4.0, GFDL \\
    \bottomrule
  \end{tabularx}
\end{table}

\begin{table}[htp]
  \centering
  \small
  \caption{Tokens per data group in the training pool and in each stage (billions of Qwen3 tokens), and epochs over training (tokens read divided by tokens in the pool). Four groups are read more than once: Wikipedia in both domains, high-quality multilingual web and high-quality math.}
  \label{tab:app-groups}
  \begin{tabular}{@{}lrrrrr@{}}
    \toprule
    \textbf{Group} & \textbf{Pool} & \textbf{Main} & \textbf{Transition} & \textbf{Anneal} & \textbf{Epochs} \\
    \midrule
    English web, high quality       & 1391.4 & 1152.9 & 108.2 & 76.1 & 0.96 \\
    English web, medium quality     & 1100.0 &  143.5 &   7.2 &  0.0 & 0.14 \\
    English synthetic, high quality & 1916.3 &  538.9 &  64.5 & 56.7 & 0.34 \\
    English PDFs                    & 1236.6 &  266.1 &  29.4 & 24.3 & 0.26 \\
    English Wikipedia               &    8.2 &   10.6 &   3.7 &  4.9 & 2.34 \\
    \midrule
    Multilingual web                & 3326.5 &  724.7 &  50.8 & 12.8 & 0.24 \\
    Multilingual web, high quality  &  638.2 &  510.2 &  73.3 & 80.0 & 1.04 \\
    Multilingual synthetic          &  658.6 &  188.2 &  38.0 & 56.3 & 0.43 \\
    Multilingual PDFs               & 1663.0 &  318.7 &  40.6 & 39.2 & 0.24 \\
    Multilingual Wikipedia          &   27.5 &   34.3 &   7.3 &  9.8 & 1.87 \\
    \midrule
    Code, real                      &  466.6 &  345.6 &  50.0 & 56.1 & 0.97 \\
    Code, synthetic                 &  434.9 &  230.4 &  37.0 & 45.9 & 0.72 \\
    \midrule
    Math, high quality              &  137.2 &  130.4 &  29.8 & 46.0 & 1.50 \\
    Math web                        &  189.0 &  109.6 &  18.2 & 20.0 & 0.78 \\
    \midrule
    SFT-like, question answering    &  363.6 &   52.8 &  23.1 & 39.6 & 0.32 \\
    SFT-like, math                  &  199.7 &   33.6 &  14.7 & 25.2 & 0.37 \\
    SFT-like, code                  &   51.5 &    9.6 &   4.2 &  7.2 & 0.41 \\
    \midrule
    \textbf{Total}                  & \textbf{13808.8} & \textbf{4800.0} & \textbf{600.0} & \textbf{600.0} & \textbf{0.43} \\
    \bottomrule
  \end{tabular}
\end{table}

\begin{figure}[htp]
  \centering
  \begin{tikzpicture}
  \begin{axis}[lrchart,
      width=0.985\linewidth, height=5.8cm,
      ybar, bar width=4.6pt, bar shift=0pt,
      enlarge x limits={abs=8pt},
      ymin=0, ymax=21, ytick={0,5,10,15,20},
      yticklabel={\pgfmathprintnumber{\tick}\%},
      ymajorgrids, axis x line*=bottom, axis y line*=left,
      ylabel={Share of multilingual tokens},
      table/col sep=comma, xtick={0,...,43},
      xticklabels from table={figures/data/languages.csv}{name},
      x tick label style={rotate=90, anchor=east, font=\sffamily\tiny, text=chartsecondary},
      legend style={at={(0.99,0.97)}, anchor=north east, legend columns=1,
        /tikz/every even column/.append style={column sep=4pt}},
      legend image code/.code={\fill[#1] (0pt,-2.2pt) rectangle (6pt,3.8pt);},
    ]
    \addplot[draw=none, fill=brandgradstart] table[col sep=comma, x=idx, y=eu]  {figures/data/languages.csv};
    \addlegendentry{EU official language}
    \addplot[draw=none, fill=brandgradend] table[col sep=comma, x=idx, y=t12] {figures/data/languages.csv};
    \addlegendentry{Other prioritized language}
    \addplot[draw=none, fill=chartmuted] table[col sep=comma, x=idx, y=oth] {figures/data/languages.csv};
    \addlegendentry{Other language ($\geq$20B tokens in pool)}
    \addplot[only marks, mark=-, mark size=3.4pt, mark options={draw=chartink, line width=1.1pt},
        legend image code/.code={\draw[chartink, line width=1.1pt] (-0.5pt,0.8pt) -- (6.5pt,0.8pt);}]
      table[col sep=comma, x=idx, y=natural] {figures/data/languages.csv};
    \addlegendentry{Proportional to unique tokens}
  \end{axis}
\end{tikzpicture}
  \caption{Share of each language in the 2.18T multilingual tokens of the 6T run (bars), against its share under sampling proportional to its unique tokens after deduplication (dashes). Tokens and epochs per language are in \cref{tab:app-languages}. The language weights increase the share of official EU languages from 46\% to 61\% of multilingual tokens.}
  \label{fig:languages}
\end{figure}
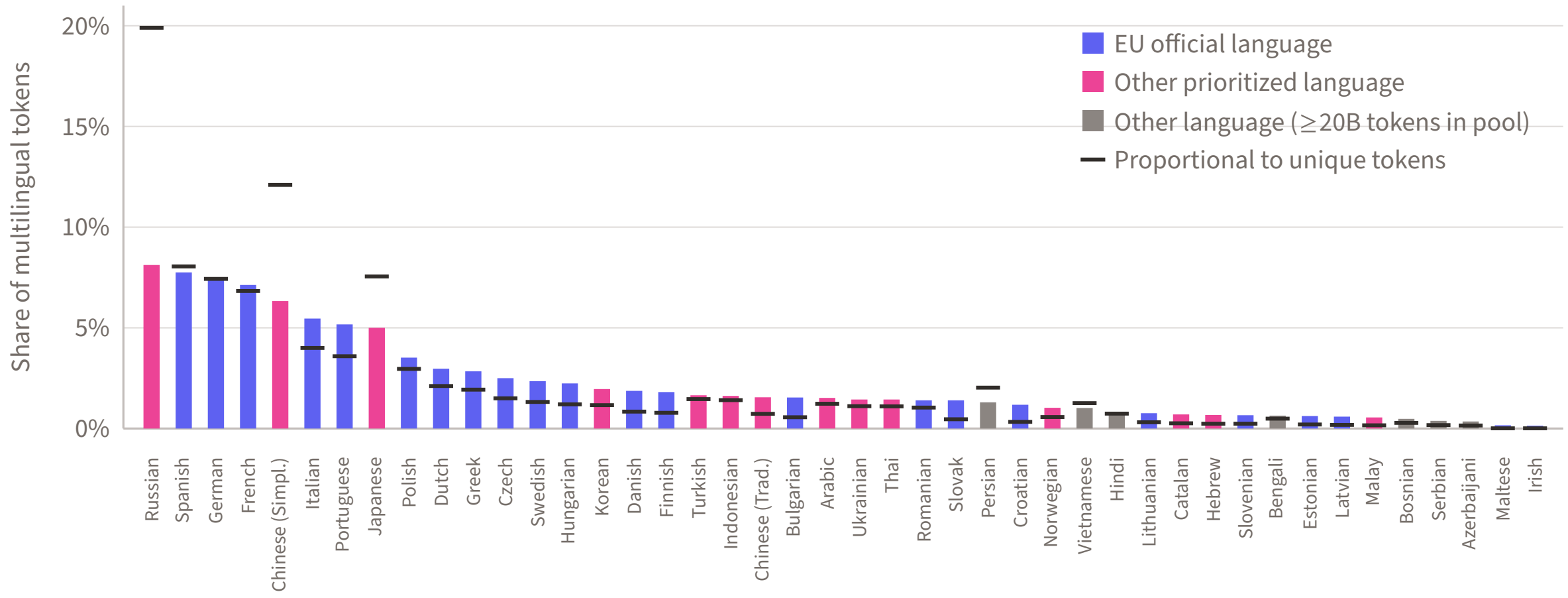

\begin{table}[htp]
  \centering
  \small
  \caption{Training tokens (billions of Qwen3 tokens) and epochs per language, over the first three stages (6T tokens). Class: official EU language (EU), other prioritized language (P), or neither (--). Each language is weighted by the square root of its unique tokens, with extra weight for official EU languages and prioritized languages (\cref{sec:data-mixture}). Only Irish and Maltese are read more than once.}
  \label{tab:app-languages}
  \begin{tabular}{@{}llrr@{\hspace{2.2em}}llrr@{}}
    \toprule
    \textbf{Language} & \textbf{Class} & \textbf{Tokens} & \textbf{Epochs} & \textbf{Language} & \textbf{Class} & \textbf{Tokens} & \textbf{Epochs} \\
    \midrule
    Russian & P & 177.2 & 0.29 & Ukrainian & P & 31.4 & 0.16 \\
    Spanish & EU, P & 169.2 & 0.33 & Thai & P & 31.4 & 0.16 \\
    German & EU, P & 162.4 & 0.35 & Romanian & EU & 30.5 & 0.44 \\
    French & EU, P & 155.7 & 0.37 & Slovak & EU, P & 30.5 & 0.37 \\
    Chinese (Simpl.) & P & 138.3 & 0.48 & Persian & -- & 28.3 & 0.30 \\
    Italian & EU, P & 119.3 & 0.44 & Croatian & EU, P & 25.7 & 0.44 \\
    Portuguese & EU, P & 113.0 & 0.43 & Norwegian & P & 22.4 & 0.22 \\
    Japanese & P & 109.3 & 0.28 & Vietnamese & -- & 22.2 & 0.44 \\
    Polish & EU, P & 76.9 & 0.42 & Hindi & -- & 17.2 & 0.13 \\
    Dutch & EU, P & 64.9 & 0.44 & Lithuanian & EU & 16.6 & 0.30 \\
    Greek & EU, P & 62.1 & 0.37 & Catalan & P & 15.2 & 0.34 \\
    Czech & EU, P & 54.7 & 0.44 & Hebrew & P & 14.6 & 0.35 \\
    Swedish & EU, P & 51.4 & 0.46 & Slovenian & EU & 14.4 & 0.35 \\
    Hungarian & EU, P & 49.0 & 0.44 & Bengali & -- & 13.9 & 0.16 \\
    Korean & P & 42.9 & 0.46 & Estonian & EU & 13.5 & 0.38 \\
    Danish & EU, P & 40.9 & 0.28 & Latvian & EU & 12.9 & 0.40 \\
    Finnish & EU, P & 39.6 & 0.29 & Malay & P & 12.0 & 0.41 \\
    Turkish & P & 35.9 & 0.48 & Bosnian & -- & 10.5 & 0.21 \\
    Indonesian & P & 35.5 & 0.46 & Serbian & -- & 8.3 & 0.27 \\
    Chinese (Trad.) & P & 34.0 & 0.54 & Azerbaijani & -- & 7.6 & 0.29 \\
    Bulgarian & EU, P & 33.5 & 0.34 & Maltese & EU & 3.5 & 1.46 \\
    Arabic & P & 33.1 & 0.32 & Irish & EU & 3.2 & 1.60 \\
    \bottomrule
  \end{tabular}
\end{table}

\subsection{Personal and Harmful Content}
\label{app:data-pii}
We do not add our own filters for personal information or harmful content and rely on each source's processing.
Common Crawl, the origin of most of our web data, respects \texttt{robots.txt} at crawl time.
FineWeb-2, FineWeb2-HQ, and FinePDFs anonymize email and IP addresses; HPLT~3.0 filters adult content and credentials; and FineWeb-Edu filters adult content.
Some personal information and harmful content remain in the training data, as in any web-scale corpus.

\section{Training}
\label{app:training}

\subsection{Training Details}
\label{app:training-details}

\paragraph{Optimizer.}
Muon~\citep{jordan2024muon} replaces the update of a weight matrix with an orthogonalized version of its momentum, computed with five Newton--Schulz iterations.
We use Nesterov momentum with coefficient 0.95.
Muon updates every two-dimensional weight inside the layers: the attention projections, the input and output projections of the Gated DeltaNet layers, and the routed and shared experts.
Queries, keys and values are orthogonalized separately, and so is each expert.
AdamW handles the tied embedding, the normalization weights, the routers, and the one-dimensional and convolutional parameters of the Gated DeltaNet layers.
Following \citet{liu2025muonscalable}, we scale each orthogonalized update by $0.2\sqrt{\max(n_{\mathrm{in}}, n_{\mathrm{out}})}$, which matches its root-mean-square size to that of an AdamW update.
All parameters therefore share a single learning rate and a decoupled weight decay of 0.1.

\paragraph{Learning rate.}
The ablations in \cref{app:ablations} used a learning rate of $1.2 \times 10^{-3}$, but the 6T run has twice their batch size and 60 times their number of tokens, so we tuned it again.
Since a sweep over 6T tokens was out of reach, we measured how the optimal learning rate varies with training length and extrapolated the trend to 6T.
The sweep ran on a quarter-width proxy with $\mu$P~\citep{yang2022mup}, so that its best learning rate carries over to the full width, at the batch size of the final run and up to 74B tokens.
At constant learning rate, the best value at a given length is hard to read, because the loss does not include the gain of the final decay.
Reading it from averages of recent checkpoints, the best learning rate decreased with training length and extrapolated to about $3 \times 10^{-4}$; this estimate is biased low, since the average recovers less of the decay gain of high learning rates as training gets longer.
Reading it after real decays at 22, 50 and 74B tokens, the best learning rate did not fall, which gives about $2.4 \times 10^{-3}$; this estimate is biased high, since the optimum may still fall beyond 74B.
We chose $8 \times 10^{-4}$, close to the geometric mean of the two estimates.

\begin{figure}[H]
  \centering
  \begin{tikzpicture}
  \begin{groupplot}[lrchart,
      group style={group size=3 by 1, horizontal sep=1.3cm},
      scale only axis, width=4.0cm, height=3.4cm,
      xmin=0, xmax=6.05, xtick={0,2,4,6},
      xticklabel={\pgfmathprintnumber{\tick}T},
      xlabel={Training tokens},
      ymajorgrids,
      table/col sep=comma,
    ]
    \nextgroupplot[ymin=1.2, ymax=2.3, ytick={1.4,1.6,1.8,2.0,2.2},
      yticklabel style={/pgf/number format/.cd, fixed, fixed zerofill, precision=1},
      title={(a) Loss},
      legend style={at={(0.97,0.97)}, anchor=north east},
      legend image post style={xscale=0.8}]
    \fill[ordinal2!25] (axis cs:4.8,1.2) rectangle (axis cs:5.4,2.3);
    \fill[gradtint=25] (axis cs:5.4,1.2) rectangle (axis cs:6.0,2.3);
    \addplot[chartink, line width=1.0pt, line join=round]
      table[x=tokens, y=loss] {figures/data/training-curves.csv};
    \addlegendentry{Training}
    \addplot[only marks, mark=*, mark size=1.5pt,
      mark options={draw=white, line width=0.4pt, fill=brandgradstart}]  %
      table[x=tokens, y=loss] {figures/data/heldout-fixed.csv};
    \addlegendentry{Held-out}

    \nextgroupplot[ymin=0, ymax=1.0, ytick={0,0.25,0.5,0.75,1.0},
      yticklabel style={/pgf/number format/.cd, fixed, fixed zerofill, precision=2},
      title={(b) Gradient norm}]
    \fill[ordinal2!25] (axis cs:4.8,0) rectangle (axis cs:5.4,1.0);
    \fill[gradtint=25] (axis cs:5.4,0) rectangle (axis cs:6.0,1.0);
    \addplot[chartink, line width=1.0pt, line join=round]
      table[x=tokens, y=gradnorm] {figures/data/training-curves.csv};

    \nextgroupplot[ymin=1.0, ymax=1.08, ytick={1.00,1.02,1.04,1.06,1.08},
      yticklabel style={/pgf/number format/.cd, fixed, fixed zerofill, precision=2},
      title={(c) Load-balancing loss}]
    \fill[ordinal2!25] (axis cs:4.8,1.0) rectangle (axis cs:5.4,1.08);
    \fill[gradtint=25] (axis cs:5.4,1.0) rectangle (axis cs:6.0,1.08);
    \addplot[chartink, line width=1.0pt, line join=round]
      table[x=tokens, y=lbl] {figures/data/training-curves.csv};
  \end{groupplot}
\end{tikzpicture}
  \caption{Training curves: (a) training loss, smoothed, and loss on the fixed held-out set at every evaluated checkpoint; (b) gradient norm before clipping; (c) load-balancing loss of the MoE layers, divided by its coefficient (1 for uniform routing). Shading marks the transition (4.8--5.4T) and anneal (5.4--6T) stages; the first 17B tokens of warmup are not shown. Gradient-norm spikes occur only at a single step, and the load-balancing loss remains close to 1 throughout training.}
  \label{fig:training-dynamics}
\end{figure}
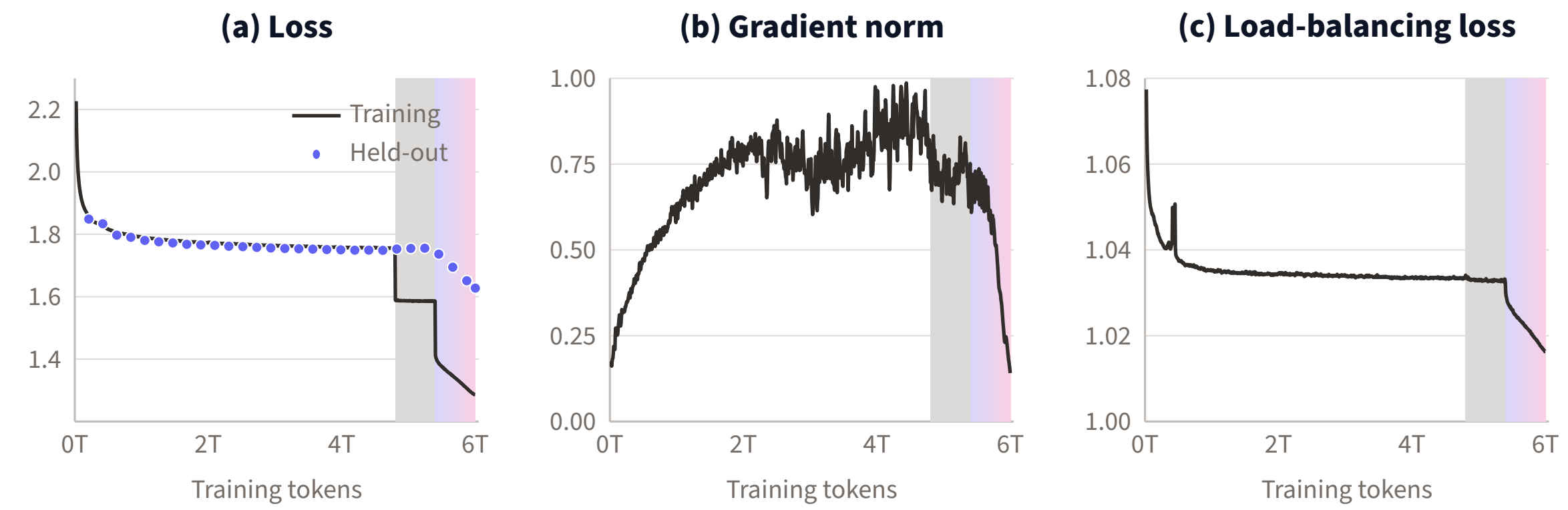

\paragraph{Stability.}
\Cref{fig:training-dynamics} shows the training curves.
The rare spikes of the gradient norm lasted a single step and did not move the loss.
The fixed held-out set in \cref{sec:pt-stability} has 134M tokens, drawn as with the main-stage data.
Its loss fell from 1.85 at 0.2T tokens to 1.75 at the end of the main stage, stayed there through the transition and fell to 1.63 after the anneal.
The gradient norm grew slowly at constant learning rate and fell with it during the anneal.
The load-balancing loss, which is 1 when all experts receive the same number of tokens, settled near 1.03 within the first trillion tokens and fell to 1.02 during the anneal.

\FloatBarrier %

\paragraph{Parallelism.}
We used data parallelism and expert parallelism of degree 4, and no tensor, pipeline, or context parallelism.

\paragraph{Long-context stage.}
\Cref{tab:ce-recipe} gives the settings and the data mixture of the two steps of the long-context stage (\cref{sec:context-extension,sec:data-long-context}).
The part of the mixture reweighted toward long documents applies the length weights of Kolibri~\citep{alephalpha2026kolibri} to the anneal mixture.

\begin{table}[H]
  \centering
  \small
  \caption{The two steps of the long-context stage: training settings and data mixture, as shares of the tokens of each step. Settings not listed are as in the first three stages (\cref{tab:pt-hparams}).}
  \label{tab:ce-recipe}
  \begin{tabular}{@{}lrrr@{}}
    \toprule
                                                     & \textbf{32k step} & \textbf{128k step} & \textbf{Overall} \\
    \midrule
    Sequence length (tokens)                         & 32{,}768 & 131{,}072 & -- \\
    Global batch size (sequences)                    & 256      & 64        & -- \\
    Steps                                            & 4{,}768  & 7{,}153   & 11{,}921 \\
    Tokens                                           & 40B      & 60B       & 100B \\
    \midrule
    Anneal mixture, as is                            & 33.3\%   & 50.0\%    & 43.3\% \\
    Anneal mixture, reweighted toward long documents & 25.5\%   & 9.0\%     & 15.6\% \\
    Long documents                                   & 30.3\%   & 30.0\%    & 30.1\% \\
    \quad FinePDFs                                   & 25.3\%   & 30.0\%    & 28.1\% \\
    \quad FineWeb2-HQ                                & 5.0\%    & --        & 2.0\% \\
    Math and code                                    & 10.9\%   & 11.0\%    & 11.0\% \\
    \quad Nemotron-CC-Math, up to 16k tokens         & 5.4\%    & 6.0\%     & 5.8\% \\
    \quad SFT-like math, 16k to 32k tokens           & 2.0\%    & --        & 0.8\% \\
    \quad Real code                                  & 3.5\%    & 5.0\%     & 4.4\% \\
    \bottomrule
  \end{tabular}
\end{table}

\subsection{Software}
\label{app:software}
We trained \model with our internal training code, built on Megatron-Bridge~0.5.0 and Megatron-Core~0.18.0~\citep{shoeybi2019megatron}.
The evaluation runs on lm-evaluation-harness and vLLM (\cref{app:eval-details}), and the inference benchmark on vLLM~0.28.0 (\cref{app:inference}).
The released model runs in Hugging Face Transformers~\citep{wolf2020transformers}, with the Gated DeltaNet kernels of flash-linear-attention~\citep{yang2024fla} when they are installed.

\section{Design Ablations}
\label{app:ablations}

The main text states the result of each ablation where it describes the corresponding choice (\cref{sec:architecture,sec:pt-recipe}).
The ablation runs were scored on the development suite defined in \cref{sec:ml-analysis}.

\subsection{Model Class, Shape and Optimizer}
\label{sec:abl-architecture}

We compared candidate architectures at about 0.5B active parameters (\cref{tab:abl-arch}) under a protocol fixed before the first run.
Each candidate was trained once on 100B tokens using the same data in the same order, with 4{,}096-token sequences and 1{,}024-sequence batches.
The learning rate peaked at $1.2 \times 10^{-3}$, stayed constant up to 80B tokens and decayed to zero over the last 20B~\citep{hagele2024scaling}.
We report the mean scores on the development suite and the final model's loss on a fixed held-out set of 128M tokens, evaluated using the same batches for every model.
This loss is not comparable with the held-out loss of \cref{app:training}.
Losses at intermediate token counts are those of the average of eight constant-learning-rate checkpoints, which approximates a model annealed at that point~\citep{li2025merging}.
These runs used global attention in every layer, and AdamW except in the last comparison.
The hybrid layers (\cref{sec:abl-hybrid}) were adopted afterward.

\begin{table}[H]
  \centering
  \small
  \caption{Model class, shape and optimizer at 100B tokens: loss on the fixed held-out set (lower is better) and mean benchmark scores, English (8 benchmarks) and multilingual (9). Experts are given as number $\times$ hidden size, with the number active per token. All rows use AdamW except the last. Highlighted is the configuration of \model, before the hybrid attention. Every MoE ends with a lower loss than both dense models.}
  \label{tab:abl-arch}
  \begin{tabular}{@{}lllrrrrr@{}}
    \toprule
    \textbf{Layers} & \textbf{Routed experts} & \textbf{Shared} & \textbf{Active} & \textbf{Total} & \textbf{Loss} & \textbf{English} & \textbf{Multilingual} \\
    \midrule
    28 & \multicolumn{2}{l}{dense}                 & 0.60B & 0.60B & 2.431 & 50.4 & 36.9 \\
    28 & \multicolumn{2}{l}{dense}                 & 1.24B & 1.24B & 2.285 & 53.7 & 38.3 \\
    24 & $64 \times 768$, top-4  & --              & 0.51B & 3.91B & 2.201 & 55.8 & 38.8 \\
    24 & $128 \times 768$, top-4 & --              & 0.51B & 7.53B & 2.160 & 56.3 & 39.5 \\
    34 & $128 \times 512$, top-4 & --              & 0.55B & 7.18B & 2.159 & 56.7 & 39.7 \\
    34 & $128 \times 512$, top-4 & $1 \times 512$  & 0.60B & 7.24B & 2.140 & 57.3 & 40.0 \\
    \ourrow \multicolumn{5}{@{}l}{Same, with Muon} & \textbf{2.098} & \textbf{58.1} & \textbf{40.7} \\
    \bottomrule
  \end{tabular}
\end{table}

\paragraph{Mixture of experts and shape.}
The MoE was ahead of both dense models from the start.
With 64 experts, the MoE ended with a lower loss than the 1.24B dense model, which has 2.4 times as many active parameters.
The gap between the MoE and the 0.6B dense model was already 0.219 nats at 5B tokens and did not shrink.
Doubling the experts at the same active compute reduced the loss further as training progressed, from 0.011 at 5B tokens to 0.041 at 100 B tokens.
The deeper, narrower shape with 34 layers started behind (0.036 higher loss at 5B tokens) and tied the 24-layer shape at 100B.
We chose it because it scored at or above the 24-layer shape on 13 of the 17 benchmarks and has 0.35B fewer total parameters.
A shared expert~\citep{dai2024deepseekmoe}, which every token passes through, adds 0.05B active parameters.
The loss reduction it brought was stable from about 25B tokens on, and the model with it scored higher than the one without on 12 of the 17 benchmarks.

\paragraph{Muon over AdamW.}
\label{sec:abl-optimizer}
Muon (\cref{sec:pt-recipe}) used the same peak learning rate as AdamW, selected by a sweep on a half-width proxy with $\mu$P~\citep{yang2022mup} and confirmed at full size.
With Muon, the held-out loss was lower at every intermediate point, by 0.19 at 5B tokens and by 0.04 to 0.05 from 40B tokens on.
AdamW needed 1.5 times as many tokens to reach Muon's training loss between 10B and 30B tokens, and about 1.9 times as many at 80B.
Muon was also ahead on 15 of the 17 benchmarks.

\begin{takeaway}
At about 0.5B active parameters, a fine-grained MoE beats dense models with up to 2.4 times as many active parameters; a shared expert and Muon further lower the loss.
\end{takeaway}

\subsection{Hybrid Attention}
\label{sec:abl-hybrid}

We tested how much of the global attention can be replaced by Gated DeltaNet and sliding-window attention without losing quality.
Starting from the 34-layer shape with one shared expert, trained with Muon (\cref{sec:abl-optimizer}), we compared three variants under the protocol of \cref{sec:abl-architecture} (\cref{tab:abl-hybrid}).
The first uses Gated DeltaNet~\citep{yang2025gdn} in 26 of the 34 layers, and the second uses sliding windows of 512 tokens in the same 26 layers.
The third is the hybrid of \model, with 14 Gated DeltaNet and 13 sliding-window layers arranged exactly as in \cref{sec:arch-attention}.
The remaining layers keep global attention.

\begin{table}[H]
  \centering
  \small
  \caption{Hybrid attention ablations at 100B tokens, against global attention in all 34 layers: change in held-out loss, mean change in score over the 17 benchmarks (points), and memory held per sequence during generation at 32k tokens, relative to global attention (computed as in \cref{sec:arch-attention}). The oracle takes, on each benchmark, the better score of the two single-mixer variants. The single-mixer variants keep global attention across 8 layers, while the hybrid uses 7. The hybrid has the largest benchmark gain and the smallest memory.}
  \label{tab:abl-hybrid}
  \begin{tabular}{@{}lrrr@{}}
    \toprule
    \textbf{Token mixers} & \textbf{$\Delta$ loss} & \textbf{$\Delta$ benchmarks} & \textbf{Memory at 32k} \\
    \midrule
    Global attention (34 layers)                      & --                & --               & 100\% \\
    Gated DeltaNet (26 layers)                        & \textbf{$-$0.017} & $+$0.37          & 24.2\% \\
    Sliding window (26 layers)                        & $+$0.002          & $+$0.17          & 24.7\% \\
    \ourrow Gated DeltaNet (14) + sliding window (13) & $-$0.015          & \textbf{$+$0.67} & \textbf{21.5\%} \\
    \midrule
    \textit{Oracle: better of Gated DeltaNet and sliding window} & -- & $+$0.65 & -- \\
    \bottomrule
  \end{tabular}
\end{table}

\paragraph{Gated DeltaNet improves quality; sliding windows alone do not.}
Gated DeltaNet lowered the held-out loss by a margin that appeared early in training and held to the end (\cref{tab:abl-hybrid}).
This is consistent with the lower loss of Gated DeltaNet hybrids at matched budgets reported by \citet{merrill2026olmohybrid}.
Sliding windows alone ended up on a level playing field with full attention.

\paragraph{The two mixers are complementary.}
On benchmarks, the two single-mixer variants were strong on different tasks (\cref{tab:abl-hybrid-families}).
Gated DeltaNet was better on text completion and commonsense reasoning; sliding windows were better on tasks that compare two sentences, such as entailment.
The difference did not come from the length of the context, since most benchmark prompts fit within the window.

\begin{table}[H]
  \centering
  \small
  \caption{Benchmark families: accuracy of the sliding-window and hybrid variants minus that of the Gated DeltaNet variant, in points. Sliding windows help on entailment and paraphrase; Gated DeltaNet on the other families.}
  \label{tab:abl-hybrid-families}
  \begin{tabular}{@{}llrr@{}}
    \toprule
    \textbf{Family} & \textbf{Benchmarks} & \textbf{Sliding window} & \textbf{Hybrid} \\
    \midrule
    Inference     & NLI, XNLI                  & $+$0.59 & $+$0.53 \\
    Paraphrase    & PAWS, PAWS-X               & $+$0.26 & $+$0.05 \\
    Causality     & COPA, XCOPA                & $-$0.58 & $+$0.17 \\
    Physical      & PIQA, Global PIQA          & $-$0.39 & $+$0.35 \\
    Completion    & HellaSwag, T-HellaSwag     & $-$0.96 & $-$0.07 \\
    Coreference   & WinoGrande                 & $-$2.84 & $-$0.16 \\
    Comprehension & Belebele                   & $-$0.17 & $+$0.15 \\
    Knowledge     & MMLU, Global MMLU, INCLUDE & $-$0.27 & $-$0.10 \\
    \bottomrule
  \end{tabular}
\end{table}

\paragraph{The hybrid matches the best of both.}
The hybrid kept most of the loss gain of Gated DeltaNet with about half as many Gated DeltaNet layers.
On benchmarks, it kept the advantage of Gated DeltaNet on completion and commonsense reasoning and recovered most of the advantage of sliding windows on entailment (\cref{tab:abl-hybrid-families}).
It scored above full attention on 16 of the 17 benchmarks.
Its mean gain over full attention matched that of the oracle in \cref{tab:abl-hybrid}, which neither single-mixer variant reached.

\paragraph{Long context.}
To test whether the ranking holds beyond the training length, we extended the three variants from their 100B checkpoints with the same long-context stage.
This stage trained each variant on the same 10B tokens, in 32k-token sequences, with attention masked across documents.
We then evaluated them on RULER~\citep{hsieh2024ruler}, its multilingual counterpart ONERULER~\citep{kim2025oneruler} and HELMET~\citep{yen2025helmet} at 16k and 32k tokens, and on the 17 short benchmarks (\cref{tab:abl-hybrid-long}).
The hybrid was the best on RULER and ONERULER at both lengths.
On HELMET, which also covers retrieval-augmented generation, re-ranking and in-context learning, the hybrid and Gated DeltaNet were level at 32k tokens and ahead of sliding windows.
The hybrid led on the two categories that require reading many passages, retrieval-augmented generation and re-ranking.
The extension did not hurt short-context quality: every variant gained slightly on the short benchmarks, and the hybrid remained the best.

\begin{table}[H]
  \centering
  \small
  \caption{Hybrid attention ablations after a long-context stage of 10B tokens at 32k: mean score over the 17 short benchmarks, and RULER (mean of 13 tasks), ONERULER (4 retrieval tasks in 26 languages) and HELMET (mean of its recall, RAG, re-ranking and in-context learning categories) at 16k and 32k tokens. The hybrid is best on RULER and ONERULER at both lengths.}
  \label{tab:abl-hybrid-long}
  \begin{tabular}{@{}lrrrrrrr@{}}
    \toprule
    & & \multicolumn{2}{c}{\textbf{RULER}} & \multicolumn{2}{c}{\textbf{ONERULER}} & \multicolumn{2}{c}{\textbf{HELMET}} \\
    \cmidrule(lr){3-4}\cmidrule(lr){5-6}\cmidrule(l){7-8}
    \textbf{Token mixers} & \textbf{Short} & \textbf{16k} & \textbf{32k} & \textbf{16k} & \textbf{32k} & \textbf{16k} & \textbf{32k} \\
    \midrule
    Gated DeltaNet (26 layers)                    & 49.66          & 57.3          & 46.1          & 63.5          & 55.9          & \textbf{44.9} & 39.0          \\
    Sliding window (26 layers)                    & 49.51          & 55.9          & 48.3          & 65.1          & 60.1          & 40.0          & 35.1          \\
    \ourrow Hybrid (14 + 13 layers)              & \textbf{50.13} & \textbf{57.5} & \textbf{52.8} & \textbf{71.7} & \textbf{64.0} & 43.4          & \textbf{39.6} \\
    \bottomrule
  \end{tabular}
\end{table}

\begin{takeaway}
Interleaving Gated DeltaNet with sliding-window layers retains nearly all the loss gain of Gated DeltaNet and yields the best short- and long-context scores among the three variants, using global attention in only 7 of 34 layers.
\end{takeaway}

\section{Evaluation Details}
\label{app:eval-details}
We run every benchmark with lm-evaluation-harness~\citep{biderman2024lmeval} (development version 0.4.14.dev0), using its task configurations unchanged except for the number of shots, which we fix as in \cref{tab:benchmarks}.
Generative tasks run on vLLM~\citep{kwon2023vllm} in bfloat16 with greedy decoding and a context of 4{,}096 tokens.

\begin{table}[h]
  \centering
  \small
  \caption{Benchmark suite. Each domain average is the mean of its benchmarks, and the Overall average is the mean of the four domain averages. Lang.: number of languages or language variants averaged. Multilingual benchmarks are averaged over the languages supported by \model, counting the country variants of Global PIQA (three of Spanish, two of French and of Portuguese) and the variants of T-HellaSwag separately. MGSM, a math benchmark, is averaged across all 11 of its languages. Acc.\ (norm.): accuracy normalized by answer length. Math-Verify: the final answer is checked for mathematical equivalence with the reference.}
  \label{tab:benchmarks}
  \begin{tabular}{@{}lllll@{}}
    \toprule
    \textbf{Domain} & \textbf{Benchmark} & \textbf{Shots} & \textbf{Metric} & \textbf{Lang.} \\
    \midrule
    \multirow{10}{*}{English}
      & XNLI (English)~\citep{conneau2018xnli}         & 0       & Acc.          & -- \\
      & PAWS (English)~\citep{zhang2019paws}           & 0       & Acc.          & -- \\
      & Belebele (English)~\citep{bandarkar2024belebele} & 0     & Acc.          & -- \\
      & COPA~\citep{roemmele2011copa}                  & 0       & Acc.          & -- \\
      & PIQA~\citep{bisk2020piqa}                      & 0       & Acc.\ (norm.) & -- \\
      & HellaSwag~\citep{zellers2019hellaswag}         & 10      & Acc.\ (norm.) & -- \\
      & WinoGrande~\citep{sakaguchi2020winogrande}     & 5       & Acc.          & -- \\
      & ARC-Challenge~\citep{clark2018arc}                   & 25      & Acc.\ (norm.) & -- \\
      & MMLU~\citep{hendrycks2021mmlu}                 & 5       & Acc.          & -- \\
      & MMLU-Pro~\citep{wang2024mmlupro}               & 5, CoT  & Exact match   & -- \\
    \midrule
    \multirow{9}{*}{Multilingual}
      & XNLI~\citep{conneau2018xnli}                   & 0       & Acc.          & 12 \\
      & PAWS-X~\citep{yang2019pawsx}                   & 0       & Acc.          & 6 \\
      & Belebele~\citep{bandarkar2024belebele}         & 0       & Acc.          & 42 \\
      & XCOPA~\citep{ponti2020xcopa}                   & 0       & Acc.          & 7 \\
      & Global PIQA, non-parallel~\citep{chang2025globalpiqa}       & 0       & Acc.          & 45 \\
      & Global PIQA, parallel                          & 0       & Acc.          & 44 \\
      & T-HellaSwag (\cref{app:t-hellaswag})          & 0       & Acc.\ (norm.) & 48 \\
      & INCLUDE~\citep{romanou2025include}             & 5       & Acc.          & 30 \\
      & Global MMLU~\citep{singh2025globalmmlu}        & 5       & Acc.          & 12 \\
    \midrule
    \multirow{3}{*}{Math}
      & GSM8K~\citep{cobbe2021gsm8k}                   & 5       & Exact match   & -- \\
      & MATH~\citep{hendrycks2021math}                 & 4       & Math-Verify   & -- \\
      & MGSM~\citep{shi2023mgsm}                       & 8, CoT  & Exact match   & 11 \\
    \midrule
    \multirow{4}{*}{Code}
      & MBPP~\citep{austin2021mbpp}                    & 3       & pass@1        & -- \\
      & MBPP+~\citep{liu2023evalplus}                  & 3       & pass@1        & -- \\
      & CRUXEval-I~\citep{gu2024cruxeval}                      & 2       & pass@1        & -- \\
      & CRUXEval-O                                     & 2       & pass@1        & -- \\
    \bottomrule
  \end{tabular}
\end{table}

\paragraph{Baselines.}
The 56 base models we compare with come from these releases:
Llama~3, 3.1 and 3.2~\citep{grattafiori2024llama3}, Qwen2.5~\citep{qwen2024qwen25}, Qwen3~\citep{yang2025qwen3} and Qwen3.5~\citep{qwenteam2026qwen35}, Gemma~\citep{gemmateam2024gemma}, Gemma~3~\citep{gemmateam2025gemma3} and Gemma~4~\citep{gemmateam2026gemma4}, Mistral-7B~\citep{jiang2023mistral} and Ministral~3~\citep{liu2026ministral3}, SmolLM2~\citep{allal2025smollm2} and SmolLM3~\citep{hf2025smollm3}, OLMoE~\citep{muennighoff2025olmoe}, Falcon-H1~\citep{falconllm2025falconh1}, Nemotron-Flash~\citep{fu2025nemotronflash}, Granite~3~\citep{granite2024granite3}, Granite~4.0~\citep{ibm2025granite4} and Granite~4.0 Nano~\citep{ibm2025granite4nano}, Granite~4.1~\citep{ibm2026granite41} and granite-swash~\citep{ibm2026graniteswash}, LFM2 and LFM2.5~\citep{amini2025lfm2,liquidai2026lfm25,liquidai2026lfm258ba1b} and LFM3~\citep{liquidai2026lfm3}, Trinity~\citep{singh2026trinity}, Hunyuan~\citep{tencent2025hunyuan4b}, MiniCPM5~\citep{openbmb2026minicpm5}, Kanana-2~\citep{kanana2026kanana2slm}, Tiny Aya~\citep{salamanca2026tinyaya}, Apertus~\citep{apertus2025}, EuroLLM~\citep{martins2024eurollm}, EuroMoE~\citep{alves2026euromoe}, Teuken~\citep{ali2024teuken}, Salamandra~\citep{gonzalezagirre2025salamandra}, Helium~\citep{kyutai2025helium}, Luciole~\citep{openllmfrance2026luciole}, Bielik~v3~\citep{ociepa2025bielikv3}, Domyn-Edge~\citep{domyn2026edge}, Limite~\citep{prignano2026limite}, Villanova2~\citep{villanova2026villanova2} and moe-7b-1b~\citep{ibm2025moe7b1b}.

\paragraph{Answer extraction.}
The harness's default extraction sometimes scores the answer format rather than the answer, so for three benchmarks we use a stricter or more complete parser, the same for every model:
\begin{itemize}
  \item GSM8K: the first answer marked as \texttt{\#\#\#\# N}; if there is none, the first \texttt{\textbackslash boxed\{N\}}; if there is none, the first ``the answer is N''.
    We never fall back to the last number in the output.
  \item MGSM: the answer phrase of each language's native chain-of-thought prompt, allowing an optional currency sign and spaces as thousands separators.
    The harness's Bengali pattern is the English one, so it scores every model at zero.
  \item MBPP+: the MBPP generations, which must pass both MBPP's own tests and the additional EvalPlus tests~\citep{liu2023evalplus}.
\end{itemize}
MATH uses the Minerva prompts and is scored with Math-Verify over its seven subjects (5{,}000 problems).
CRUXEval-I and CRUXEval-O use direct prediction and the official answer parser.

\subsection{Full Results}
\label{app:full-results}
\Cref{tab:app-full-results} gives the domain averages of every model we evaluated, and \cref{tab:app-english,tab:app-multilingual,tab:app-multilingual-all,tab:app-math-code} give every benchmark.
Models are grouped by active parameters as in \cref{tab:main-results}, larger models first, and sorted by Overall average within each group.

\begin{table}[H]
  \centering
  \scriptsize
  \setlength{\tabcolsep}{3.2pt}
  \renewcommand{\arraystretch}{0.85} %
  \caption{Domain averages of all 56 open base models we evaluated and \model, grouped by active parameters as in \cref{tab:main-results}. Active, Total: active and total parameters. Multilingual: averaged over the languages supported by \model. Best result in each group in bold, second best underlined.}
  \label{tab:app-full-results}
  \begin{tabular}{@{}lrrrrrrr@{}}
    \toprule
    & \multicolumn{2}{c}{\textbf{Parameters}} & \multicolumn{4}{c}{\textbf{Domain average}} & \\
    \cmidrule(lr){2-3}\cmidrule(lr){4-7}
    \textbf{Model} & \textbf{Active} & \textbf{Total} & \textbf{English} & \textbf{Multilingual} & \textbf{Math} & \textbf{Code} & \textbf{Overall} \\
    \midrule
    \multicolumn{8}{@{}l}{\textit{Models with more than 4B active parameters}} \\
    Teuken-7B & 7.5B & 7.5B & 59.8 & 47.1 & 7.4 & 19.1 & 33.4 \\
    Mistral-7B-v0.3 & 7.3B & 7.3B & 68.1 & 49.0 & 24.6 & 37.4 & 44.8 \\
    Llama-3-8B & 8.0B & 8.0B & 68.8 & 53.5 & 35.8 & 43.3 & 50.4 \\
    Gemma-7B & 8.5B & 8.5B & 68.5 & 55.2 & 39.0 & 40.5 & 50.8 \\
    Llama-3.1-8B & 8.0B & 8.0B & 69.6 & 55.2 & 36.6 & 43.6 & 51.2 \\
    Apertus-v1.5-8B & 8.0B & 8.0B & 69.0 & \textbf{59.2} & 50.6 & 44.3 & 55.8 \\
    Qwen2.5-7B & 7.6B & 7.6B & \textbf{72.3} & \underline{57.3} & \underline{64.9} & \underline{55.5} & \underline{62.5} \\
    Qwen3-8B & 8.2B & 8.2B & \underline{71.9} & \textbf{59.2} & \textbf{71.3} & \textbf{61.0} & \textbf{65.9} \\
    \midrule
    \multicolumn{8}{@{}l}{\textit{Models with 2B to 4B active parameters}} \\
    salamandra-2b & 2.3B & 2.3B & 48.9 & 39.4 & 2.9 & 10.4 & 25.4 \\
    Villanova2-2B & 2.4B & 2.4B & 59.8 & 45.3 & 28.8 & 33.9 & 41.9 \\
    tiny-aya & 3.4B & 3.4B & 62.5 & 54.5 & 38.8 & 29.4 & 46.3 \\
    SmolLM3-3B & 3.1B & 3.1B & 64.7 & 48.7 & 51.0 & 43.6 & 52.0 \\
    Nemotron-Flash-3B & 2.7B & 2.7B & 65.6 & 42.0 & 59.7 & 44.5 & 52.9 \\
    granite-4.1-3b & 3.4B & 3.4B & 64.3 & 48.1 & 53.8 & 47.6 & 53.5 \\
    Ministral-3-3B & 3.4B & 3.4B & 67.7 & 55.4 & 49.9 & 42.3 & 53.8 \\
    Domyn-Edge & 3.9B & 3.9B & 68.2 & 52.0 & 54.1 & 47.3 & 55.4 \\
    Gemma-4-E4B & 4.0B & 7.9B & 70.1 & \textbf{61.1} & 44.4 & 46.0 & 55.4 \\
    MiniCPM5-2B & 2.5B & 2.5B & 65.3 & 48.0 & 66.0 & 50.8 & 57.5 \\
    Qwen3.5-4B & 4.7B & 4.7B & \underline{70.5} & \underline{58.0} & 66.1 & 52.3 & 61.7 \\
    Hunyuan-4B & 4.2B & 4.2B & 70.4 & 51.6 & \textbf{72.0} & \textbf{58.6} & \underline{63.1} \\
    Qwen3-4B & 4.0B & 4.0B & \textbf{71.2} & 57.5 & \underline{69.2} & \underline{57.0} & \textbf{63.7} \\
    \midrule
    \multicolumn{8}{@{}l}{\textit{Models with up to 2B active parameters}} \\
    Apertus-v1.1-0.5B & 0.4B & 0.4B & 45.8 & 37.8 & 2.0 & 11.1 & 24.2 \\
    Gemma-3-1B & 1.0B & 1.0B & 49.3 & 40.6 & 2.2 & 12.6 & 26.2 \\
    EuroLLM-1.7B & 1.7B & 1.7B & 48.5 & 40.1 & 3.9 & 18.0 & 27.6 \\
    EuroMoE-2.6B-A0.6B & 0.5B & 2.6B & 49.6 & 40.7 & 6.2 & 21.1 & 29.4 \\
    helium-1-2b & 2.0B & 2.0B & 59.8 & 43.3 & 6.1 & 13.5 & 30.7 \\
    Llama-3.2-1B & 1.2B & 1.2B & 51.9 & 39.7 & 5.0 & 28.0 & 31.2 \\
    Apertus-v1.1-1.5B & 1.5B & 1.5B & 53.6 & 44.8 & 5.9 & 20.6 & 31.2 \\
    Luciole-1B & 1.3B & 1.3B & 51.5 & 38.9 & 11.0 & 24.0 & 31.3 \\
    limite-1b-soup & 1.0B & 1.0B & 44.9 & 35.5 & 47.2 & 18.9 & 36.6 \\
    SmolLM2-1.7B & 1.7B & 1.7B & 58.5 & 39.1 & 17.6 & 31.8 & 36.8 \\
    OLMoE-1B-7B & 1.3B & 6.9B & 61.9 & 40.4 & 27.2 & 21.1 & 37.7 \\
    granite-swash-3b-a600m & 0.6B & 3.0B & 57.5 & 41.5 & 26.9 & 26.1 & 38.0 \\
    Nemotron-Flash-1B & 1.0B & 1.0B & 54.7 & 37.3 & 33.1 & 32.2 & 39.3 \\
    granite-3.1-3b-a800m & 0.9B & 3.3B & 59.1 & 40.9 & 25.6 & 32.6 & 39.5 \\
    Qwen3.5-0.8B & 0.8B & 0.8B & 56.1 & 44.3 & 30.3 & 28.2 & 39.7 \\
    LFM2.5-1.2B & 1.2B & 1.2B & 57.2 & 42.2 & 37.8 & 23.9 & 40.3 \\
    limite-1b & 1.0B & 1.0B & 46.8 & 36.3 & 51.0 & 27.5 & 40.4 \\
    Bielik-1.5B & 1.6B & 1.6B & 58.6 & 40.4 & 26.8 & 37.9 & 40.9 \\
    Falcon-H1-0.5B & 0.5B & 0.5B & 56.1 & 36.8 & 39.2 & 32.9 & 41.2 \\
    Gemma-4-E2B & 2.0B & 5.1B & 63.3 & \underline{53.3} & 18.0 & 32.8 & 41.9 \\
    Qwen3-0.6B & 0.6B & 0.6B & 55.3 & 43.1 & 37.2 & 34.8 & 42.6 \\
    kanana-2-1.3b & 1.3B & 1.3B & 59.5 & 40.7 & 35.5 & 39.9 & 43.9 \\
    granite-4.0-h-1b & 1.5B & 1.5B & 57.6 & 43.0 & 37.8 & 39.5 & 44.5 \\
    moe-7b-1b & 1.1B & 6.4B & 62.8 & 42.8 & 34.1 & 39.5 & 44.8 \\
    granite-4.0-1b & 1.6B & 1.6B & 60.0 & 42.5 & 43.3 & 43.0 & 47.2 \\
    Qwen2.5-1.5B & 1.5B & 1.5B & 62.6 & 46.8 & 43.0 & 38.1 & 47.6 \\
    Falcon-H1-1.5B & 1.6B & 1.6B & 63.2 & 43.4 & 46.1 & 42.6 & 48.8 \\
    Qwen3.5-2B & 1.9B & 1.9B & 63.3 & 50.0 & 45.6 & 36.7 & 48.9 \\
    LFM3-1.7B & 1.7B & 1.7B & 60.0 & 46.7 & 52.2 & 38.0 & 49.2 \\
    Trinity-Nano & 1.0B & 6.1B & \textbf{68.4} & 47.4 & 45.0 & 42.0 & 50.7 \\
    LFM2.5-8B-A1B & 1.7B & 8.5B & 65.7 & 48.6 & 52.3 & 36.6 & 50.8 \\
    Hunyuan-1.8B & 1.8B & 1.8B & 62.7 & 44.3 & 52.0 & \textbf{49.7} & 52.2 \\
    Falcon-H1-1.5B-Deep & 1.6B & 1.6B & 65.3 & 45.2 & \underline{55.2} & 47.2 & 53.2 \\
    Qwen3-1.7B & 1.7B & 1.7B & 63.8 & 50.3 & 54.6 & 46.6 & 53.8 \\
    granite-4.0-h-tiny & 1.0B & 6.9B & 66.2 & 51.8 & 49.4 & \underline{48.4} & \underline{53.9} \\
    \ourrow \model & 0.6B & 7.2B & \underline{66.4} & \textbf{56.7} & \textbf{59.9} & 45.1 & \textbf{57.0} \\
    \bottomrule
  \end{tabular}
\end{table}
\begin{table}[p]
  \centering
  \scriptsize
  \setlength{\tabcolsep}{3.2pt}
  \caption{English benchmarks for all 56 models we evaluated and \model. Bele.: Belebele; HS: HellaSwag; WG: WinoGrande; MMLU-P: MMLU-Pro. Settings in \cref{tab:benchmarks}; best result in each group in bold, second best underlined.}
  \label{tab:app-english}
  \begin{tabular}{@{}lrrrrrrrrrrr@{}}
    \toprule
    \textbf{Model} & \textbf{NLI} & \textbf{PAWS} & \textbf{Bele.} & \textbf{COPA} & \textbf{PIQA} & \textbf{HS} & \textbf{WG} & \textbf{ARC-C} & \textbf{MMLU} & \textbf{MMLU-P} & \textbf{Avg.} \\
    \midrule
    \multicolumn{12}{@{}l}{\textit{Models with more than 4B active parameters}} \\
    Teuken-7B & \textbf{56.9} & 59.4 & 45.9 & 86.0 & 78.7 & 77.4 & 71.7 & 54.0 & 49.8 & 17.9 & 59.8 \\
    Mistral-7B-v0.3 & \underline{55.3} & 68.4 & 72.1 & 88.2 & \textbf{82.1} & \textbf{83.2} & \textbf{78.8} & 60.2 & 62.5 & 30.5 & 68.1 \\
    Llama-3-8B & 54.8 & \underline{70.1} & 75.2 & 87.8 & 80.7 & 82.3 & \underline{78.5} & 58.4 & 65.5 & 34.7 & 68.8 \\
    Gemma-7B & 52.9 & 69.2 & 76.0 & \textbf{89.4} & \underline{81.0} & \underline{82.4} & 77.4 & 60.0 & 63.1 & 33.5 & 68.5 \\
    Llama-3.1-8B & \underline{55.3} & \textbf{70.4} & 81.0 & 88.2 & 80.9 & 82.3 & 78.3 & 58.0 & 65.3 & 35.8 & 69.6 \\
    Apertus-v1.5-8B & 55.0 & 65.5 & 77.9 & 87.8 & 80.1 & 80.3 & 74.4 & \underline{64.5} & 66.3 & 38.2 & 69.0 \\
    Qwen2.5-7B & 53.9 & 68.8 & \underline{88.8} & \underline{89.2} & 79.9 & 80.3 & 75.9 & 63.8 & \underline{74.2} & \underline{48.5} & \textbf{72.3} \\
    Qwen3-8B & 46.8 & 55.5 & \textbf{91.3} & 89.0 & 79.5 & 79.6 & 76.6 & \textbf{67.6} & \textbf{76.8} & \textbf{56.1} & \underline{71.9} \\
    \midrule
    \multicolumn{12}{@{}l}{\textit{Models with 2B to 4B active parameters}} \\
    salamandra-2b & 51.9 & 57.8 & 23.7 & 79.2 & 73.3 & 63.7 & 62.2 & 41.5 & 25.5 & 9.8 & 48.9 \\
    Villanova2-2B & 52.9 & 59.2 & 63.9 & 78.6 & 75.4 & 70.4 & 65.1 & 55.0 & 53.5 & 24.1 & 59.8 \\
    tiny-aya & 54.3 & 66.7 & 66.2 & 84.2 & 77.3 & 72.4 & 68.4 & 52.7 & 58.4 & 24.8 & 62.5 \\
    SmolLM3-3B & 49.5 & 61.0 & 65.8 & 85.6 & 78.5 & 77.1 & 73.4 & 60.1 & 61.3 & 35.2 & 64.7 \\
    Nemotron-Flash-3B & \textbf{55.2} & 65.6 & 70.8 & 83.8 & \underline{80.0} & 73.4 & 69.1 & 57.3 & 61.1 & 39.3 & 65.6 \\
    granite-4.1-3b & 48.1 & 58.9 & 62.7 & 83.8 & 79.5 & 75.7 & 72.9 & 59.9 & 64.3 & 37.4 & 64.3 \\
    Ministral-3-3B & 53.1 & 64.9 & 82.0 & 85.2 & 78.2 & 74.4 & 69.9 & 63.1 & 68.0 & 38.6 & 67.7 \\
    Domyn-Edge & 53.7 & \textbf{69.4} & 85.8 & \underline{86.2} & 79.3 & 75.8 & \underline{73.7} & 57.8 & 63.6 & 36.5 & 68.2 \\
    Gemma-4-E4B & 54.1 & 65.8 & 81.0 & \textbf{87.6} & \textbf{81.2} & \textbf{81.5} & \textbf{75.6} & \underline{66.0} & 70.6 & 37.9 & 70.1 \\
    MiniCPM5-2B & 53.6 & 67.0 & 74.0 & 82.6 & 77.0 & 69.1 & 66.2 & 55.0 & 65.3 & 43.2 & 65.3 \\
    Qwen3.5-4B & 44.0 & 66.3 & \underline{87.4} & 85.2 & 79.3 & \underline{77.2} & 73.1 & \textbf{66.5} & \textbf{73.1} & \underline{52.5} & \underline{70.5} \\
    Hunyuan-4B & \underline{54.8} & 67.9 & 87.3 & 84.2 & 77.5 & 72.1 & 73.6 & 59.0 & 72.5 & \textbf{55.1} & 70.4 \\
    Qwen3-4B & 54.4 & \underline{68.6} & \textbf{89.3} & 85.8 & 78.0 & 75.7 & 72.0 & 64.4 & \underline{73.0} & 51.2 & \textbf{71.2} \\
    \midrule
    \multicolumn{12}{@{}l}{\textit{Models with up to 2B active parameters}} \\
    Apertus-v1.1-0.5B & 54.1 & 53.5 & 25.6 & 74.8 & 70.4 & 52.3 & 57.2 & 36.4 & 24.7 & 9.3 & 45.8 \\
    Gemma-3-1B & 52.8 & 60.7 & 25.3 & 79.6 & 75.1 & 62.9 & 61.4 & 39.3 & 26.2 & 9.3 & 49.3 \\
    EuroLLM-1.7B & 49.4 & 59.9 & 27.9 & 76.6 & 73.9 & 60.8 & 59.6 & 40.9 & 26.9 & 9.6 & 48.5 \\
    EuroMoE-2.6B-A0.6B & 52.3 & 58.1 & 24.4 & 79.0 & 76.8 & 65.0 & 61.3 & 44.8 & 24.9 & 9.3 & 49.6 \\
    helium-1-2b & 53.1 & 65.7 & 54.6 & 82.8 & 78.1 & 73.1 & 68.9 & 52.1 & 52.0 & 17.7 & 59.8 \\
    Llama-3.2-1B & 54.5 & 60.0 & 36.2 & 81.8 & 75.0 & 65.8 & 62.8 & 40.4 & 31.1 & 11.2 & 51.9 \\
    Apertus-v1.1-1.5B & 53.5 & 62.2 & 32.1 & 81.0 & 75.4 & 65.3 & 63.3 & 47.3 & 42.7 & 13.6 & 53.6 \\
    Luciole-1B & 53.8 & 56.9 & 34.9 & 78.4 & 73.6 & 62.5 & 61.7 & 45.0 & 35.2 & 12.5 & 51.5 \\
    limite-1b-soup & 54.6 & 65.6 & 26.3 & 57.4 & 60.6 & 35.6 & 55.3 & 34.3 & 35.6 & 23.7 & 44.9 \\
    SmolLM2-1.7B & 51.0 & 66.1 & 43.7 & 82.8 & 77.6 & 73.1 & 67.9 & 53.8 & 50.1 & 19.2 & 58.5 \\
    OLMoE-1B-7B & 52.6 & 64.5 & 51.7 & \underline{87.4} & 79.7 & \textbf{79.7} & 72.1 & 58.4 & 55.0 & 18.1 & 61.9 \\
    granite-swash-3b-a600m & 53.2 & 58.0 & 48.0 & 80.2 & 77.4 & 69.2 & 66.1 & 49.6 & 54.4 & 18.9 & 57.5 \\
    Nemotron-Flash-1B & 51.2 & 60.2 & 42.6 & 79.8 & 75.4 & 61.3 & 61.1 & 46.0 & 44.6 & 25.0 & 54.7 \\
    granite-3.1-3b-a800m & \textbf{56.9} & 67.3 & 40.9 & 84.4 & 78.9 & 74.2 & 69.3 & 51.1 & 48.2 & 19.4 & 59.1 \\
    Qwen3.5-0.8B & 50.1 & 62.0 & 68.6 & 74.8 & 71.6 & 55.5 & 60.1 & 43.7 & 48.3 & 26.5 & 56.1 \\
    LFM2.5-1.2B & 52.1 & 58.7 & 63.1 & 76.4 & 73.0 & 61.0 & 61.0 & 56.0 & 55.1 & 15.3 & 57.2 \\
    limite-1b & 49.2 & 56.6 & 36.8 & 65.2 & 62.9 & 39.8 & 54.9 & 39.0 & 39.3 & 24.9 & 46.8 \\
    Bielik-1.5B & 49.8 & 66.0 & 66.2 & 77.4 & 72.3 & 64.7 & 64.2 & 46.5 & 55.2 & 23.8 & 58.6 \\
    Falcon-H1-0.5B & 52.3 & 65.2 & 50.1 & 72.0 & 71.1 & 56.3 & 58.4 & 47.1 & 55.1 & 33.0 & 56.1 \\
    Gemma-4-E2B & 53.6 & 67.8 & 64.8 & 84.8 & 78.9 & 75.3 & 68.4 & 57.6 & 57.8 & 23.6 & 63.3 \\
    Qwen3-0.6B & 52.9 & 60.5 & 56.8 & 76.0 & 70.0 & 53.5 & 60.4 & 45.1 & 52.5 & 25.7 & 55.3 \\
    kanana-2-1.3b & 51.2 & 65.5 & 60.4 & 79.2 & 76.0 & 64.1 & 62.2 & 49.9 & 56.4 & 29.9 & 59.5 \\
    granite-4.0-h-1b & 45.8 & 64.7 & 43.2 & 81.0 & 76.3 & 69.7 & 65.1 & 50.9 & 56.6 & 22.4 & 57.6 \\
    moe-7b-1b & 55.8 & 65.0 & 53.1 & \textbf{87.6} & 78.7 & \underline{78.5} & 72.4 & 56.5 & 55.8 & 24.7 & 62.8 \\
    granite-4.0-1b & 49.3 & 55.1 & 59.8 & 82.6 & 75.6 & 70.2 & 67.6 & 53.8 & 57.2 & 29.3 & 60.0 \\
    Qwen2.5-1.5B & 51.9 & 64.1 & 71.7 & 84.6 & 76.0 & 67.9 & 64.6 & 54.7 & 60.8 & 29.5 & 62.6 \\
    Falcon-H1-1.5B & \underline{56.8} & 67.1 & 65.9 & 79.8 & 74.9 & 66.8 & 65.5 & 52.9 & 61.7 & \textbf{40.4} & 63.2 \\
    Qwen3.5-2B & 53.6 & 61.7 & \textbf{82.8} & 82.0 & 74.8 & 67.6 & 66.3 & 54.1 & 54.2 & 36.4 & 63.3 \\
    LFM3-1.7B & 52.5 & 65.3 & 69.3 & 76.8 & 68.4 & 59.5 & 62.4 & 51.0 & 58.8 & 35.7 & 60.0 \\
    Trinity-Nano & 54.7 & 67.1 & \underline{79.0} & 87.2 & \textbf{81.1} & 78.3 & \textbf{74.2} & \underline{62.6} & \underline{66.1} & 33.6 & \textbf{68.4} \\
    LFM2.5-8B-A1B & 52.3 & 62.9 & 69.1 & 85.4 & 77.7 & 74.0 & 67.6 & \textbf{64.9} & 63.2 & 39.8 & 65.7 \\
    Hunyuan-1.8B & 50.7 & \textbf{71.0} & 75.4 & 78.6 & 72.9 & 62.2 & 66.1 & 50.1 & 62.9 & 37.5 & 62.7 \\
    Falcon-H1-1.5B-Deep & 55.9 & \underline{68.0} & 70.9 & 82.2 & 76.4 & 69.8 & 67.3 & 55.7 & \textbf{66.6} & \underline{40.1} & 65.3 \\
    Qwen3-1.7B & 54.2 & 61.2 & 75.9 & 82.4 & 75.5 & 67.2 & 65.4 & 55.7 & 62.6 & 37.4 & 63.8 \\
    granite-4.0-h-tiny & 54.2 & 63.3 & 69.9 & 85.4 & 79.5 & 77.7 & \underline{73.5} & 61.7 & 65.5 & 31.6 & 66.2 \\
    \ourrow \model & 52.7 & 65.3 & 76.0 & 84.8 & \underline{80.0} & 74.8 & 69.6 & 58.6 & 63.3 & 39.0 & \underline{66.4} \\
    \bottomrule
  \end{tabular}
\end{table}

\begin{table}[p]
  \centering
  \scriptsize
  \setlength{\tabcolsep}{3.2pt}
  \caption{Multilingual benchmarks for all 56 models we evaluated and \model, each averaged over the languages supported by \model (\cref{tab:benchmarks}). Bele.: Belebele; GP-NP, GP-P: Global PIQA non-parallel and parallel; T-HS: T-HellaSwag (\cref{app:t-hellaswag}); INCL.: INCLUDE; G-MMLU: Global MMLU. Best result in each group in bold, second best underlined.}
  \label{tab:app-multilingual}
  \begin{tabular}{@{}lrrrrrrrrrr@{}}
    \toprule
    \textbf{Model} & \textbf{XNLI} & \textbf{PAWS-X} & \textbf{Bele.} & \textbf{XCOPA} & \textbf{GP-NP} & \textbf{GP-P} & \textbf{T-HS} & \textbf{INCL.} & \textbf{G-MMLU} & \textbf{Avg.} \\
    \midrule
    \multicolumn{11}{@{}l}{\textit{Models with more than 4B active parameters}} \\
    Teuken-7B & 44.2 & 54.6 & 34.8 & 61.1 & 66.1 & 29.9 & 51.9 & 43.2 & 37.9 & 47.1 \\
    Mistral-7B-v0.3 & 42.6 & 59.3 & 47.3 & 58.8 & 62.7 & 28.4 & 48.5 & 47.0 & 46.3 & 49.0 \\
    Llama-3-8B & \underline{45.7} & \underline{59.8} & 50.4 & 65.7 & 66.2 & 32.5 & 54.9 & 54.2 & 52.1 & 53.5 \\
    Gemma-7B & 43.5 & 59.7 & 57.4 & 68.2 & \underline{69.0} & 33.0 & \underline{58.5} & 54.8 & 52.8 & 55.2 \\
    Llama-3.1-8B & \textbf{46.2} & \textbf{60.8} & 60.3 & 67.1 & 66.5 & 32.6 & 55.2 & 54.7 & 52.9 & 55.2 \\
    Apertus-v1.5-8B & 45.1 & 59.1 & 64.2 & \textbf{70.5} & \textbf{73.6} & \textbf{36.7} & \textbf{62.6} & \underline{61.8} & 59.0 & \textbf{59.2} \\
    Qwen2.5-7B & 44.0 & 58.8 & \underline{74.9} & 67.5 & 63.2 & 32.9 & 50.8 & 60.9 & \underline{62.9} & \underline{57.3} \\
    Qwen3-8B & 39.1 & 51.3 & \textbf{82.3} & \underline{69.4} & 67.1 & \underline{33.2} & 56.8 & \textbf{66.8} & \textbf{67.2} & \textbf{59.2} \\
    \midrule
    \multicolumn{11}{@{}l}{\textit{Models with 2B to 4B active parameters}} \\
    salamandra-2b & 40.9 & 51.5 & 24.3 & 56.3 & 60.0 & 28.1 & 42.2 & 25.7 & 25.3 & 39.4 \\
    Villanova2-2B & 42.2 & 53.6 & 43.0 & 58.0 & 59.7 & 26.6 & 41.6 & 40.7 & 42.0 & 45.3 \\
    tiny-aya & \textbf{46.2} & 57.9 & 54.0 & \underline{67.8} & \underline{69.3} & 30.7 & \underline{56.3} & 56.7 & 51.7 & 54.5 \\
    SmolLM3-3B & 41.0 & 55.5 & 47.2 & 62.5 & 59.8 & 29.2 & 45.5 & 47.0 & 50.5 & 48.7 \\
    Nemotron-Flash-3B & 38.2 & 56.0 & 38.8 & 55.0 & 52.3 & 26.0 & 35.3 & 36.0 & 40.2 & 42.0 \\
    granite-4.1-3b & 37.8 & 50.7 & 49.3 & 60.1 & 58.8 & 27.8 & 45.9 & 50.4 & 52.5 & 48.1 \\
    Ministral-3-3B & 44.0 & 56.5 & 69.5 & 62.4 & 64.0 & 32.6 & 53.5 & 58.3 & 58.1 & 55.4 \\
    Domyn-Edge & 42.0 & \textbf{63.3} & 64.9 & 61.8 & 60.3 & 28.5 & 46.6 & 51.4 & 49.5 & 52.0 \\
    Gemma-4-E4B & \underline{45.3} & 59.5 & 70.5 & \textbf{71.7} & \textbf{75.2} & \textbf{35.6} & \textbf{63.8} & \textbf{65.7} & 62.7 & \textbf{61.1} \\
    MiniCPM5-2B & 43.2 & 56.2 & 46.3 & 58.1 & 59.2 & 28.3 & 43.5 & 48.0 & 49.3 & 48.0 \\
    Qwen3.5-4B & 38.4 & 56.7 & \underline{76.2} & 66.1 & 68.3 & \underline{33.4} & 55.3 & \underline{64.0} & \textbf{63.7} & \underline{58.0} \\
    Hunyuan-4B & 40.7 & \underline{59.7} & 66.5 & 60.5 & 58.6 & 27.6 & 43.2 & 52.0 & 55.2 & 51.6 \\
    Qwen3-4B & 43.3 & 57.7 & \textbf{78.5} & 66.6 & 63.0 & 31.1 & 51.2 & 62.6 & \underline{63.2} & 57.5 \\
    \midrule
    \multicolumn{11}{@{}l}{\textit{Models with up to 2B active parameters}} \\
    Apertus-v1.1-0.5B & 41.6 & 48.7 & 23.7 & 57.6 & 58.7 & 23.9 & 36.4 & 25.5 & 24.5 & 37.8 \\
    Gemma-3-1B & 41.9 & 51.9 & 26.5 & 62.8 & 61.8 & 26.4 & 42.1 & 25.9 & 26.3 & 40.6 \\
    EuroLLM-1.7B & 41.0 & 50.3 & 25.2 & 57.8 & 62.0 & 28.2 & 44.2 & 25.8 & 26.3 & 40.1 \\
    EuroMoE-2.6B-A0.6B & 40.3 & 50.6 & 25.3 & 58.1 & 65.2 & 28.8 & 47.1 & 25.4 & 25.3 & 40.7 \\
    helium-1-2b & 40.3 & 53.4 & 35.7 & 55.7 & 59.3 & 26.7 & 43.2 & 38.4 & 36.6 & 43.3 \\
    Llama-3.2-1B & 41.3 & 52.6 & 29.7 & 59.0 & 56.2 & 24.9 & 37.4 & 28.7 & 27.7 & 39.7 \\
    Apertus-v1.1-1.5B & \underline{44.6} & 54.2 & 27.6 & \underline{63.5} & 63.9 & 28.7 & 47.3 & 37.6 & 36.1 & 44.8 \\
    Luciole-1B & 38.2 & 51.9 & 28.9 & 54.4 & 55.8 & 25.0 & 35.8 & 29.8 & 30.1 & 38.9 \\
    limite-1b-soup & 33.9 & 48.0 & 26.3 & 51.6 & 50.4 & 22.2 & 30.6 & 27.1 & 29.3 & 35.5 \\
    SmolLM2-1.7B & 37.6 & 54.9 & 30.4 & 53.6 & 51.4 & 24.3 & 33.3 & 32.4 & 33.5 & 39.1 \\
    OLMoE-1B-7B & 40.0 & 53.4 & 30.7 & 56.1 & 54.0 & 24.9 & 34.9 & 34.5 & 35.3 & 40.4 \\
    granite-swash-3b-a600m & 39.7 & 50.9 & 35.3 & 54.2 & 54.5 & 26.0 & 37.1 & 37.2 & 38.9 & 41.5 \\
    Nemotron-Flash-1B & 36.5 & 52.2 & 27.8 & 52.9 & 50.5 & 23.8 & 31.0 & 29.4 & 31.2 & 37.3 \\
    granite-3.1-3b-a800m & 39.1 & 53.7 & 30.4 & 57.6 & 55.7 & 25.4 & 37.3 & 33.7 & 35.3 & 40.9 \\
    Qwen3.5-0.8B & 38.3 & 53.9 & 49.1 & 57.1 & 56.2 & 25.0 & 37.1 & 40.0 & 42.0 & 44.3 \\
    LFM2.5-1.2B & 37.1 & 54.8 & 34.7 & 54.7 & 54.7 & 25.8 & 36.9 & 37.0 & 43.6 & 42.2 \\
    limite-1b & 34.8 & 48.8 & 27.9 & 51.7 & 50.6 & 23.4 & 29.5 & 29.4 & 30.9 & 36.3 \\
    Bielik-1.5B & 36.1 & 53.3 & 37.0 & 52.9 & 54.3 & 23.7 & 35.8 & 35.1 & 35.7 & 40.4 \\
    Falcon-H1-0.5B & 33.9 & 52.2 & 26.3 & 51.6 & 50.5 & 23.8 & 31.3 & 29.4 & 32.2 & 36.8 \\
    Gemma-4-E2B & 43.8 & 55.3 & 50.1 & \textbf{68.1} & \underline{70.4} & \underline{33.3} & \underline{57.7} & 52.5 & 48.7 & \underline{53.3} \\
    Qwen3-0.6B & 39.6 & 52.2 & 47.1 & 56.0 & 53.1 & 24.7 & 34.6 & 40.5 & 40.3 & 43.1 \\
    kanana-2-1.3b & 39.1 & 52.6 & 35.1 & 54.0 & 52.0 & 23.6 & 33.2 & 36.6 & 40.4 & 40.7 \\
    granite-4.0-h-1b & 37.0 & 53.0 & 36.4 & 56.2 & 55.1 & 24.9 & 39.0 & 41.9 & 43.6 & 43.0 \\
    moe-7b-1b & 39.2 & 54.6 & 34.4 & 57.8 & 55.1 & 26.8 & 39.1 & 38.1 & 40.3 & 42.8 \\
    granite-4.0-1b & 36.5 & 49.8 & 36.6 & 56.3 & 54.4 & 26.2 & 38.5 & 41.1 & 42.8 & 42.5 \\
    Qwen2.5-1.5B & 40.8 & 54.8 & 51.4 & 59.9 & 55.9 & 26.0 & 38.7 & 46.3 & 46.9 & 46.8 \\
    Falcon-H1-1.5B & 39.5 & \underline{58.4} & 34.1 & 54.5 & 56.4 & 26.5 & 37.5 & 39.0 & 44.8 & 43.4 \\
    Qwen3.5-2B & 40.2 & 55.7 & \textbf{61.9} & 61.6 & 61.7 & 28.6 & 44.9 & 45.3 & 50.1 & 50.0 \\
    LFM3-1.7B & 40.3 & 57.2 & 43.2 & 59.2 & 57.4 & 27.4 & 39.6 & 46.5 & 49.5 & 46.7 \\
    Trinity-Nano & 41.2 & \textbf{58.5} & 52.2 & 58.1 & 56.1 & 26.4 & 41.1 & 46.1 & 46.9 & 47.4 \\
    LFM2.5-8B-A1B & 41.1 & 54.4 & 40.5 & 62.4 & 61.8 & 29.0 & 46.6 & 48.2 & 53.7 & 48.6 \\
    Hunyuan-1.8B & 38.3 & \textbf{58.5} & 45.1 & 55.3 & 54.5 & 23.5 & 35.1 & 43.4 & 44.6 & 44.3 \\
    Falcon-H1-1.5B-Deep & 40.4 & 56.8 & 40.3 & 54.8 & 56.8 & 26.5 & 39.3 & 42.5 & 49.1 & 45.2 \\
    Qwen3-1.7B & 42.0 & 55.7 & \underline{61.8} & 62.2 & 58.1 & 28.4 & 42.0 & 51.4 & 50.8 & 50.3 \\
    granite-4.0-h-tiny & 42.8 & 57.0 & 51.1 & 61.5 & 63.1 & 30.3 & 51.3 & \underline{53.4} & \underline{55.4} & 51.8 \\
    \ourrow \model & \textbf{44.9} & 56.9 & 59.3 & \textbf{68.1} & \textbf{72.6} & \textbf{34.4} & \textbf{58.4} & \textbf{58.7} & \textbf{56.5} & \textbf{56.7} \\
    \bottomrule
  \end{tabular}
\end{table}

\begin{table}[p]
  \centering
  \scriptsize
  \setlength{\tabcolsep}{3.2pt}
  \caption{Multilingual benchmarks for all 56 models we evaluated and \model, each averaged over all its languages, including those \model does not support. Abbreviations as in \cref{tab:app-multilingual}. Best result in each group in bold, second best underlined.}
  \label{tab:app-multilingual-all}
  \begin{tabular}{@{}lrrrrrrrrrr@{}}
    \toprule
    \textbf{Model} & \textbf{XNLI} & \textbf{PAWS-X} & \textbf{Bele.} & \textbf{XCOPA} & \textbf{GP-NP} & \textbf{GP-P} & \textbf{T-HS} & \textbf{INCL.} & \textbf{G-MMLU} & \textbf{Avg.} \\
    \midrule
    \multicolumn{11}{@{}l}{\textit{Models with more than 4B active parameters}} \\
    Teuken-7B & 42.8 & 54.6 & 30.2 & 57.8 & 57.4 & 24.9 & 41.9 & 40.4 & 36.6 & 43.0 \\
    Mistral-7B-v0.3 & 41.5 & 59.3 & 36.1 & 56.7 & 56.5 & 24.9 & 40.5 & 44.0 & 44.2 & 44.9 \\
    Llama-3-8B & 44.4 & \underline{59.8} & 40.3 & 62.0 & 59.7 & 27.1 & 46.8 & 51.7 & 50.0 & 49.1 \\
    Gemma-7B & 42.9 & 59.7 & 43.1 & \underline{64.1} & \underline{61.3} & 27.5 & \underline{48.7} & 53.1 & 50.6 & 50.1 \\
    Llama-3.1-8B & \textbf{44.9} & \textbf{60.8} & 45.7 & 62.7 & 60.0 & 27.6 & 46.9 & 52.1 & 50.6 & 50.1 \\
    Apertus-v1.5-8B & \underline{44.6} & 59.1 & 48.2 & \textbf{65.2} & \textbf{64.3} & \textbf{30.1} & \textbf{52.7} & \underline{59.3} & 56.3 & \textbf{53.3} \\
    Qwen2.5-7B & 42.6 & 58.8 & \underline{53.9} & 62.0 & 56.9 & 27.7 & 41.9 & 56.0 & \underline{58.9} & 51.0 \\
    Qwen3-8B & 38.6 & 51.3 & \textbf{62.8} & 63.9 & 60.1 & \underline{28.1} & 47.4 & \textbf{63.3} & \textbf{63.2} & \underline{53.2} \\
    \midrule
    \multicolumn{11}{@{}l}{\textit{Models with 2B to 4B active parameters}} \\
    salamandra-2b & 40.0 & 51.5 & 23.8 & 55.0 & 55.2 & 24.3 & 36.9 & 26.1 & 25.2 & 37.6 \\
    Villanova2-2B & 41.0 & 53.6 & 33.0 & 55.7 & 54.9 & 23.6 & 36.1 & 38.0 & 40.2 & 41.8 \\
    tiny-aya & \textbf{45.6} & 57.9 & 41.2 & \underline{63.4} & \underline{61.1} & 26.2 & \underline{46.4} & 52.5 & 50.0 & 49.4 \\
    SmolLM3-3B & 40.0 & 55.5 & 36.4 & 58.4 & 56.2 & 25.2 & 38.8 & 43.8 & 47.8 & 44.7 \\
    Nemotron-Flash-3B & 37.7 & 56.0 & 32.0 & 54.1 & 52.8 & 24.0 & 33.6 & 34.8 & 38.6 & 40.4 \\
    granite-4.1-3b & 37.3 & 50.7 & 38.7 & 57.2 & 55.1 & 24.9 & 40.0 & 48.0 & 49.9 & 44.6 \\
    Ministral-3-3B & 42.7 & 56.5 & 49.4 & 59.6 & 58.2 & 27.0 & 44.8 & 55.8 & 54.8 & 49.9 \\
    Domyn-Edge & 41.0 & \textbf{63.3} & 47.3 & 58.4 & 55.9 & 25.2 & 40.0 & 48.2 & 47.0 & 47.4 \\
    Gemma-4-E4B & \underline{45.1} & 59.5 & \underline{56.7} & \textbf{67.0} & \textbf{66.3} & \textbf{29.8} & \textbf{56.3} & \textbf{63.5} & \underline{60.7} & \textbf{56.1} \\
    MiniCPM5-2B & 42.4 & 56.2 & 36.3 & 56.6 & 55.2 & 24.7 & 37.9 & 44.8 & 47.1 & 44.6 \\
    Qwen3.5-4B & 38.1 & 56.7 & 56.3 & 61.3 & 60.7 & \underline{27.8} & \underline{46.4} & \underline{60.2} & \textbf{60.8} & \underline{52.0} \\
    Hunyuan-4B & 39.9 & \underline{59.7} & 49.0 & 57.7 & 55.5 & 25.0 & 37.8 & 48.9 & 52.3 & 47.3 \\
    Qwen3-4B & 42.4 & 57.7 & \textbf{58.5} & 62.1 & 57.5 & 26.8 & 43.5 & 59.4 & 59.3 & 51.9 \\
    \midrule
    \multicolumn{11}{@{}l}{\textit{Models with up to 2B active parameters}} \\
    Apertus-v1.1-0.5B & 40.5 & 48.7 & 24.2 & 55.6 & 54.5 & 22.6 & 34.1 & 25.5 & 24.5 & 36.7 \\
    Gemma-3-1B & 41.0 & 51.9 & 26.5 & 59.5 & 56.9 & 24.1 & 37.7 & 25.8 & 26.2 & 38.9 \\
    EuroLLM-1.7B & 39.9 & 50.3 & 24.5 & 56.0 & 56.5 & 24.6 & 38.0 & 26.0 & 26.0 & 38.0 \\
    EuroMoE-2.6B-A0.6B & 39.4 & 50.6 & 24.9 & 55.9 & 58.2 & 25.0 & 39.0 & 25.9 & 25.2 & 38.3 \\
    helium-1-2b & 39.3 & 53.4 & 30.1 & 54.5 & 55.1 & 23.7 & 37.6 & 36.3 & 35.5 & 40.6 \\
    Llama-3.2-1B & 40.2 & 52.6 & 27.9 & 56.6 & 53.8 & 23.4 & 34.8 & 28.7 & 27.4 & 38.4 \\
    Apertus-v1.1-1.5B & \textbf{43.5} & 54.2 & 25.7 & 59.8 & 57.7 & 25.3 & 40.6 & 37.2 & 35.1 & 42.1 \\
    Luciole-1B & 37.6 & 51.9 & 27.9 & 53.6 & 54.0 & 22.9 & 34.2 & 29.2 & 29.5 & 37.9 \\
    limite-1b-soup & 33.9 & 48.0 & 26.2 & 51.7 & 51.3 & 22.4 & 30.5 & 26.9 & 29.2 & 35.6 \\
    SmolLM2-1.7B & 37.2 & 54.9 & 28.7 & 53.3 & 52.2 & 22.7 & 32.8 & 32.1 & 32.9 & 38.5 \\
    OLMoE-1B-7B & 39.3 & 53.4 & 28.5 & 54.8 & 53.3 & 23.0 & 33.1 & 33.5 & 34.4 & 39.3 \\
    granite-swash-3b-a600m & 38.9 & 50.9 & 30.0 & 53.4 & 53.4 & 23.9 & 34.3 & 35.6 & 37.4 & 39.7 \\
    Nemotron-Flash-1B & 36.0 & 52.2 & 26.8 & 53.0 & 51.8 & 22.7 & 31.2 & 29.3 & 30.6 & 37.1 \\
    granite-3.1-3b-a800m & 38.4 & 53.7 & 28.4 & 55.7 & 53.5 & 23.2 & 34.6 & 32.7 & 34.1 & 39.4 \\
    Qwen3.5-0.8B & 37.6 & 53.9 & 37.1 & 55.6 & 54.2 & 23.4 & 34.5 & 37.8 & 40.5 & 41.6 \\
    LFM2.5-1.2B & 36.6 & 54.8 & 29.1 & 54.2 & 53.2 & 23.6 & 34.5 & 34.7 & 41.3 & 40.2 \\
    limite-1b & 34.7 & 48.8 & 27.3 & 52.3 & 51.2 & 22.5 & 30.2 & 29.2 & 30.4 & 36.3 \\
    Bielik-1.5B & 35.8 & 53.3 & 30.5 & 53.0 & 53.6 & 22.6 & 34.3 & 33.9 & 34.6 & 39.1 \\
    Falcon-H1-0.5B & 33.9 & 52.2 & 25.1 & 51.7 & 51.9 & 22.8 & 31.9 & 29.8 & 31.8 & 36.8 \\
    Gemma-4-E2B & \textbf{43.5} & 55.3 & 40.1 & \textbf{64.1} & \textbf{62.2} & \textbf{28.3} & \textbf{49.8} & 50.6 & 47.2 & \underline{49.0} \\
    Qwen3-0.6B & 38.9 & 52.2 & 36.8 & 55.2 & 52.1 & 23.4 & 33.8 & 38.5 & 38.7 & 41.1 \\
    kanana-2-1.3b & 38.2 & 52.6 & 30.5 & 53.8 & 52.6 & 22.7 & 32.9 & 35.1 & 38.9 & 39.7 \\
    granite-4.0-h-1b & 36.7 & 53.0 & 31.3 & 54.6 & 53.7 & 23.4 & 35.7 & 40.0 & 41.9 & 41.1 \\
    moe-7b-1b & 38.5 & 54.6 & 30.2 & 55.9 & 53.7 & 23.9 & 35.7 & 36.5 & 38.8 & 40.9 \\
    granite-4.0-1b & 36.1 & 49.8 & 32.2 & 54.7 & 53.3 & 24.1 & 35.2 & 39.4 & 41.1 & 40.7 \\
    Qwen2.5-1.5B & 39.9 & 54.8 & 38.2 & 57.3 & 53.5 & 23.8 & 35.5 & 42.9 & 44.5 & 43.4 \\
    Falcon-H1-1.5B & 38.9 & \underline{58.4} & 30.4 & 53.7 & 54.7 & 24.2 & 34.7 & 36.9 & 42.6 & 41.6 \\
    Qwen3.5-2B & 39.4 & 55.7 & \underline{44.5} & 57.9 & 56.6 & 24.9 & 39.1 & 43.1 & 47.9 & 45.5 \\
    LFM3-1.7B & 39.5 & 57.2 & 34.4 & 57.4 & 55.1 & 24.7 & 36.5 & 43.9 & 46.7 & 43.9 \\
    Trinity-Nano & 40.4 & \textbf{58.5} & 38.3 & 55.9 & 54.2 & 23.8 & 36.6 & 43.5 & 45.0 & 44.0 \\
    LFM2.5-8B-A1B & 40.2 & 54.4 & 31.5 & 58.6 & 56.9 & 25.1 & 39.2 & 44.5 & \underline{50.3} & 44.5 \\
    Hunyuan-1.8B & 37.8 & \textbf{58.5} & 34.1 & 54.6 & 53.4 & 22.4 & 33.1 & 41.4 & 42.8 & 42.0 \\
    Falcon-H1-1.5B-Deep & 39.8 & 56.8 & 32.2 & 53.9 & 54.7 & 24.2 & 35.5 & 39.7 & 46.6 & 42.6 \\
    Qwen3-1.7B & 40.9 & 55.7 & \textbf{45.9} & 58.6 & 54.5 & 25.6 & 37.6 & 48.7 & 48.2 & 46.2 \\
    granite-4.0-h-tiny & 41.8 & 57.0 & 40.1 & 58.7 & 58.0 & 26.0 & 43.9 & \underline{51.6} & \textbf{52.9} & 47.8 \\
    \ourrow \model & \underline{43.4} & 56.9 & 40.7 & \underline{62.5} & \underline{61.2} & \underline{27.8} & \underline{45.5} & \textbf{53.5} & \textbf{52.9} & \textbf{49.4} \\
    \bottomrule
  \end{tabular}
\end{table}

\begin{table}[p]
  \centering
  \scriptsize
  \setlength{\tabcolsep}{3.2pt}
  \caption{Math and code benchmarks for all 56 models we evaluated and \model. Avg.: mean of the math and code averages. CRUX-I, CRUX-O: CRUXEval input and output prediction, answered directly. Best result in each group in bold, second best underlined.}
  \label{tab:app-math-code}
  \begin{tabular}{@{}lrrrrrrrr@{}}
    \toprule
    & \multicolumn{3}{c}{\textbf{Math}} & \multicolumn{4}{c}{\textbf{Code}} & \\
    \cmidrule(lr){2-4}\cmidrule(lr){5-8}
    \textbf{Model} & \textbf{GSM8K} & \textbf{MATH} & \textbf{MGSM} & \textbf{MBPP} & \textbf{MBPP+} & \textbf{CRUX-I} & \textbf{CRUX-O} & \textbf{Avg.} \\
    \midrule
    \multicolumn{9}{@{}l}{\textit{Models with more than 4B active parameters}} \\
    Teuken-7B & 10.9 & 4.4 & 7.0 & 13.4 & 14.0 & 22.4 & 26.6 & 13.3 \\
    Mistral-7B-v0.3 & 36.4 & 14.0 & 23.5 & 38.0 & 41.5 & 36.0 & 34.1 & 31.0 \\
    Llama-3-8B & 51.0 & 19.0 & 37.2 & 49.8 & 53.4 & 32.9 & 37.3 & 39.6 \\
    Gemma-7B & 51.9 & 24.8 & 40.2 & 45.2 & 47.1 & 33.3 & 36.5 & 39.7 \\
    Llama-3.1-8B & 50.6 & 21.2 & 37.8 & 49.2 & 52.7 & 35.1 & 37.4 & 40.1 \\
    Apertus-v1.5-8B & 67.1 & 30.2 & 54.6 & 50.0 & 55.8 & 33.6 & 37.8 & 47.5 \\
    Qwen2.5-7B & \underline{83.9} & \underline{49.9} & \underline{60.8} & \underline{63.6} & \underline{68.3} & \underline{44.3} & \underline{45.8} & \underline{60.2} \\
    Qwen3-8B & \textbf{86.7} & \textbf{55.3} & \textbf{71.9} & \textbf{69.6} & \textbf{70.4} & \textbf{52.4} & \textbf{51.6} & \textbf{66.1} \\
    \midrule
    \multicolumn{9}{@{}l}{\textit{Models with 2B to 4B active parameters}} \\
    salamandra-2b & 2.4 & 2.6 & 3.6 & 5.8 & 7.4 & 13.6 & 14.9 & 6.7 \\
    Villanova2-2B & 38.7 & 26.3 & 21.5 & 35.0 & 41.0 & 32.3 & 27.4 & 31.4 \\
    tiny-aya & 57.4 & 23.6 & 35.2 & 27.8 & 32.5 & 27.6 & 29.5 & 34.1 \\
    SmolLM3-3B & 68.7 & 42.5 & 41.9 & 50.0 & 54.5 & 34.9 & 35.1 & 47.3 \\
    Nemotron-Flash-3B & 65.4 & \underline{67.7} & 46.1 & 52.2 & 58.7 & 32.4 & 34.6 & 52.1 \\
    granite-4.1-3b & 72.2 & 41.9 & 47.3 & 51.4 & 51.6 & 44.9 & 42.5 & 50.7 \\
    Ministral-3-3B & 64.7 & 35.8 & 49.2 & 44.0 & 50.0 & 38.3 & 36.9 & 46.1 \\
    Domyn-Edge & 74.6 & 40.6 & 47.0 & 53.8 & 55.8 & 42.5 & 37.3 & 50.7 \\
    Gemma-4-E4B & 56.8 & 24.0 & 52.5 & 51.2 & 54.2 & 38.1 & 40.4 & 45.2 \\
    MiniCPM5-2B & 73.5 & 66.2 & 58.3 & 58.6 & \underline{67.2} & 38.8 & 38.8 & 58.4 \\
    Qwen3.5-4B & \underline{82.4} & 54.2 & 61.6 & 52.6 & 58.7 & \underline{50.0} & \underline{48.0} & 59.2 \\
    Hunyuan-4B & 78.0 & \textbf{74.8} & \underline{63.1} & \underline{58.8} & 62.4 & \textbf{56.3} & \textbf{57.0} & \textbf{65.3} \\
    Qwen3-4B & \textbf{87.8} & 52.8 & \textbf{66.9} & \textbf{67.0} & \textbf{68.0} & 45.1 & \underline{48.0} & \underline{63.1} \\
    \midrule
    \multicolumn{9}{@{}l}{\textit{Models with up to 2B active parameters}} \\
    Apertus-v1.1-0.5B & 1.1 & 3.0 & 2.0 & 5.0 & 6.1 & 18.0 & 15.5 & 6.6 \\
    Gemma-3-1B & 1.9 & 2.7 & 2.1 & 8.6 & 10.9 & 12.4 & 18.5 & 7.4 \\
    EuroLLM-1.7B & 5.3 & 3.5 & 2.8 & 12.2 & 15.9 & 23.5 & 20.3 & 10.9 \\
    EuroMoE-2.6B-A0.6B & 9.0 & 5.9 & 3.7 & 19.6 & 21.7 & 24.8 & 18.3 & 13.6 \\
    helium-1-2b & 9.0 & 4.1 & 5.1 & 4.2 & 6.4 & 26.1 & 17.5 & 9.8 \\
    Llama-3.2-1B & 6.5 & 4.3 & 4.2 & 26.4 & 32.8 & 28.5 & 24.4 & 16.5 \\
    Apertus-v1.1-1.5B & 7.7 & 4.4 & 5.6 & 15.4 & 21.7 & 25.0 & 20.1 & 13.2 \\
    Luciole-1B & 15.7 & 9.6 & 7.6 & 22.6 & 26.7 & 25.9 & 20.9 & 17.5 \\
    limite-1b-soup & \underline{76.6} & \underline{53.8} & 11.2 & 14.8 & 14.8 & 22.0 & 24.1 & 33.1 \\
    SmolLM2-1.7B & 29.6 & 12.5 & 10.7 & 33.6 & 39.2 & 27.1 & 27.4 & 24.7 \\
    OLMoE-1B-7B & 51.0 & 17.4 & 13.4 & 21.6 & 23.0 & 19.6 & 20.0 & 24.2 \\
    granite-swash-3b-a600m & 36.7 & 25.8 & 18.2 & 26.6 & 32.3 & 18.6 & 26.8 & 26.5 \\
    Nemotron-Flash-1B & 32.2 & 44.9 & 22.0 & 34.8 & 45.2 & 25.8 & 23.0 & 32.6 \\
    granite-3.1-3b-a800m & 41.8 & 19.3 & 15.8 & 35.0 & 43.9 & 24.3 & 27.1 & 29.1 \\
    Qwen3.5-0.8B & 40.3 & 25.2 & 25.6 & 26.0 & 29.9 & 31.6 & 25.3 & 29.3 \\
    LFM2.5-1.2B & 55.1 & 32.1 & 26.2 & 22.4 & 32.3 & 17.4 & 23.5 & 30.9 \\
    limite-1b & \textbf{78.6} & 53.4 & 21.1 & 27.4 & 29.4 & 26.4 & 27.0 & 39.3 \\
    Bielik-1.5B & 36.7 & 28.7 & 15.2 & 41.8 & 47.6 & 32.4 & 29.8 & 32.4 \\
    Falcon-H1-0.5B & 63.4 & 39.2 & 15.0 & 36.0 & 40.7 & 29.4 & 25.5 & 36.1 \\
    Gemma-4-E2B & 22.8 & 10.7 & 20.6 & 36.6 & 40.0 & 27.5 & 27.1 & 25.4 \\
    Qwen3-0.6B & 49.7 & 30.4 & 31.4 & 39.2 & 45.5 & 27.4 & 27.3 & 36.0 \\
    kanana-2-1.3b & 49.5 & 31.4 & 25.7 & 47.6 & 51.6 & 27.9 & 32.5 & 37.7 \\
    granite-4.0-h-1b & 49.4 & 32.5 & 31.4 & 40.2 & 47.9 & 34.3 & 35.6 & 38.6 \\
    moe-7b-1b & 54.9 & 25.1 & 22.2 & 45.2 & 53.2 & 27.1 & 32.5 & 36.8 \\
    granite-4.0-1b & 59.4 & 35.5 & 35.0 & 46.4 & 49.7 & \textbf{39.6} & 36.1 & 43.1 \\
    Qwen2.5-1.5B & 61.3 & 31.3 & 36.4 & 43.4 & 50.3 & 29.9 & 28.8 & 40.5 \\
    Falcon-H1-1.5B & 55.6 & 46.6 & 36.3 & 49.4 & 55.3 & 31.5 & 34.1 & 44.4 \\
    Qwen3.5-2B & 60.1 & 35.7 & 41.1 & 33.6 & 42.1 & \underline{37.6} & 33.4 & 41.1 \\
    LFM3-1.7B & 66.9 & 43.6 & 46.2 & 39.0 & 46.3 & 35.1 & 31.4 & 45.1 \\
    Trinity-Nano & 58.5 & 41.8 & 34.8 & 47.0 & 52.1 & 32.9 & 35.9 & 43.5 \\
    LFM2.5-8B-A1B & 67.6 & 45.8 & 43.3 & 40.4 & 48.4 & 28.4 & 29.1 & 44.4 \\
    Hunyuan-1.8B & 71.6 & 42.3 & 42.1 & \textbf{58.8} & \textbf{62.2} & \textbf{39.6} & \underline{38.4} & 50.9 \\
    Falcon-H1-1.5B-Deep & 72.3 & 49.2 & 44.3 & 56.8 & 58.5 & 35.8 & 37.8 & \underline{51.2} \\
    Qwen3-1.7B & 71.8 & 41.0 & \underline{50.9} & 55.4 & \underline{59.8} & 35.9 & 35.1 & 50.6 \\
    granite-4.0-h-tiny & 59.1 & 45.4 & 43.6 & \underline{58.4} & 58.5 & 36.6 & \textbf{40.0} & 48.9 \\
    \ourrow \model & 67.2 & \textbf{60.9} & \textbf{51.7} & 53.8 & 58.2 & 34.4 & 33.9 & \textbf{52.5} \\
    \bottomrule
  \end{tabular}
\end{table}

\clearpage
\subsection{Per-Language Results}
\label{app:per-language}
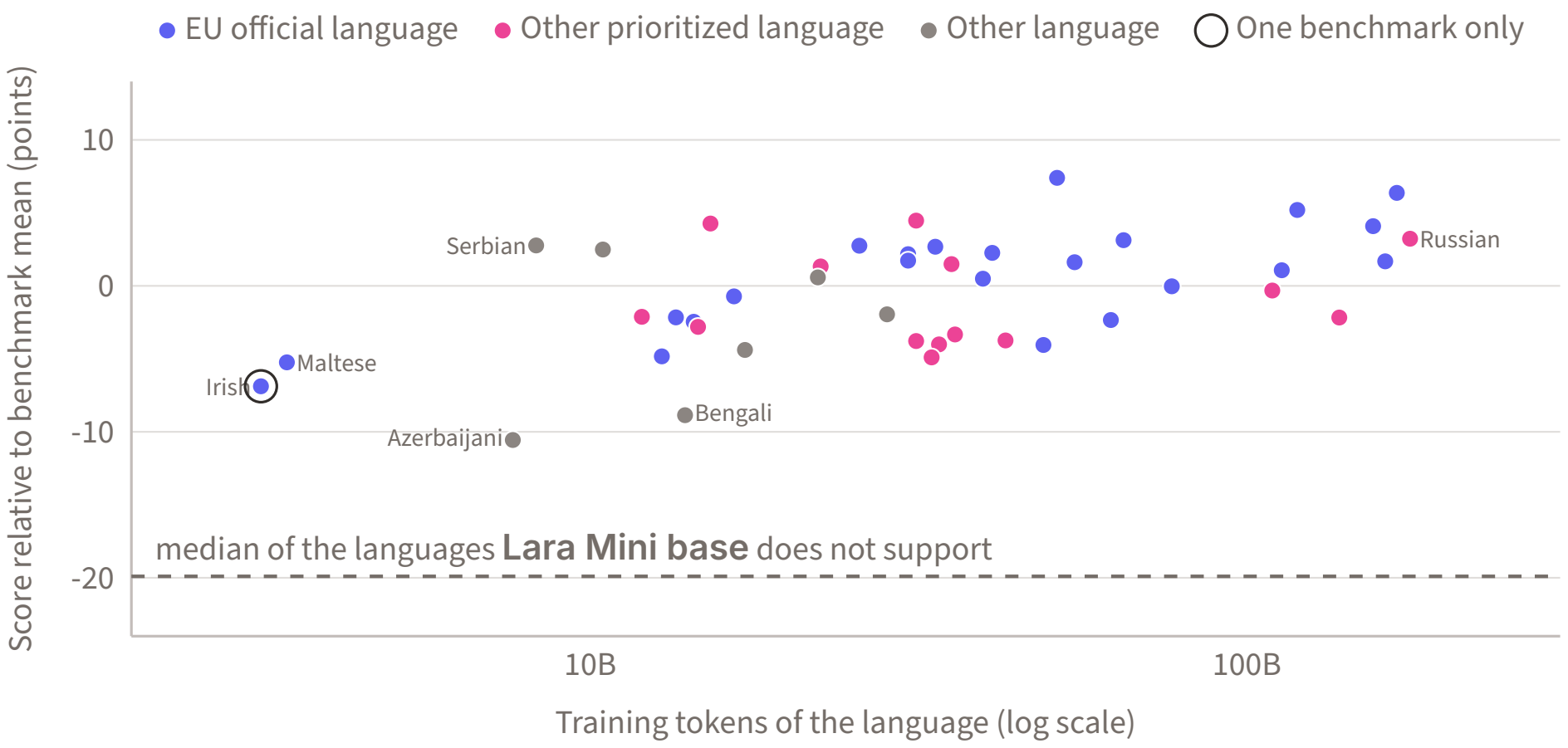
\begin{figure}[H]
  \centering
  \begin{tikzpicture}
  \begin{axis}[lrchart,
      width=0.80\linewidth, height=6.0cm,
      xmode=log, xmin=2, xmax=300,
      xtick={10,100}, xticklabels={10B,100B},
      ymin=-24, ymax=14, ytick={-20,-10,0,10},
      ymajorgrids,
      xlabel={Training tokens of the language (log scale)},
      ylabel={Score relative to benchmark mean (points)},
      table/col sep=comma,
      legend style={at={(0.5,1.03)}, anchor=south, legend columns=4,
        /tikz/every even column/.append style={column sep=6pt}},
    ]
    \draw[chartsecondary, dashed, line width=0.8pt] (axis cs:2,-19.9) -- (axis cs:300,-19.9);
    \node[anchor=south west, font=\sffamily\scriptsize, text=chartsecondary, inner sep=2pt]
      at (axis cs:2.1,-19.9) {median of the languages \model does not support};
    \addplot[only marks, mark=*, mark size=2.0pt, mark options={draw=white, line width=0.4pt, fill=brandgradstart}]
      table[x=tokens, y=eu] {figures/data/per-language.csv};
    \addlegendentry{EU official language}
    \addplot[only marks, mark=*, mark size=2.0pt, mark options={draw=white, line width=0.4pt, fill=brandgradend}]
      table[x=tokens, y=t12] {figures/data/per-language.csv};
    \addlegendentry{Other prioritized language}
    \addplot[only marks, mark=*, mark size=2.0pt, mark options={draw=white, line width=0.4pt, fill=chartmuted}]
      table[x=tokens, y=oth] {figures/data/per-language.csv};
    \addlegendentry{Other language}
    \addplot[only marks, mark=o, mark size=3.6pt, mark options={draw=chartink, line width=0.6pt}]
      table[x=tokens, y=one] {figures/data/per-language.csv};
    \addlegendentry{One benchmark only}
    \node[anchor=east, font=\sffamily\tiny, text=chartsecondary, inner sep=2pt] at (axis cs:7.62,-10.56) {Azerbaijani};
    \node[anchor=west, font=\sffamily\tiny, text=chartsecondary, inner sep=2pt] at (axis cs:13.94,-8.86) {Bengali};
    \node[anchor=east, font=\sffamily\tiny, text=chartsecondary, inner sep=2pt] at (axis cs:3.15,-6.88) {Irish}; %
    \node[anchor=west, font=\sffamily\tiny, text=chartsecondary, inner sep=2pt] at (axis cs:3.45,-5.24) {Maltese};
    \node[anchor=east, font=\sffamily\tiny, text=chartsecondary, inner sep=2pt] at (axis cs:8.27,2.77) {Serbian}; %
    \node[anchor=west, font=\sffamily\tiny, text=chartsecondary, inner sep=2pt] at (axis cs:177.24,3.23) {Russian};
  \end{axis}
\end{tikzpicture}
  \caption{Score of \model in each supported language against its training tokens in that language (log scale). Scores are relative to the mean of each benchmark across the non-English-supported languages and are averaged across the multilingual benchmarks that cover the language. The dashed line is the median of the languages and varieties \model does not support. The ring marks Irish, which only T-HellaSwag covers. Every supported language scores above the median of the languages \model does not support.}
  \label{fig:per-language}
\end{figure}

\Cref{fig:per-language} plots the score of \model in each supported language against its training tokens in that language.
Each point is the score in the language minus the benchmark mean across the non-English-supported languages, averaged over multilingual benchmarks that cover the language.
Subtracting the benchmark mean removes the differences in chance level and difficulty between benchmarks.
The median score of the languages \model does not support is 20 points below that of its supported languages.
Among the supported languages, the score grows only moderately with the number of training tokens (Spearman correlation 0.4).
Of the three classes of language in the figure, the official EU languages, which the language weights favor (\cref{sec:data-mixture}), have the highest median.
Serbian, with 8B training tokens, scores close to Russian, which has 21 times as many.
The weakest supported languages are Azerbaijani and Bengali, followed by Irish and Maltese, the two languages with the fewest training tokens.

\Cref{tab:app-per-language} gives the score of \model on every multilingual benchmark in each supported language.
For each language, it compares the mean of \model with the best mean over the same benchmarks among the 35 other models with up to 2B active parameters.
\model is ahead in 41 of the 44 languages.
It trails in Simplified Chinese (Qwen3-1.7B, 2.7 points) and Bosnian (Gemma-4-E2B, 1.3 points), and is level with kanana-2-1.3b in Korean.

\begin{table}[H]
  \centering
  \footnotesize
  \setlength{\tabcolsep}{3pt}
  \caption{Scores of \model per supported language on each multilingual benchmark that covers it (--: not covered; country variants averaged into their language), its mean over these benchmarks (Avg.), and the best mean over the same benchmarks among the 35 other models with up to 2B active parameters (Peer, with the model that holds it: $^a$EuroMoE-2.6B-A0.6B, $^b$Gemma-4-E2B, $^c$LFM2.5-8B-A1B, $^d$Qwen3-1.7B, $^e$Qwen3.5-2B, $^f$granite-4.0-h-tiny, $^g$kanana-2-1.3b). Languages in order of training tokens; bold: \model ahead. Abbreviations as in \cref{tab:app-multilingual}. \model is ahead in 41 of the 44 languages.}
  \label{tab:app-per-language}
  \begin{tabular}{@{}lrrrrrrrrrrr@{}}
    \toprule
    \textbf{Language} & \textbf{Bele.} & \textbf{G-MMLU} & \textbf{INCL.} & \textbf{XNLI} & \textbf{PAWS-X} & \textbf{XCOPA} & \textbf{GP-NP} & \textbf{GP-P} & \textbf{T-HS} & \textbf{Avg.} & \textbf{Peer} \\
    \midrule
    Russian & 67.1 & -- & 54.4 & 48.6 & -- & -- & 76.0 & 31.1 & 63.5 & \textbf{56.8} & 56.2$^b$ \\
    Spanish & 64.3 & 59.6 & 64.0 & 45.9 & 63.8 & -- & 87.3 & 41.4 & 62.0 & \textbf{61.0} & 58.8$^f$ \\
    German & 62.9 & 58.9 & 51.8 & 47.9 & 59.2 & -- & 77.0 & 36.9 & 56.0 & \textbf{56.3} & 54.7$^f$ \\
    French & 63.2 & 58.8 & 60.9 & 49.8 & 59.3 & -- & 75.0 & 43.7 & 60.8 & \textbf{58.9} & 55.7$^b$ \\
    Chinese (Simpl.) & 65.7 & 57.3 & 51.2 & 36.7 & 53.6 & 70.6 & 58.0 & 35.0 & 57.5 & 53.9 & 56.6$^d$ \\
    Italian & 67.2 & 59.4 & 71.2 & -- & -- & 71.8 & 77.0 & 39.8 & 58.0 & \textbf{63.5} & 60.0$^f$ \\
    Portuguese & 68.6 & 60.0 & 53.9 & -- & -- & -- & 70.5 & 34.5 & 59.0 & \textbf{57.7} & 56.7$^f$ \\
    Japanese & 54.3 & 55.7 & 63.3 & -- & 51.6 & -- & 79.0 & 40.8 & 60.5 & \textbf{57.9} & 56.1$^f$ \\
    Polish & 60.6 & -- & 54.4 & -- & -- & -- & 75.0 & 36.9 & 56.5 & \textbf{56.7} & 52.7$^b$ \\
    Dutch & 65.1 & -- & 62.1 & -- & -- & -- & 74.0 & 35.0 & 63.0 & \textbf{59.8} & 57.3$^b$ \\
    Greek & 58.1 & -- & 54.5 & 45.2 & -- & -- & 60.0 & 35.0 & 61.5 & \textbf{52.4} & 49.1$^b$ \\
    Czech & 62.2 & -- & -- & -- & -- & -- & 79.0 & 32.0 & 58.0 & \textbf{57.8} & 52.5$^f$ \\
    Swedish & 68.0 & -- & -- & -- & -- & -- & 82.0 & 36.9 & 67.5 & \textbf{63.6} & 60.1$^b$ \\
    Hungarian & 60.1 & -- & 51.6 & -- & -- & -- & 63.0 & 34.0 & 54.5 & \textbf{52.7} & 48.5$^b$ \\
    Korean & 58.7 & 54.2 & 57.6 & -- & 54.2 & -- & 58.0 & 30.1 & 58.0 & 53.0 & 53.0$^g$ \\
    Danish & 64.0 & -- & -- & -- & -- & -- & 66.0 & 38.8 & 65.0 & \textbf{58.5} & 53.5$^b$ \\
    Finnish & 61.2 & -- & 45.9 & -- & -- & -- & 80.0 & 39.8 & 59.0 & \textbf{57.2} & 49.7$^b$ \\
    Turkish & 58.1 & -- & 55.3 & 48.0 & -- & 63.0 & 74.0 & 25.2 & 49.5 & \textbf{53.3} & 51.6$^b$ \\
    Indonesian & 56.3 & 57.8 & 62.0 & -- & -- & 71.8 & 76.0 & 33.0 & 61.5 & \textbf{59.8} & 57.4$^e$ \\
    Chinese (Trad.) & 60.8 & -- & -- & -- & -- & -- & 56.0 & 34.0 & 58.0 & \textbf{52.2} & 52.1$^d$ \\
    Bulgarian & 62.2 & -- & 66.9 & 47.2 & -- & -- & 78.0 & 31.1 & 59.0 & \textbf{57.4} & 52.8$^b$ \\
    Arabic & 60.2 & 54.5 & 57.1 & 33.5 & -- & -- & 61.0 & 27.2 & 57.0 & \textbf{50.1} & 48.0$^c$ \\
    Ukrainian & 57.6 & -- & 65.8 & -- & -- & -- & 86.0 & 34.0 & 62.5 & \textbf{61.2} & 57.7$^b$ \\
    Thai & 56.8 & -- & -- & 45.1 & -- & 61.6 & 70.0 & 33.0 & 52.5 & \textbf{53.2} & 52.5$^b$ \\
    Romanian & 63.1 & -- & -- & -- & -- & -- & 80.0 & 29.1 & 59.5 & \textbf{57.9} & 54.3$^b$ \\
    Slovak & 60.3 & -- & -- & -- & -- & -- & 85.0 & 28.2 & 60.0 & \textbf{58.4} & 54.0$^b$ \\
    Persian & 58.8 & -- & 45.6 & -- & -- & -- & 76.0 & 37.9 & 55.5 & \textbf{54.8} & 51.3$^b$ \\
    Croatian & 60.1 & -- & 71.6 & -- & -- & -- & 76.0 & 33.0 & 56.5 & \textbf{59.5} & 56.0$^b$ \\
    Norwegian & 62.7 & -- & -- & -- & -- & -- & 69.0 & 35.0 & 63.5 & \textbf{57.5} & 51.2$^b$ \\
    Vietnamese & 61.6 & -- & 58.6 & 44.3 & -- & 72.6 & 71.0 & 35.0 & 57.5 & \textbf{57.2} & 53.6$^e$ \\
    Hindi & 48.7 & 51.3 & 56.1 & 45.9 & -- & -- & 65.0 & 32.0 & 55.0 & \textbf{50.6} & 48.1$^b$ \\
    Lithuanian & 53.0 & -- & 62.9 & -- & -- & -- & 69.0 & 35.0 & 60.0 & \textbf{56.0} & 49.5$^b$ \\
    Catalan & 63.0 & -- & -- & -- & -- & -- & 80.0 & 36.9 & 62.0 & \textbf{60.5} & 55.1$^b$ \\
    Hebrew & 53.0 & -- & 66.0 & -- & -- & -- & 55.0 & 35.9 & 59.5 & \textbf{53.9} & 53.8$^b$ \\
    Slovenian & 56.9 & -- & -- & -- & -- & -- & 66.0 & 31.1 & 61.0 & \textbf{53.7} & 48.7$^b$ \\
    Bengali & 46.4 & 50.0 & 50.4 & -- & -- & -- & 69.0 & 29.1 & 51.5 & \textbf{49.4} & 46.0$^b$ \\
    Estonian & 53.4 & -- & 67.4 & -- & -- & 65.2 & 58.0 & 32.0 & 62.5 & \textbf{56.4} & 52.5$^b$ \\
    Latvian & 54.6 & -- & -- & -- & -- & -- & -- & -- & 53.5 & \textbf{54.0} & 49.3$^b$ \\
    Malay & 58.1 & -- & 60.3 & -- & -- & -- & 63.0 & 33.0 & 58.5 & \textbf{54.6} & 48.3$^b$ \\
    Bosnian & -- & -- & -- & -- & -- & -- & 78.0 & -- & 58.0 & 68.0 & 69.3$^b$ \\
    Serbian & 58.7 & -- & 65.8 & -- & -- & -- & 88.0 & 29.1 & 55.8 & \textbf{59.5} & 54.9$^b$ \\
    Azerbaijani & 43.2 & -- & 52.2 & -- & -- & -- & 63.0 & 23.3 & 49.0 & \textbf{46.1} & 42.1$^b$ \\
    Maltese & 53.2 & -- & -- & -- & -- & -- & -- & -- & 54.0 & \textbf{53.6} & 43.4$^b$ \\
    Irish & -- & -- & -- & -- & -- & -- & -- & -- & 51.5 & \textbf{51.5} & 43.5$^a$ \\
    \bottomrule
  \end{tabular}
\end{table}

\subsection{Long-Context Evaluation}
\label{app:lc-eval}
Some tasks of RULER, ONERULER and HELMET assume a model that answers and then stops, while a base model keeps writing after its answer.
\Cref{tab:app-lc-eval} lists what we changed and why.
Every model in \cref{tab:ce-results} is scored with the same prompts, samples and code.
Among them, granite-4.0-h-tiny scores low on RULER and ONERULER because it copies only two to five digits of each 7-digit number it has to retrieve.

\begin{table}[H]
  \centering
  \small
  \caption{Changes to the long-context benchmarks for base models.}
  \label{tab:app-lc-eval}
  \begin{tabularx}{\linewidth}{@{}l>{\raggedright\arraybackslash}p{6.2cm}L@{}}
    \toprule
    \textbf{Benchmark} & \textbf{What we changed} & \textbf{Why} \\
    \midrule
    RULER    & Fixed samples: 100 per task and length, generated once, with lengths counted in each model's tokens & The harness draws new random samples at every run; fixed samples give every model the same inputs \\
    \midrule
    ONERULER & No answer template at the end of the prompt; instead, the prompt ends with the start of the answer, the needle sentence up to the number (``The special magic number for \ldots{} is:'') & Base models copy the answer template instead of answering \\
    \addlinespace[3pt]
             & In the task with four numbers for one key, the start of the answer states how many there are (``The 4 special magic numbers for \ldots{} are:'') & Without the count, models often stop after one to three numbers, so the score measures when they stop rather than what they find \\
    \addlinespace[3pt]
             & An answer is correct if it contains the right numbers, whatever text follows them & The original scoring gives zero to any answer that contains the word for ``none'', which a base model often copies from the question after a correct answer \\
    \addlinespace[3pt]
             & The task whose correct answer is ``none'' (no needle has the key asked for) is left out & Without examples in the prompt, the base models we tested did not answer ``none''; with examples, the score depends on how closely a model follows them rather than on what it retrieves, and a model can then answer ``none'' even when the needle is there \\
    \addlinespace[3pt]
             & The two common-word tasks are left out & They score close to zero for every model from 16k tokens \\
    \addlinespace[3pt]
             & Answers of up to 64 tokens; 20 samples per language and task up to 32k tokens, 10 at 64k and 128k & Room for four 7-digit numbers with tokenizers that split numbers into single digits; fewer samples at the longest lengths to limit the cost \\
    \midrule
    HELMET   & Summarization, citation and long-document question answering left out & Base models score on summarization by copying the example summary given in the prompt; citation needs an entailment model that we do not run; long-document question answering has weak automatic metrics and 100 questions per dataset \\
    \addlinespace[3pt]
             & At 128k, retrieval-augmented generation prompts that are too long are shortened by dropping their lowest-ranked distractor passages, never a passage with the answer; every model gets the same prompts & HELMET instead cuts the end of the prompt by an amount that depends on the tokenizer, which can remove the passage with the answer \\
    \addlinespace[3pt]
             & In-context learning on four of its five datasets & The fifth, NLU, is no longer publicly available \\
    \bottomrule
  \end{tabularx}
\end{table}

\subsection{Benchmark Contamination}
\label{sec:contamination}
Here we test whether contamination inflates the scores of \model.
HumanEval is excluded from the suite because nearly all of its items, including solutions, appear in public datasets in our pool (\cref{sec:data-decontam}).
Every other benchmark is used as released, with all its items.
On MMLU and MBPP, two tests verify that items from the training pool do not inflate \ model's score, against three references from its group: Llama-3.2-1B, granite-4.0-h-tiny, and Qwen3-1.7B.
We divide benchmark items into clean and potentially contaminated subsets based on whether our scan identifies overlap with the training pool.
This partition is applied consistently to all four models, although the training data of the reference models may differ from ours.
Neither test shows signs of memorization: \ model's scores are consistent with generalization.

\paragraph{Accuracy.}
A model that had memorized the contaminated items would score higher on them than on the clean ones, relative to the references (\cref{tab:contam-acc}).
\model does not: its difference is $-0.2$ points on MMLU (references: $-0.7$ to $+2.0$) and $-9.2$ on MBPP ($-15.6$ to $-8.4$).
Both differences fall within the reference-model ranges, providing no clear evidence of an unusual accuracy advantage on potentially contaminated items.

\begin{table}[!h]
  \centering
  \footnotesize
  \caption{Accuracy (\%; pass@1 for MBPP) on the clean and contaminated items of MMLU and MBPP. $\Delta$: contaminated minus clean.}
  \label{tab:contam-acc}
  \begin{tabular}{@{}lrrrrrr@{}}
    \toprule
    & \multicolumn{3}{c}{\textbf{MMLU}} & \multicolumn{3}{c@{}}{\textbf{MBPP}} \\
    \cmidrule(lr){2-4} \cmidrule(l){5-7}
    \textbf{Model} & \textbf{Clean} & \textbf{Contam.} & $\boldsymbol{\Delta}$ & \textbf{Clean} & \textbf{Contam.} & $\boldsymbol{\Delta}$ \\
    \midrule
    Llama-3.2-1B       & 31.5 & 30.8 & $-0.7$ & 39.5 & 23.9 & $-15.6$ \\
    granite-4.0-h-tiny & 64.4 & 66.4 & $+2.0$ & 65.4 & 57.0 & $-8.4$ \\
    Qwen3-1.7B         & 61.5 & 63.4 & $+1.9$ & 66.7 & 53.2 & $-13.5$ \\
    \ourrow \model     & 63.6 & 63.4 & $-0.2$ & 61.7 & 52.5 & $-9.2$ \\
    \bottomrule
  \end{tabular}
\end{table}

\paragraph{Likelihood.}
Memorization can also make the reference text more predictable without changing the answer. 
This shows in the bits per byte (BPB) of the reference text given the prompt: the reference code for MBPP, the question and its correct answer for MMLU.
BPB varies across models, so only the ratio of contaminated to clean items is comparable; it is below 1 even for a model that has never seen the items, because the two groups differ in length and style.
The ratio of \model lies within the reference ranges on both benchmarks (\cref{tab:contam-bpb}): 0.982 on MMLU (0.940 to 0.990) and 0.763 on MBPP (0.749 to 0.806).
Thus, \model does not exhibit an unusually large likelihood advantage on potentially contaminated items relative to the reference models.

\begin{table}[!h]
  \centering
  \footnotesize
  \caption{Bits per byte of the reference text on the clean and contaminated items of MMLU and MBPP, and their ratio (contaminated / clean). BPB values depend on the model, so only the ratios are comparable; a ratio clearly below those of the other models would indicate memorization.}
  \label{tab:contam-bpb}
  \begin{tabular}{@{}lrrrrrr@{}}
    \toprule
    & \multicolumn{3}{c}{\textbf{MMLU}} & \multicolumn{3}{c@{}}{\textbf{MBPP}} \\
    \cmidrule(lr){2-4} \cmidrule(l){5-7}
    \textbf{Model} & \textbf{Clean} & \textbf{Contam.} & \textbf{Ratio} & \textbf{Clean} & \textbf{Contam.} & \textbf{Ratio} \\
    \midrule
    Llama-3.2-1B       & 0.847 & 0.816 & 0.964 & 0.562 & 0.429 & 0.763 \\
    granite-4.0-h-tiny & 0.855 & 0.847 & 0.990 & 1.207 & 0.972 & 0.806 \\
    Qwen3-1.7B         & 0.826 & 0.776 & 0.940 & 0.408 & 0.306 & 0.749 \\
    \ourrow \model     & 0.790 & 0.776 & 0.982 & 0.507 & 0.387 & 0.763 \\
    \bottomrule
  \end{tabular}
\end{table}

\section{Inference Benchmark}
\label{app:inference}
We measured the inference speed of \model and of 45 of the 56 baselines of \cref{sec:evaluation}.
The other 11 were not measured: 10 are not natively supported by vLLM~0.28.0 or fail to load in vLLM~0.28.0, and Villanova2-2B was added to the comparison after the benchmark.
\Cref{tab:inference-user} reports the results, and \cref{fig:pareto-speed} plots the Overall average against the speed for one user.

\paragraph{Setup.}
Each model runs on a single NVIDIA A100 GPU with vLLM~0.28.0~\citep{kwon2023vllm} in bf16.
All models use the same engine settings, with CUDA graphs captured for batches of up to 2{,}048 tokens.
Prompts are real text from our held-out set, cut to the exact lengths specified by each model's tokenizer.
Every number is the median of up to three repeats.
Since tokenizers differ in density, from 245 to 495 tokens per 1{,}000 characters (273 for \model), we report characters per second.

\begin{table}[H]
  \centering
  \scriptsize
  \setlength{\tabcolsep}{3pt}
  \renewcommand{\arraystretch}{0.9} %
  \caption{Inference speed on one A100 GPU with vLLM, for 1{,}024-token prompts and 256 generated tokens: characters/s for one user, thousands of characters/s for 64 and 256 users, and median time to first token (TTFT, ms; under load it includes queueing). 128k: thousands of characters/s on 126{,}976-token prompts with 1{,}024 generated tokens (--: longer than the context of the model). Active: active parameters (B). Groups as in \cref{tab:main-results}, sorted from slowest to fastest for one user, \model last. $^\dagger$Some measurements from one or two repeats. \model is the eighth-fastest of the 46 models for one user and ties for fifth of the 29 that support 128k-token prompts.}
  \label{tab:inference-user}
  \begin{tabular}{@{}lrrrrrrrr@{}}
    \toprule
    & & \multicolumn{2}{c}{\textbf{1 user}} & \multicolumn{2}{c}{\textbf{64 users}} & \multicolumn{2}{c}{\textbf{256 users}} & \textbf{128k} \\
    \cmidrule(lr){3-4}\cmidrule(lr){5-6}\cmidrule(lr){7-8}\cmidrule(l){9-9}
    \textbf{Model} & \textbf{Active} & \textbf{Chars/s} & \textbf{TTFT} & \textbf{k chars/s} & \textbf{TTFT} & \textbf{k chars/s} & \textbf{TTFT} & \textbf{k chars/s} \\
    \midrule
    \multicolumn{9}{@{}l}{\textit{Models with more than 4B active parameters}} \\
    Mistral-7B-v0.3$^\dagger$ & 7.25 & 257 & 77.6 & 5.4 & 487 & 5.5 & 939 & -- \\
    Gemma-7B$^\dagger$ & 8.54 & 258 & 81.9 & 4.5 & 509 & 4.4 & 36{,}924 & -- \\
    Teuken-7B$^\dagger$ & 7.45 & 274 & 66.4 & 7.1 & 437 & 6.9 & 1{,}054 & -- \\
    Qwen3-8B$^\dagger$ & 8.19 & 281 & 74.2 & 5.9 & 499 & 5.7 & 3{,}799 & 9.2 \\
    Apertus-v1.5-8B$^\dagger$ & 8.00 & 294 & 75.3 & 6.4 & 476 & 6.6 & 930 & 10.2 \\
    Llama-3-8B & 8.03 & 298 & 77.0 & 6.4 & 483 & 6.7 & 938 & -- \\
    Llama-3.1-8B$^\dagger$ & 8.03 & 298 & 77.1 & 6.4 & 478 & 6.7 & 934 & 10.4 \\
    Qwen2.5-7B$^\dagger$ & 7.62 & 300 & 64.1 & 7.2 & 428 & 8.0 & 779 & 15.0 \\
    \midrule
    \multicolumn{9}{@{}l}{\textit{Models with 2B to 4B active parameters}} \\
    Gemma-4-E4B$^\dagger$ & 4.00 & 383 & 57.2 & 9.9 & 281 & 14.0 & 446 & 16.0 \\
    Hunyuan-4B & 4.22 & 427 & 46.6 & 8.1 & 335 & 7.5 & 805 & 10.1 \\
    Qwen3-4B & 4.02 & 440 & 43.2 & 8.4 & 327 & 7.7 & 785 & 10.3 \\
    Qwen3.5-4B$^\dagger$ & 4.66 & 465 & 52.6 & 9.4 & 312 & 10.7 & 537 & 32.2 \\
    granite-4.1-3b$^\dagger$ & 3.40 & 484 & 40.3 & 10.5 & 281 & 8.7 & 713 & 13.9 \\
    SmolLM3-3B & 3.08 & 538 & 36.3 & 12.5 & 228 & 12.9 & 481 & -- \\
    tiny-aya & 3.35 & 542 & 40.2 & 11.7 & 233 & 11.2 & 489 & -- \\
    MiniCPM5-2B$^\dagger$ & 2.50 & 596 & 33.0 & 15.1 & 200 & 13.2 & 486 & 19.5 \\
    salamandra-2b & 2.30 & 756 & 23.8 & 11.4 & 136 & 13.0 & 1{,}404 & -- \\
    \midrule
    \multicolumn{9}{@{}l}{\textit{Models with up to 2B active parameters}} \\
    Bielik-1.5B & 1.60 & 432 & 26.0 & 11.3 & 149 & 11.0 & 606 & -- \\
    Falcon-H1-1.5B-Deep$^\dagger$ & 1.55 & 432 & 76.9 & 7.3 & 367 & 9.5 & 501 & 21.3 \\
    Trinity-Nano & 1.00 & 573 & 34.9 & 10.7 & 167 & 13.0 & 370 & 71.4 \\
    moe-7b-1b & 1.10 & 613 & 33.0 & 8.9 & 176 & 12.0 & 383 & -- \\
    Gemma-4-E2B$^\dagger$ & 2.00 & 655 & 36.2 & 17.1 & 166 & 25.0 & 283 & 18.2 \\
    granite-4.0-h-tiny & 1.00 & 745 & 40.1 & 10.4 & 190 & 16.8 & 343 & 95.3 \\
    granite-4.0-1b$^\dagger$ & 1.60 & 749 & 27.4 & 14.4 & 190 & 13.7 & 431 & 18.3 \\
    SmolLM2-1.7B & 1.70 & 754 & 26.1 & 10.1 & 166 & 10.6 & 1{,}181 & -- \\
    Falcon-H1-1.5B & 1.55 & 771 & 34.1 & 13.9 & 147 & 15.3 & 364 & 45.1 \\
    Hunyuan-1.8B & 1.80 & 781 & 27.3 & 16.6 & 184 & 16.2 & 375 & 22.9 \\
    Falcon-H1-0.5B & 0.50 & 796 & 42.6 & 14.4 & 200 & 18.8 & 284 & -- \\
    Qwen2.5-1.5B & 1.50 & 833 & 23.4 & 23.6 & 139 & 23.0 & 399 & 35.7 \\
    granite-3.1-3b-a800m$^\dagger$ & 0.90 & 842 & 25.8 & 12.7 & 141 & 14.2 & 351 & 22.1 \\
    Qwen3-1.7B$^\dagger$ & 1.70 & 872 & 26.3 & 15.2 & 148 & 16.1 & 342 & 19.7 \\
    Qwen3.5-2B & 1.90 & 904 & 33.3 & 20.3 & 166 & 24.4 & 295 & 77.5 \\
    granite-4.0-h-1b & 1.50 & 918 & 41.3 & 16.2 & 204 & 20.7 & 285 & 112 \\
    Luciole-1B$^\dagger$ & 1.32 & 987 & 22.5 & 19.9 & 142 & 16.8 & 398 & 26.6 \\
    LFM2.5-8B-A1B & 1.70 & 995 & 34.2 & 13.5 & 160 & 23.3 & 297 & 76.7 \\
    OLMoE-1B-7B & 1.30 & 1{,}005 & 27.6 & 10.2 & 141 & 15.9 & 292 & -- \\
    EuroLLM-1.7B & 1.70 & 1{,}037 & 23.5 & 17.2 & 126 & 18.2 & 311 & -- \\
    Gemma-3-1B & 1.00 & 1{,}068 & 19.6 & 31.0 & 83 & 38.3 & 199 & -- \\
    Llama-3.2-1B & 1.20 & 1{,}351 & 19.8 & 29.9 & 114 & 26.7 & 355 & 43.6 \\
    LFM2.5-1.2B & 1.20 & 1{,}362 & 19.9 & 35.1 & 125 & 40.5 & 1{,}441 & 85.7 \\
    Apertus-v1.1-1.5B & 1.50 & 1{,}392 & 19.2 & 29.8 & 112 & 27.0 & 355 & -- \\
    EuroMoE-2.6B-A0.6B & 0.50 & 1{,}440 & 18.1 & 23.9 & 73 & 33.0 & 230 & -- \\
    Qwen3.5-0.8B & 0.80 & 1{,}561 & 32.2 & 26.3 & 125 & 32.5 & 212 & 92.7 \\
    Qwen3-0.6B$^\dagger$ & 0.60 & 1{,}567 & 16.4 & 20.0 & 92 & 20.1 & 267 & 21.7 \\
    Apertus-v1.1-0.5B & 0.40 & 2{,}243 & 12.7 & 49.5 & 79 & 55.2 & 779 & -- \\
    \ourrow \model & 0.65 & 1{,}088 & 33.9 & 12.9 & 143 & 18.5 & 305 & 77.5 \\
    \bottomrule
  \end{tabular}
\end{table}

\newpage
\section{T-HellaSwag: Construction and Release}
\label{app:t-hellaswag}
T-HellaSwag is a higher-quality machine translation of HellaSwag~\citep{zellers2019hellaswag} into 110 language variants, which we will release.
The HellaSwag test-split labels are not public, so we translated the 10{,}042 documents in its validation split into each variant using Lara\footnote{\url{https://laratranslate.com}}, Translated's machine translation service.
T-HellaSwag contains 200 of these documents per language variant, selected for translation quality and stratified by source and difficulty.
T-HellaSwag-full contains all 10{,}042 translated validation documents per variant, without selection.

\begin{figure}[H]
  \centering
  \begin{tikzpicture}[font=\sffamily\footnotesize,
    fam/.style={anchor=east, inner sep=3pt, text=chartink},
    num/.style={anchor=west, inner sep=3pt, text=chartink, font=\sffamily\footnotesize\bfseries},
    inS/.style={text=white, font=\sffamily\scriptsize\bfseries, inner sep=0pt},
    inO/.style={text=chartink, font=\sffamily\scriptsize, inner sep=0pt},
    sup/.style={fill=brandaccent, draw=white, line width=0.6pt},
    oth/.style={fill=chartaxis!70, draw=white, line width=0.6pt}]
  \def\U{0.19}\def\H{0.34}\def\R{0.46}
  \newcommand{\cv}[4]{%
    \node[fam] at (0,-#1*\R) {#4};
    \ifnum#2>0 \fill[sup] (0,-#1*\R-\H/2) rectangle (#2*\U,-#1*\R+\H/2);\fi
    \ifnum#3>0 \fill[oth] (#2*\U,-#1*\R-\H/2) rectangle ({(#2+#3)*\U},-#1*\R+\H/2);\fi
    \ifnum#2>1 \node[inS] at ({#2*\U/2},-#1*\R) {#2};\fi
    \ifnum#3>1 \node[inO] at ({(#2+#3/2)*\U},-#1*\R) {#3};\fi
    \pgfmathtruncatemacro\tot{#2+#3}\node[num] at ({(#2+#3)*\U},-#1*\R) {\tot};}
  \foreach \x in {10,20,30,40,50} {\draw[chartgrid, line width=0.4pt] (\x*\U,{\R/2}) -- (\x*\U,{-14*\R-\R/2});
    \node[anchor=north, text=chartsecondary, font=\sffamily\scriptsize] at (\x*\U,{-14*\R-\R/2-0.04}) {\x};}
  \cv{0}{32}{26}{Indo-European}
  \cv{1}{0}{13}{Atlantic-Congo}
  \cv{2}{3}{4}{Afro-Asiatic}
  \cv{3}{2}{5}{Austronesian}
  \cv{4}{2}{4}{Turkic}
  \cv{5}{0}{4}{Dravidian}
  \cv{6}{2}{1}{Sino-Tibetan}
  \cv{7}{3}{0}{Uralic}
  \cv{8}{1}{1}{Austroasiatic}
  \cv{9}{1}{1}{Tai-Kadai}
  \cv{10}{0}{1}{Basque (isolate)}
  \cv{11}{1}{0}{Japonic}
  \cv{12}{0}{1}{Kartvelian}
  \cv{13}{1}{0}{Koreanic}
  \cv{14}{0}{1}{Mongolic-Khitan}
  \draw[chartaxis, line width=0.6pt] (0,{\R/2}) -- (0,{-14*\R-\R/2});
  \node[anchor=north, text=chartsecondary, font=\sffamily\scriptsize] at (29*\U,{-14*\R-\R/2-0.38}) {Language variants};
  \node[anchor=south west, inner sep=0pt, text=chartsecondary] at (0,{\R/2+0.15})
    {\textcolor{brandaccent}{\rule{7pt}{7pt}}~Languages supported by \model\ (48 variants)\quad
     \textcolor{chartaxis!70}{\rule{7pt}{7pt}}~Other languages (62 variants)};
\end{tikzpicture}
  \caption{The 110 language variants of T-HellaSwag by language family in Glottolog~\citep{hammarstrom2026glottolog}. Indo-European accounts for 58 of the variants, and five families, including Atlantic-Congo with 13 variants, occur only among languages that \model does not support.}
  \label{fig:t-hellaswag-coverage}
\end{figure}
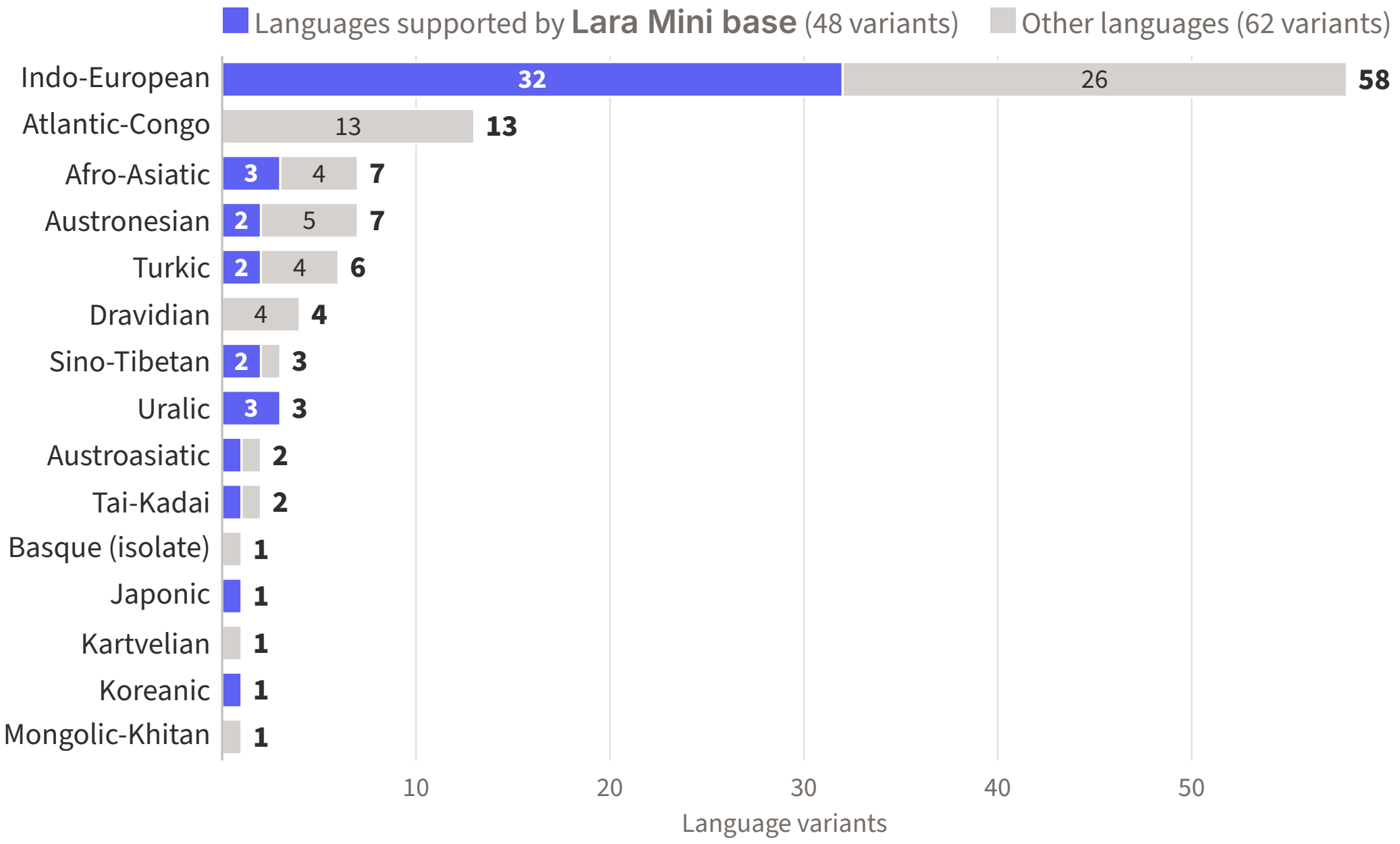

\paragraph{Languages.}
\Cref{tab:app-t-hellaswag-languages} lists the 110 language variants: 48 for the 44 non-English languages supported by \model and 62 for other languages.
French, Portuguese and Spanish each have two regional variants, while Serbian has two script variants.
The 105 languages belong to 15 families (\cref{fig:t-hellaswag-coverage}).

\begin{table}[H]
  \centering
  \footnotesize
  \caption{The 110 language variants of T-HellaSwag and T-HellaSwag-full.}
  \label{tab:app-t-hellaswag-languages}
  \begin{tabularx}{\linewidth}{@{}lL@{}}
    \toprule
    \begin{tabular}[t]{@{}l@{}}44 non-English languages\\supported by \model\\(48 variants)\end{tabular}
      & Arabic, Azerbaijani, Bengali, Bosnian, Bulgarian, Catalan, Chinese (Simplified), Chinese (Traditional), Croatian, Czech, Danish, Dutch, Estonian, Finnish, French (Canada), French (France), German, Greek, Hebrew, Hindi, Hungarian, Indonesian, Irish, Italian, Japanese, Korean, Latvian, Lithuanian, Malay, Maltese, Norwegian (Bokm\aa l), Persian, Polish, Portuguese (Brazil), Portuguese (Portugal), Romanian, Russian, Serbian (Cyrillic), Serbian (Latin), Slovak, Slovenian, Spanish (Mexico), Spanish (Spain), Swedish, Thai, Turkish, Ukrainian, Vietnamese \\
    \midrule
    \begin{tabular}[t]{@{}l@{}}Other languages\\(62 variants)\end{tabular}
      & Afrikaans, Albanian, Amharic, Armenian, Assamese, Basque, Belarusian, Bhojpuri, Burmese, Cebuano, Chichewa, Galician, Georgian, Gujarati, Haitian Creole, Hausa, Icelandic, Igbo, Javanese, Kannada, Kazakh, Khmer, Kinyarwanda, Kurdish (Kurmanji), Kurdish (Sorani), Kyrgyz, Lao, Lingala, Luganda, Luxembourgish, Macedonian, Maithili, Malagasy, Malayalam, Marathi, Mongolian, Nepali, Northern Sotho, Odia, Oromo (West Central), Pashto, Punjabi, Shona, Sindhi, Sinhala, Somali, Southern Sotho, Sundanese, Swahili, Tagalog, Tajik, Tamil, Telugu, Tswana, Urdu, Uyghur, Uzbek, Welsh, Xhosa, Yiddish, Yoruba, Zulu \\
    \bottomrule
  \end{tabularx}
\end{table}

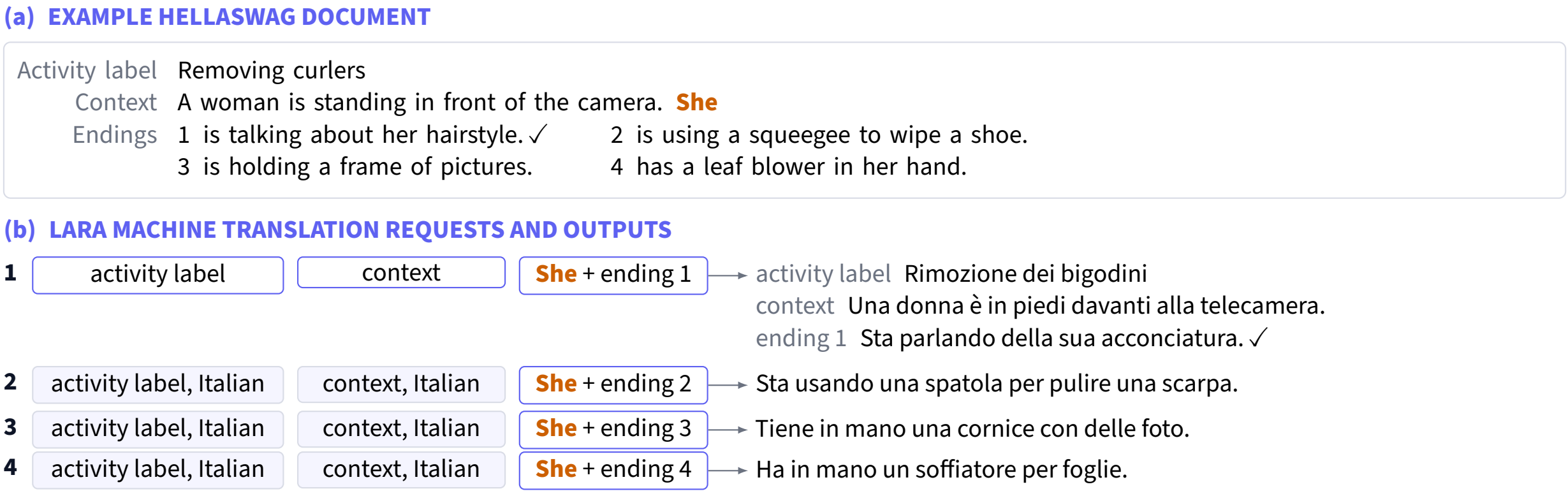
\begin{figure}[H]
  \centering
  \begin{tikzpicture}[font=\sffamily\scriptsize,
      hdr/.style={anchor=south west, font=\sffamily\scriptsize\bfseries, text=brandaccent, inner sep=0pt},
      seg/.style={anchor=north west, draw=brandaccent, fill=white, rounded corners=1.5pt, inner xsep=2.5pt,
        inner ysep=2pt, minimum width=#1},
      lock/.style={seg=#1, draw=brandrule, fill=brandtint},
      ret/.style={anchor=north west, text width=8.4cm, align=flush left, inner xsep=0pt, inner ysep=2pt},
      req/.style={anchor=north west, font=\sffamily\scriptsize\bfseries, text=brandink, inner xsep=0pt, inner ysep=2pt},
      to/.style={-{Latex[length=3pt]}, draw=brandmuted},
    ]
    \providecommand{\checkmark}{\ensuremath{\surd}}%
    \def\tg#1{\textcolor{brandmuted}{#1}}
    \def\fr#1{\textcolor{series3}{\textbf{#1}}}
    \node[hdr] at (0,0.12) {(a)\enspace EXAMPLE HELLASWAG DOCUMENT};
    \node[anchor=north west, draw=brandrule, rounded corners=2pt, inner sep=4pt, text width=\linewidth-8pt] (doc) at (0,0) {%
      \noindent\begin{tabular}{@{}r@{\hspace{6pt}}l@{\hspace{18pt}}l@{}}
        \tg{Activity label} & Removing curlers \\
        \tg{Context} & \multicolumn{2}{@{}l@{}}{A woman is standing in front of the camera. \fr{She}} \\
        \tg{Endings} & 1\enspace is talking about her hairstyle.\,$\checkmark$ & 2\enspace is using a squeegee to wipe a shoe. \\
                     & 3\enspace is holding a frame of pictures. & 4\enspace has a leaf blower in her hand. \\
      \end{tabular}};
    \node[hdr] at ([yshift=-13pt]doc.south west) {(b)\enspace LARA MACHINE TRANSLATION REQUESTS AND OUTPUTS};
    \coordinate (xa) at (0.3,0); \coordinate (xb) at (3.05,0); \coordinate (xc) at (5.35,0); \coordinate (xo) at (7.8,0);
    \def\threq#1#2#3#4{%
      \node[req] at (#1) {#2};
      #3
      \node[seg=1.95cm] (e#2) at (#1 -| xc) {\fr{She} + ending #2};
      \node[ret] (o#2) at (#1 -| xo) {#4};
      \draw[to] (e#2.east) -- ([xshift=-2pt]e#2.east -| xo);}
    \def\locked#1{\node[lock=2.6cm] at (#1 -| xa) {activity label, Italian}; \node[lock=2.15cm] at (#1 -| xb) {context, Italian};}
    \coordinate (r1) at ([yshift=-17pt]doc.south west);
    \threq{r1}{1}{\node[seg=2.6cm] at (r1 -| xa) {activity label}; \node[seg=2.15cm] at (r1 -| xb) {context};}%
      {\tg{activity label}\enspace Rimozione dei bigodini\\
       \tg{context}\enspace Una donna \`e in piedi davanti alla telecamera.\\
       \tg{ending 1}\enspace Sta parlando della sua acconciatura.\,$\checkmark$}
    \coordinate (r2) at ([yshift=-2pt]o1.south west -| 0,0);
    \threq{r2}{2}{\locked{r2}}{Sta usando una spatola per pulire una scarpa.}
    \coordinate (r3) at ([yshift=-2pt]o2.south west -| 0,0);
    \threq{r3}{3}{\locked{r3}}{Tiene in mano una cornice con delle foto.}
    \coordinate (r4) at ([yshift=-2pt]o3.south west -| 0,0);
    \threq{r4}{4}{\locked{r4}}{Ha in mano un soffiatore per foglie.}
  \end{tikzpicture}
  \caption{A HellaSwag document and its translation into Italian. (a)~The source document, with the subject fragment at the end of the context shown in orange and the correct ending marked ($\checkmark$). (b)~The four requests to Lara. The subject fragment is moved to the start of each ending; the translated activity label and context from request 1 are sent as non-translatable segments in requests 2--4 (shaded).}
  \label{fig:app-t-hellaswag-requests}
\end{figure}

\paragraph{Translation.}
HellaSwag draws its documents from ActivityNet Captions~\citep{krishna2017dense} and wikiHow\footnote{\url{https://www.wikihow.com}}.
Each document contains an activity label, a context and four candidate endings.
The task is to choose the ending that best continues the context (\cref{fig:app-t-hellaswag-requests}(a)).
The translated document must retain one context shared by all four endings, and each translated ending must remain consistent with that context.
We translated each document in four requests, one per ending.
Each request sent the activity label, context and ending as separate segments (\cref{fig:app-t-hellaswag-requests}(b)).
The first request translated all three segments; requests 2 to 4 resent the translated activity label and context as non-translatable segments.
For ActivityNet documents, HellaSwag splits a sentence into context and endings: the context ends with the subject, most often a short phrase such as ``he'' or ``the man''.
A subject cut off from its verb often cannot be translated correctly because, in many languages, the verb determines its case or position, or allows it to be dropped.
We therefore moved this subject fragment to the start of each ending before translation.
To keep activity labels consistent across documents, we also maintained a lexicon of them for each language.
Each label was translated once, and subsequent requests sent its translation as a non-translatable segment.
After translation, we restored wikiHow markup, such as \texttt{[title]}, using a rule-based script in the fields where Lara had translated.
We used Claude Opus~5.5~\citep{anthropic2026opus55} to correct four types of translation error: fields in the wrong language or script; characters or words of another script within a field; English text left untranslated within a field; and two endings in the same document with identical translations.

\begin{figure}[H]
  \centering
  \begin{tikzpicture}[font=\sffamily\scriptsize,
      stage/.style={draw=brandaccent, fill=white, rounded corners=1.5pt, text width=2.3cm, align=flush center,
        minimum height=1.4cm, inner xsep=2pt, inner ysep=2pt, execute at begin node={\hyphenpenalty=10000}},
      relnote/.style={text width=2.3cm, align=flush center, inner sep=1pt, font=\sffamily\scriptsize\itshape, text=brandink},
      to/.style={-{Latex[length=3pt]}, draw=brandmuted},
    ]
    \def\st#1#2{{\bfseries\color{brandink}#1}\\[1pt]{\color{brandmuted}#2}}
    \node[stage] (a) {\st{Translate}{10{,}042 validation\\documents}};
    \node[stage, right=0.25cm of a] (b) {\st{Score}{quality estimation\\by the QE models\\that pass}};
    \node[stage, right=0.25cm of b] (c) {\st{Screen}{automatic checks:\\7{,}553--10{,}008 left}};
    \node[stage, right=0.25cm of c] (d) {\st{Stratify}{13 strata by\\source and\\difficulty}};
    \node[stage, right=0.25cm of d] (e) {\st{Select}{top-scoring share\\of each stratum:\\200 documents}};
    \foreach \u/\v in {a/b, b/c, c/d, d/e} \draw[to] (\u.east) -- (\v.west);
    \node[relnote, below=3pt of a] {to be released as\\T-HellaSwag-full};
    \node[relnote, below=3pt of e] {to be released as\\T-HellaSwag};
  \end{tikzpicture}
  \caption{Construction of each T-HellaSwag language variant, with the number of documents remaining per variant after each step. The translated validation split will be released in full as T-HellaSwag-full. Selecting the same proportion from every stratum preserves the source and difficulty distributions of the English originals.}
  \label{fig:app-t-hellaswag-selection}
\end{figure}
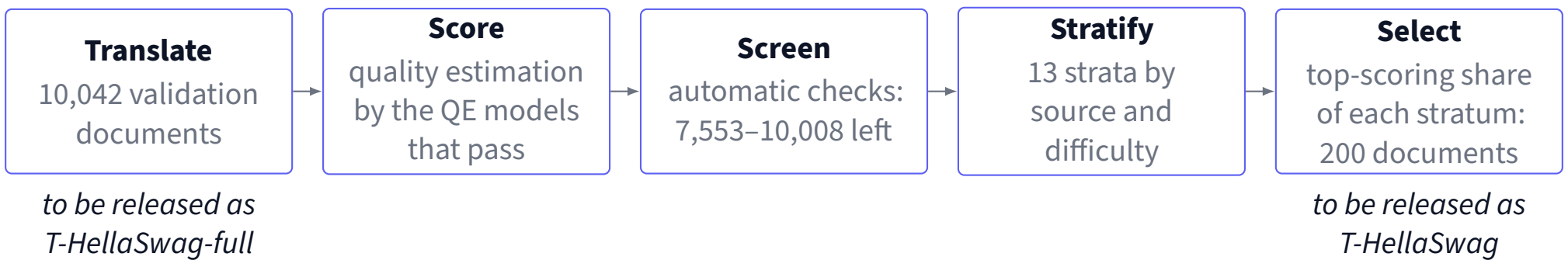

\paragraph{Selection.}
For each language variant, we selected the 200 documents with the highest translation-quality scores, stratifying the selection by source and by the difficulty of the English originals (\cref{fig:app-t-hellaswag-selection}).
We evaluated eight reference-free quality estimation (QE) models in each language variant to assess how reliably they ranked translations: CometKiwi-22~\citep{rei2022cometkiwi}, CometKiwi-XL and CometKiwi-XXL~\citep{rei2023scaling}, xCOMET-XL and xCOMET-XXL~\citep{guerreiro2024xcomet}, MetricX-24-Hybrid-XXL~\citep{juraska2024metricx}, BLASER~2.0-QE~\citep{dale2024blaser}, and COMET-poly~\citep{zufle2025cometpoly}.
We tested each model on 1{,}000 sampled documents per language variant with four corruptions of an ending: word deletion, word shuffling, replacement with another ending of the same document and replacement with an ending of another document.
A model passed for a language if, for each corruption, it scored the original ending above the corrupted one in at least 95\% of the comparisons.
Six of the eight models passed in at least one language.
In the three languages where no model passed, we used MetricX-24 alone with lower thresholds: 90\% for Luxembourgish and Luganda, and 85\% for West Central Oromo.
The models that passed for a language scored all four endings of every document.
Because a single faulty ending invalidates a document, each model rated a document by the percentile rank of its worst ending, and a document's score was the mean of these ranks over the models that passed.

Before selection, automatic checks excluded documents containing untranslated English, fields in the wrong script or language~\citep{kargaran2023glotlid}, or retranslated fields.
Selecting by translation quality alone under-represented ActivityNet: its share of the selected documents fell from 32.4\% to 15.0\% on average.
Such a selection would therefore distort both the source and the difficulty distributions of the dataset.
We divided the remaining documents into 13 strata by source (ActivityNet or wikiHow) and by the difficulty of the English original, estimated with 50 public base models.
We selected the same proportion of the highest-scoring documents from each stratum, yielding 200 documents per language variant.
To verify that the difficulty was preserved, we evaluated 35 additional public base models, not used to build the strata, on the selected documents.
The mean absolute difference in English accuracy between the selected documents and the original split was 1.6 points, against 3.2 points when selecting by quality alone and 1.5 points for random documents from the same strata.
Each language variant has its own selection of documents, so part of a score difference between two variants comes from the documents.
T-HellaSwag is built to compare models within each variant.

\paragraph{Quality.}
Automatic checks verified that every translated document was complete and that the four endings of each document shared the same translated activity label and context.
On the final data, every check passed in both sets and in all 110 language variants.
We compared T-HellaSwag with Multilingual HellaSwag~\citep{lai2023okapi}.
The latter's construction differs from ours in two ways.
It keeps the activity labels in English, so that each of its prompts begins with English text.
It also translates the subject that ends each ActivityNet context separately from the endings, so that the two parts of the split sentence may not agree.
For each of the 32 languages in both datasets, we drew 50 random T-HellaSwag documents that Multilingual HellaSwag also translates, and compared the two translations of each document.
Two LLM judges, Gemini~4 Argon~\citep{google2026gemini4argon} and Claude Opus~5.5, preferred our translation for 98.6\% and 98.7\% of the documents, respectively.
A human linguist reviewed 50 documents in each of Italian, French and Russian.
Of the 150 documents, 145 were rated OK, 5 as having a minor error and none as having a major error.

\paragraph{Limitations.}
Apart from the linguist review in three languages, the evidence of translation quality comes from QE models and LLM judges.
No human translator revised either set, so both can contain machine-translation errors, such as stray characters and line breaks.
T-HellaSwag-full is not filtered: it also keeps the documents that the automatic checks excluded from T-HellaSwag.
Both sets retain the situations described in the English source, so they measure commonsense reasoning in each language variant, not knowledge specific to its culture.
Because HellaSwag is public, a model that encountered an English document during training may recognize its translation.

\paragraph{Release.}
Both sets will be released with one configuration per set and language variant, e.g., \texttt{it-IT} and \texttt{it-IT-full} for Italian.
They share 50 few-shot documents per language variant from the HellaSwag training split, so that no validation document serves as a few-shot example.
We translated the few-shot documents in the same way as the validation documents and selected them based on the scores from the same QE models.
We score each language variant in the same way as the standard HellaSwag task scores English.
The release will credit HellaSwag, wikiHow and ActivityNet Captions~\citep{krishna2017dense} as sources.

\section{Model Card}
\label{app:model-card}

\begin{table}[H]
  \centering
  \small
  \caption{Model card of \model.}
  \label{tab:model-card}
  \begin{tabularx}{\linewidth}{@{}lL@{}}
    \toprule
    Developer        & Translated \\
    Model type       & Decoder-only mixture of experts base model; hybrid of Gated DeltaNet, sliding-window and global attention (\cref{sec:architecture}) \\
    Parameters       & 7.2B total, 0.6B active per token (embeddings included) \\
    Layers           & 34: 14 Gated DeltaNet (GDN), 13 sliding-window attention (window 512, RoPE), 7 global attention (no positional encoding) \\
    Layer pattern    & GDN, sliding, GDN, global, then (sliding, GDN, sliding, GDN, global) $\times$ 6 \\
    Hidden size      & 1{,}024 \\
    Attention        & 16 query heads, 4 key-value heads, head dimension 128, QK-norm \\
    Gated DeltaNet   & 8 key heads, 16 value heads, head dimension 128, short convolution of width 4 \\
    Mixture of experts & In every layer: 128 routed experts (top-4, fp32 router) and 1 shared expert, all SwiGLU with hidden size 512 \\
    Vocabulary       & 151{,}936 tokens; input and output embeddings tied \\
    Context length   & 4{,}096 tokens in the first three stages; 131{,}072 after the long-context stage (\cref{sec:context-extension}) \\
    Precision        & bf16 \\
    Training data    & 6T tokens of public data: English and multilingual text, code, math and SFT-like data (\cref{sec:data}), then 100B tokens in the long-context stage \\
    Languages        & English and 44 others, which together cover all 24 official EU languages (\cref{tab:app-languages}) \\
    License          & CC BY-NC-SA 4.0 (\url{https://creativecommons.org/licenses/by-nc-sa/4.0/}) \\
    Intended use     & Research, and a starting point for fine-tuning and post-training; a base model, not tuned to follow instructions or for safety \\
    Limitations      & See \cref{sec:conclusion} \\
    \bottomrule
  \end{tabularx}
\end{table}

\clearpage
\section*{Contributors}
\phantomsection
\addcontentsline{toc}{section}{Contributors}
\label{app:contributors}
\model is the work of the Lara Team at Translated.
All contributors are with Translated and are listed in alphabetical order by last name.

\medskip
{\raggedright
Maria~Pia~Adinolfi,
Isabelle~Andrieu,
Grazia~Fiorella~Anfuso,
Melania~Annunziata,
Daniele~Antonini,
Marta~Arena,
Luca~Barone,
Dana~Barras~Franco,
Francesca~Basiric{\`o},
Francesca~Bee,
Sofia~Bertini,
Nicola~Bertoldi,
Giorgio~Biancini,
Alessandra~Bigi,
Giulia~Boccalini,
Tommaso~Bonomo,
Enrico~Boscardi,
Sebastien~Bratieres,
Francesco~Brigante,
Giulia~Brignola,
Romolo~Buondestino,
Brenda~Patricia~Byrne,
Giorgia~Campora,
Edoardo~Cappelli,
Veronica~Carascosi,
Francesca~Carducci,
Cody~Carlson,
Davide~Caroselli,
Ambra~Cascio,
Dario~Cascio,
Mauro~Cassani,
Andrea~Castro,
Alessandro~Cattelan,
Giulia~Ceccacci,
Paola~Cellini,
Giulia~Chiriatti,
Federico~Cialini,
Valentina~Ciarlariello,
Daniele~Cocozza,
Emanuela~Coghe,
Kathleen~Compares,
Marco~Comunit{\`a},
Davide~Conti,
Helen~Rebecca~Corfield,
Michele~Cortellessa,
Anna~Cracco,
Aurora~Cuppone,
Arianna~D'Andrea,
Maurizio~Davoli,
Andrea~De~Angelis,
Andrea~De~Blasi,
Rebecca~De~Giorgi,
Annalisa~De~Petra,
Arianna~De~Pilla,
Joris~Demiraj,
Elena~Deufemia,
Federica~Di~Antonio,
Pierluigi~Di~Cianni,
Vincenzo~Di~Cicco,
Michela~Di~Clemente,
Alan~Di~Francesco,
Ylenia~Di~Gese,
Federica~Di~Iulio,
Claudia~Di~Lorenzo,
Andrea~Di~Paola,
Giulia~Di~Pilla,
Jessica~Dickson,
Francesco~Elefante,
Ilaria~Fabrizio,
Alessandro~Falappa,
Viviana~Fantozzi,
Antonio~Farina,
Andrea~Fernandez~Tostado~Ro.Diego~Andrea,
Andrea~Ferrari,
Sara~Ferri,
Simona~Ferri,
Fabio~Fioramonti,
Stefano~Florio,
Mariachiara~Fornaro,
Anna~Maria~Foti,
Niccol{\`o}~Fransoni,
Alessandro~Fusacchia,
Emma~Gamba,
Francesco~Gasperoni,
Davide~Gattini,
Marta~Giani,
Claudia~Giannetti,
Valerio~Giannini,
Antonella~Giudici,
Mara~Gramendola,
Michele~Grandinetti,
Eleonora~Graziani,
L'Meese~Hope~Greaney,
Dario~Gubernari,
Filomena~Guglielmi,
Jonathan~Hammond,
Stephen~William~Harris,
Rowena~Elizabeth~Hogg,
Marta~Hontanon~Del~Olmo,
Marco~Ilari,
Kenneth~Larry~Jackson~Jr,
Sagar~Joshi,
Bartosz~Alberto~Karman,
Iuliia~Khimenko,
Valerio~Lanni,
Giulio~Leonardi,
Veronica~Levco,
Matteo~Licci,
Alessio~Ligios,
Noemi~Lomartire,
Domenico~Lupinetti,
Jennifer~Mirjam~Maechtlen,
Chiara~Maggiori,
Maria~Luisa~Malerba,
Francesco~Marconi,
Domenico~Marini,
Francesca~Marsili,
Andrea~Martinelli,
James~Mc~Groary,
Matteo~Menicacci,
Andrew~Steven~Miller,
Marco~Moauro,
Giuliana~Molon,
Ilaria~Mongelluzzi,
Emiliano~Morghen,
Giovanni~Moscati,
Filippo~Musmeci,
Flavia~Napoleoni,
Nicola~Giuseppe~Ninni,
Beatrice~Ottoni,
Federica~Pagliani,
Nicol{\`o}~Palmiero,
Daniele~Palumbo,
Chiara~Pantani,
Stefano~Paolucci,
Deborah~Paritario,
Michela~Pascale,
Thomas~Passera,
Daniele~Patrioli,
Chiara~Patrizi,
Silvia~Patrizi,
Giulia~Pelos,
Simone~Giovanni~Perone,
Cristiano~Perrone,
Joshua~Peterson,
Katia~Piredda,
Francesca~Piscitelli,
Alessandra~Posca,
Lorenzo~Pratesi,
Francesca~Prencipe,
Daniela~Proietto~Cuco,
Mattia~Pupino~Marangio,
Disheng~Qiu,
Mirco~Ramo,
Federica~Rapagn{\`a},
Waldo~Luis~Ribeiro,
Federico~Ricciuti,
Alessia~Ridoni,
Eleonora~Risi~Valdettaro,
Jon~Matthew~Ritzdorf,
Serena~Rivera,
Federico~Roma,
Arianna~Ruggiero,
Eleonora~Ruggiu,
Mariafrancesca~Sagliocco,
Vanessa~Salvi,
Alessio~Sampieri,
Elena~Santanelli,
Roberto~Sastre,
Luca~Scofano,
Alessandro~Sdei,
Giuseppe~Sebastiani,
Claudia~Nicole~Siegenthaler,
Monica~Signoretti,
Nicola~Sobieski,
Biancamaria~Soellner,
Valentina~Stati,
Michael~Stevens,
Arianna~Stissi,
Fabiana~Stomaci,
Leonardo~Taddei,
Christopher~Thomas,
Maurizio~Tiberi,
John~Tinsley,
Riccardo~Tittarelli,
Carla~Tocco,
Martina~Totaro,
Barbara~Veronika~Totzauer,
Alessio~Tranzocchi,
Letizia~Treossi,
Mara~Trombetti,
Marco~Trombetti,
Andrea~Valenzi,
Kirti~Vashee,
Giulia~Vestita,
Teodosio~Vezza,
Giulia~Vignola,
Sara~Virtuoso,
Valentina~Volpini,
Francesca~Vurro,
Jolita~Zaksauskaite,
Giulia~Zanarella,
Paolo~Zanotti,
Francesca~Zavettieri,
Irene~Zorzi.
\par}

\end{document}